\documentclass[english]{article}
\usepackage{geometry}
\usepackage[T1]{fontenc}
\usepackage[latin9]{inputenc}
\usepackage[unicode=true,
 bookmarks=false,
 breaklinks=false,pdfborder={0 0 1},colorlinks=false]
 {hyperref}
\hypersetup{colorlinks,
linkcolor=red,
anchorcolor=blue,
citecolor=blue,
urlcolor=blue}

\makeatletter
\usepackage{bm}
\usepackage{amsmath,mathtools}
\usepackage{amssymb}
\usepackage{amsthm}  
\usepackage{comment}
\usepackage{natbib}
\usepackage{enumitem}
\usepackage{booktabs}
\usepackage{graphicx}
\usepackage[linesnumbered,ruled,vlined]{algorithm2e}
\usepackage{float}
\usepackage{multirow}
\usepackage{dsfont}
\usepackage{tcolorbox}
\usepackage{hyperref}
\usepackage{url}
\usepackage{color}
\allowdisplaybreaks

\definecolor{jw}{RGB}{0,0,200}
\definecolor{cc}{RGB}{231,117,0}

\usepackage{graphicx}
\usepackage{subcaption}
\usepackage{geometry}

\usepackage{booktabs}
\usepackage{multirow}
\usepackage{subcaption}
\usepackage{tcolorbox}

\title{Transformer-Based Masked Diffusion Models Provably Learn Distributions with Latent Structure}
\date{\today}

\author{
  Jingda Wu\thanks{Department of Industrial and Operations Engineering, University of Michigan, Ann Arbor, USA.
  {Email: \texttt{\{wujingda,cxcai\}@umich.edu}.}
  }
  \and
  Changxiao Cai\footnotemark[1]
}

\theoremstyle{plain} 
\newtheorem{lemma}{\bf Lemma} 
\newtheorem{proposition}{\bf Proposition}
\newtheorem{theorem}{\bf Theorem}

\theoremstyle{remark}
\newtheorem{assumption}{\bf Assumption} 
\newtheorem{definition}{\bf Definition}

\newtheorem{remark}{\bf Remark}

\newcommand{\bE}{\mathbb E}
\newcommand{\bP}{\mathbb P}
\newcommand{\bR}{\mathbb R}
\newcommand{\bV}{\mathbb V}
\newcommand{\bN}{\mathbb N}

\newcommand{\KL}{D_{\mathrm{KL}}}
\newcommand{\TC}{\mathsf{TC}}
\newcommand{\DTC}{\mathsf{DTC}}

\newcommand{\Proj}{\mathsf{Proj}}
\newcommand{\Softmax}{\mathsf{Softmax}}
\newcommand{\softmax}{\mathrm{softmax}}
\newcommand{\relu}{\operatorname{ReLU}}
\newcommand{\mask}{\mathtt M}

\newcommand{\head}{\mathrm{head}}
\newcommand{\attn}{\mathrm{attn}}
\newcommand{\Attn}{\mathsf{Attn}}
\newcommand{\FFN}{\mathsf{FFN}}
\newcommand{\ff}{\mathrm{ff}}
\newcommand{\diff}{\mathrm{d}}
\newcommand{\polylog}{\mathsf{poly}\log}
\newcommand{\TF}{\mathsf{TF}}
\newcommand{\Embed}{\mathsf{Embed}}
\newcommand{\embed}{\mathrm{embed}}
\newcommand{\OutEmbed}{\mathsf{OutEmbed}}
\newcommand{\tf}{\mathrm{tf}}
\newcommand{\supp}{\mathsf{supp}}
\newcommand{\loss}{\mathcal L}
\newcommand{\F}{\mathcal F}

\newcommand{\NN}{\mathsf{NN}}
\newcommand{\ind}{\mathbf1}
\newcommand{\LSE}{\operatorname{LSE}}
\newcommand{\ReLU}{\operatorname{ReLU}}
\newcommand{\Clip}{\mathrm{Clip}}
\newcommand{\pa}{\operatorname{pa}}
\newcommand{\ch}{\operatorname{ch}}
\newcommand{\defn}{\coloneqq}
\newcommand{\parall}{\mathsf{parall}}

\begin{document}

\maketitle
\begin{abstract}

Masked diffusion models (MDMs) generate discrete data by iteratively unmasking multiple tokens in parallel using a learned mask predictor.
Their output carries two errors: parallelization error due to parallel sampling, and estimation error from the mask predictor learned with finite samples.
Existing theory treats these errors separately, and the available statistical bounds typically scale with the number $V^L$ of length-$L$ sequences over a vocabulary
of size $V$, suffering from the curse of dimensionality.
In this work, we establish end-to-end statistical guarantees showing that transformer-based MDMs learn structured distributions with a sample complexity governed by their latent structure.
Specifically, we model this structure through mixtures of hidden Markov trees, which capture latent dependencies within components and heterogeneity across components.
When the data distribution is well approximated by such a mixture, a transformer-based MDM with parallel unmasking schedules achieves $\varepsilon$-accurate sampling in KL divergence with $\widetilde{O}(LK_{\mathrm{par}}/\varepsilon)$ training samples, where $K_{\mathrm{par}}$ is polynomial in $L$, $V$, the number of hidden states, and the number of mixture components. The transformer network and sampling procedure require no knowledge of the tree topologies or component parameters, thereby adapting to the unknown latent structure automatically. Overall, our results provide theoretical justification for the empirical effectiveness of MDMs on real-world data.
\end{abstract}

\tableofcontents

\section{Introduction}
\label{sec:introduction}
Discrete diffusion models \citep{austin2021structured,campbell2022continuous} have emerged as a powerful generative modeling paradigm for discrete data.
%
Inspired by continuous diffusion models \citep{ho2020denoising,song2020score}, a discrete diffusion model generates samples from a data distribution $p^\star$ by learning to reverse a categorical corruption process over a discrete state space.
The central object that determines this reverse process is the \textit{concrete score} \citep{meng2022concrete,lou2023discrete}. It encodes ratios between the marginal probabilities of corrupted states, serving as the discrete counterpart of the score function in continuous diffusion models \citep{song2020score,chen2022sampling}.


Discrete diffusion models differ in how the forward process corrupts the data.
Two common corruption mechanisms are uniform replacement and masking.
Uniform diffusion models \citep{austin2021structured,hoogeboom2021argmax} replace each token with one drawn uniformly at random from the vocabulary, while \textit{masked diffusion models} (MDMs) \citep{austin2021structured,campbell2022continuous} replace each token with a special absorbing symbol, referred to as the \textit{mask}.\footnote{We use ``tokens'' to refer to discrete states for convenience; our framework and results apply to any discrete data beyond text.}
Notably, MDMs have demonstrated strong empirical performance in language modeling \citep{sahoo2024simple,shi2024simplified}, code generation \citep{gong2026diffucoder}, and protein sequence modeling \citep{wang2024diffusion}.

Concretely, the forward process of an MDM gradually masks a sequence drawn from $p^\star$ until it is fully masked.
The reverse process starts from a fully masked sequence and, at each step, unmasks a subset of positions.
In particular, for MDMs, the concrete score factorizes into a known time-dependent scalar and a time-independent conditional token probability under $p^\star$ \citep{ou2025your,zheng2025masked}.
Learning the reverse dynamics therefore reduces to learning a \textit{mask predictor} that estimates the conditional distribution of masked tokens given the revealed context.


Ideally, each unmasking step would sample the tokens at the selected positions jointly from their conditional distribution given the revealed context.
Since directly learning these joint conditionals is computationally prohibitive, practical MDMs learn conditional marginals instead and sample from their product \citep{nie2026large}.
A transformer-based mask predictor, trained on finite samples from $p^\star$, can estimate these conditional marginals for all masked positions in a single forward pass.
This enables parallel token generation and breaks the sequential bottleneck of autoregressive models.
However, sampling tokens independently given the same context ignores dependencies among simultaneously generated tokens.
%
The output of an MDM thus carries two errors: a \textit{parallelization error}, which
persists even with exact marginals, and an \textit{estimation error}, because the mask predictor is learned from finite samples.

Despite the empirical success of MDMs, their end-to-end sampling guarantees remain largely unexplored, as existing theory typically treats training and sampling separately.
One line of work takes the mask predictor as given, either exact or with a prescribed error.
 \citet{li2026breaking} derived a sampling error bound for parallel generation with a fixed-cardinality unmasking schedule via information-theoretic measures, and \citet{zhao2026adaptation}, \citet{dmitriev2026efficient} and \citet{wainwright2026data} developed unmasking schedules that exploit intrinsic data complexity.
In addition, \citet{cai2026confidence} established the sampling guarantees of confidence-based unmasking.
However, these works do not characterize the statistical cost of learning the mask predictor.
%

Another line of work studies finite-sample score estimation for discrete diffusion models.
\citet{wakasugi2025state} obtained sampling guarantees independent of the full sequence-state size $V^L$, where $V$ is the vocabulary size and $L$ is the sequence length, under continuous embedding assumptions, but their analysis excludes MDMs.
For MDMs, \citet{zhanggeneralization} and \citet{cho2026minimax} derived sampling guarantees governed by the target support size and the full-sequence size $V^L$, respectively.
Such bounds do not consider the transformer architectures that enable parallel sampling in practice, 
and can remain exponential in the sequence length even when the data admit a compact latent representation, thus suffering from the curse of dimensionality. 
These analyses offer limited insight into the empirical success of transformer-based MDMs on structured data. 

Filling this gap requires an end-to-end analysis that connects the statistical cost of learning a transformer-based mask predictor to the accuracy of parallel generation, with guarantees that capture latent structure in the data.
The dependencies observed in text and biological sequences \citep{chiu2020scaling,frazer2021disease} motivate studying distributions with compact latent representations.
Such distributions can nevertheless assign positive probability to all $V^L$ sequences, so bounds governed by the support size can remain exponential in $L$, obscuring the potential statistical benefits of latent structure.
This leads to our central question:
\begin{center}
\textit{How many training samples does a transformer-based MDM need to sample accurately from $p^\star$, and how does this number depend on the latent structure of the data?}
\end{center}

\subsection{Main contributions}
In this work, we establish end-to-end statistical guarantees for transformer-based MDMs, connecting the finite-sample learning of the mask predictor to the KL accuracy of the generated distribution.

To overcome the curse of dimensionality, we model low-dimensional structure by mixtures of hidden Markov trees, which capture structured latent
dependence within each component and heterogeneity across components. In particular, the data distribution need not belong to this class exactly.

When the data distribution is approximated by such a model and the parallelization error is sufficiently small, we show that a transformer-based MDM achieves $\varepsilon$-accurate sampling in KL divergence with
\begin{align}
    \widetilde{O}(LK_{\rm par}/\varepsilon) \label{eq:sample complexity}
\end{align}
training samples. Here, $L$ is the sequence length and the structural complexity $K_{\rm par}$ is polynomial in the sequence length, vocabulary size $V$, number of hidden states, and number of components, rather than scaling with the size $V^L$ of the sequence space.
Moreover, the transformer architecture for the mask predictor and the parallel sampling procedure require no prior knowledge of the tree topologies or component parameters, so the MDM adapts to the unknown latent structure of the data.

To the best of our knowledge, these are the first end-to-end statistical guarantees for transformer-based MDMs that cover both mask-predictor estimation and parallel unmasking while adapting to latent structure, providing theoretical support for the effectiveness of MDMs on structured data.

\subsection{Related work}
\paragraph{Sample complexity for diffusion models.}
For discrete diffusion models, \citet{wakasugi2025state} analyzed ReLU-based score estimation networks and derived rates governed by smoothness and dimension in a continuous embedding space. 
However, their analysis based on the symmetric-transition structure excludes MDMs.
\citet{zhanggeneralization} analyzed uniform discrete diffusion models and MDMs using empirical kernel-based score estimators and derived sampling guarantees dependent on the support size of the target distribution for MDMs.
\citet{cho2026minimax} established minimax lower bounds for score estimation and nearly matching upper bounds using an MLE-based score estimator, with score estimation error dependent on the full sequence-space size $V^L$. 
\citet{srikanth2025discrete} established KL sampling guarantees when applying neural score estimators trained by stochastic gradient descent (SGD) under regularity conditions. 
For continuous diffusion models, existing work established statistical guarantees for distributions with smooth densities and showed that diffusion models can achieve minimax-optimal sampling \citep{oko2023diffusion,pmlr-v247-wibisono24a,dou2024optimal,zhang2024minimax,cai2025minimax}.
Furthermore, statistical guarantees have been established for learning distributions with low-dimensional structure \citep{pmlr-v202-chen23o,azangulov2024convergence,pmlr-v238-tang24a,fan2025optimal,wu2026diffusion}.

\paragraph{Convergence theory for discrete diffusion models.}
For MDMs, even with exact conditional marginals, parallel unmasking introduces error by sampling conditionally dependent tokens independently. 
\citet{li2026breaking} and \citet{chen2025optimal} characterized this error using information-theoretic measures, 
while \citet{zhao2026adaptation} and \citet{dmitriev2026efficient} developed unmasking schedules that adapt to intrinsic data complexity. 
\citet{lavenant2025error} and \citet{secchi2026schedule} studied schedule optimization through information-theoretic measures, with the latter analyzing the random block sizes induced by $\tau$-leaping.
\citet{cai2026confidence} analyzed entropy-based confidence decoding with exact conditional marginals. 
Beyond MDMs, \citet{dmitriev2026provably} established convergence guarantees governed by dual total correlation for uniform and remasking diffusion, 
while \citet{chen2026parallelism} obtained analogous guarantees for uniform and Gaussian diffusion and compared them with MDMs. 
These analyses assumed exact predictors or prescribed prediction errors without quantifying the finite-sample cost of learning the predictors.


\paragraph{Theoretical analysis of transformers.}
Recent papers study both the approximation capabilities and statistical guarantees of transformers.
\citet{yun2019transformers} established approximation theory for continuous permutation equivariant sequence-to-sequence functions,
while \citet{li2025transformers} developed approximation guarantees for in-context learning over general function classes.
For diffusion transformers (DiTs), \citet{fu2025ditgp} derived score approximation theory and estimation guarantees for Gaussian process data.
\citet{hu2024statistical,hu2025statistical} studied latent and conditional diffusion models, deriving corresponding approximation theory and statistical guarantees.
For next-token prediction, \citet{yuksel2025sample} analyzed generalization error for self-attention predictors, 
and \citet{kong2026generalization} obtained generalization bounds for transformers under an extended log-bilinear language model.

\subsection{Notation}
For $a,b\in\bR$, write $a \vee b \defn \max\{a,b\}$ and $a\wedge b \defn \min\{a,b\}$.
For a positive integer $M$, let $[M]\defn\{1,\cdots,M\}$. 
Let $\ind\{ \cdot \}$ denote the indicator function. 
For $x\in \mathbb{R}^d$, define $\| x \|_p$ as its $p$-norm, i.e., $\|x\|_p \defn (\sum_{i=1}^d |x_i|^p)^{1/p}$, and write $\|x\|_{\infty} \defn \max_{i}|x_i|$. 
We further define $\|x\|_0$ by $\|x\|_0:= \sum_{i\in [d]} \ind\{x_i\neq 0\}$. 
For $W\subseteq [d]$, let $x_{W} \defn (x_i)_{i\in W}$. When used as a predictor input, $x_W$ denotes the length-$L$ sequence retaining $x$ on $W$ and containing masks elsewhere.  
For any matrix $A\in \bR^{m\times n}$, denote its $i$-th row, $j$-th column and Frobenius norm  by $A_{i,\cdot}$, $A_{\cdot,j}$ and $\|A\|_F$, respectively. 
We denote by $\supp(p)$ the support of a probability distribution $p$.
We use $\polylog (\cdot)$ to denote an unspecified polynomial in the logarithm of its arguments.
For positive functions $f,g$, write $f(n)\lesssim g(n)$ or $f(n) = O(g(n))$ if $f(n) \leq Cg(n)$ for some universal constant $C>0$.
The notation $\widetilde{O}$ denotes the respective bounds up to some logarithmic factors.
\section{Preliminaries}
\label{sec:prelim}


\subsection{Masked diffusion models}
\label{sec:mdm}
Let $\bV$ denote the vocabulary set with $|\bV| = V$, and let $L \in \bN_{+}$ be the sequence length.
Denote $\mask \notin \bV$ as the mask symbol, and define the extended vocabulary $\overline{\bV} := \bV \cup \{\mask\}$.

\paragraph{Forward masking process.}
Given a token sequence $X = (X_1,\ldots,X_L) \sim p^\star$ on $\bV^L$, consider a fixed-cardinality masking schedule $s = (s_1,\ldots,s_T)$ satisfying $s_t \in \bN_{+}$ and $\sum_{t=1}^T s_t = L$. 
Initialize $M^0 = \emptyset$. At each step $t\in [T]$, form a random set $S^t$ by selecting $s_t$ positions uniformly from $[L]\setminus M^{t-1}$ 
and then update by $M^t= M^{t-1} \cup S^{t}$. 
The sequence $X$ at time step $t$ is corrupted by replacing the tokens at positions in $M^t$ with $\mask$, while leaving other tokens unchanged.
After step $T$, we have a fully masked sequence $(\mask,\ldots,\mask)$ with $M^T = [L]$.

\paragraph{Mask predictor training.}
For a masked sequence $x \in \overline{\bV}^L$, let $W(x) := \{i\in [L]: x_i\neq \mask\}$ be the index set of unmasked positions.
The target conditional marginal distributions are
\begin{align}
    p_i^\star(v \mid x) := \bP_{X\sim p^\star} \big[X_i=v \mid X_{W(x)}=x_{W(x)}\big],  \quad \forall x\in \overline{\bV}^L,\, v \in \bV, \, i \in [L]\setminus W(x). \label{eq:exact conditional distribution}
\end{align}
To train the mask predictor that approximates these exact conditional distributions, draw $\tau \in [T]$ with $\bP[\tau=t] := s_t/L$, sample $M^{\tau}$ uniformly among subsets of $[L]$ of size $\sum_{t=1}^{\tau} s_t$ and set $W^{\tau} := [L]\setminus M^\tau$. 
We define the per-sequence loss by
\begin{align}
    \ell_{q}(x) := \bE_{\tau, M^\tau} \Big[\frac{1}{|M^\tau|} \sum_{i\in M^\tau} -\log q_{i}(x_i \mid x_{W^{\tau}})\Big], \quad \forall \, x \in \bV^L \label{eq:per-sequence loss},
\end{align}
for any conditional mask predictor $q=\big(q_i\big)_{i=1}^L$. 
Here $x_{W^{\tau}}$ is the corrupted sequence of $x$ that replaces tokens outside $W^{\tau}$ with $\mask$. 
A mask predictor $p_{\theta,i}$ is trained via \citep{shi2024simplified,sahoo2024simple}
\begin{align}
  \min_{\theta}\overline{R}(p_{\theta}) := \bE_{X\sim p^\star} [\ell_{p_{\theta}}(X)] \label{eq:objective func}
\end{align}

Define the excess risk 
\begin{align}
  \mathcal{L}(\theta):= \overline{R}(p_{\theta})-\overline{R}(p^\star), \label{eq:excess risk}
\end{align}
where $\overline R(p^\star)$ denotes the population loss evaluated at the true conditional marginals in \eqref{eq:exact conditional distribution}. 
Minimizing $\mathcal{L}(\theta)$ is equivalent to minimizing $\overline{R}(p_{\theta})$.
As shown in Lemma \ref{lem:kl} in Appendix~\ref{app:kl}, $\mathcal{L}(\theta)$ corresponds to the averaged KL divergence between the true and predicted conditional distributions.
Given $n$ i.i.d.~training samples $\{X^{(j)}\}_{j=1}^n \sim p^\star$, define the empirical risk and empirical risk minimizer (ERM) respectively by
\begin{align}
  \widehat{R}_n(p_{\theta}) := \frac{1}{n} \sum_{j=1}^n \ell_{p_{\theta}}(X^{(j)}), \quad \text{and} \quad \widehat{\theta} \in \arg\min_{\theta} \widehat{R}_n(p_{\theta}). \label{eq:ERM estimator}
\end{align}
%
%
%
Since the masking distribution is prescribed, we keep the expectation over $(\tau,M^\tau)$ in the empirical loss. In practice, it can be evaluated exactly when tractable or approximated by Monte Carlo sampling independent of training data.

\paragraph{Sampling procedure.}
Given a mask predictor $p_{\theta}$ and a fixed-cardinality unmasking schedule $s = (s_1,\ldots,s_T)$, initialize the sequence $Y = (\mask,\dots,\mask)$ and the set of unmasked positions $W^T = \emptyset$.
For iteration $t=T,\ldots, 1$, draw a random unmasking set $S^t$ uniformly among the subsets of $[L] \setminus W^t$ with size $s_t$, set $W^{t-1} = W^{t} \cup S^t$, sample \textit{simultaneously} and \textit{independently}
\begin{align} 
    Y_i  
    \sim p_{\theta,i}(\cdot \mid Y_{W^{t}}) \quad &\text{for}~~i \in S^t,
    \label{eq:reverse process}
\end{align} 
and keep other tokens unchanged. Since $\sum_{t=1}^T s_t = L$, the final iterate $Y\in \mathbb V^L$ is fully unmasked and is the generated sample. We write $p_{Y\mid S}$ for its distribution to emphasize its dependence on the random unmasking sets $S = (S^1,\ldots,S^T)$.

\subsection{Transformer structure for estimation}
\label{sec:estimator}
We specify the transformer structure \citep{vaswani2017attention} used for learning the mask predictor.

\paragraph{Transformer block.} 
Each layer maps $Z\in \bR^{L\times D}$ to $ \bR^{L\times D}$, with each row containing the $D$-dimensional representation. 
A single head is defined by 
\begin{align*}
  \head \big(Z; Q,K,V\big) := \frac{1}{L}\sigma_{\attn} \big((ZQ)(ZK)^{\top}\big) ZV, \quad \text{with } \sigma_{\attn}(x) := \frac{1}{1+e^{-x}}. 
\end{align*}
Here $Q,K,V \in \bR^{D\times D}$ denote query, key and value matrices, respectively; sigmoid activation $\sigma_{\attn}$ is applied entrywise. 
Combining $M$ heads with a residual connection gives:
\begin{align}
    \Attn_{\theta_{\attn}} (Z) := Z + \sum_{m=1}^M \head\big(Z; Q_m,K_m,V_m\big), \quad \text{with }\theta_{\attn}:= \big(Q_m,K_m,V_m\big)_{m\in [M]} . \label{eq:attention layer}
  \end{align} 
The following rowwise feed-forward layer satisfies 
\begin{align*}
  \FFN_{\theta_{\ff}}(Z) := Z + \ReLU \big(ZW\big) U, \quad \text{with } \ReLU(x) := x\vee 0.
\end{align*}
Here ReLU is applied entrywise, and parameters are $\theta_{\ff}:=(W,U)$ with $W \in \bR^{D\times D_{\ff}},U \in \bR^{D_{\ff}\times D}$. 
A transformer block is then defined by:
\begin{align}
     \mathsf{T}_{\theta}(Z) := \FFN_{\theta_{\ff}} \big(\Attn_{\theta_{\attn}} (Z) \big), \quad \text{with }
    \theta:=(\theta_{\mathrm{attn}},\theta_{\mathrm{ff}}), \label{eq:single transformer block}
  \end{align}
  and preserves the $L\times D$ dimensions.

\paragraph{Sparse transformer architecture for mask predictor.}
We build the predictor function class by composing an input embedding, $N$ transformer blocks, an output embedding and a clipped rowwise softmax. 
\begin{itemize}
  \item 
  \textit{Input embedding.} 
  For a masked sequence $x\in \overline{\bV}^L$, let $E \in \bR^{(V+1) \times D}$ be its input embedding matrix, whose 
  rows give the representations of each token $v \in \overline{\bV}$ in $\bR^D$ with $V = |\bV|$.
  Let $P \in \bR^{L\times D}$ be the positional embedding matrix. The input embedding is defined by 
  \begin{align}
    \Embed_{E,P}: \overline{\bV}^L \rightarrow \bR^{L \times D}, \quad \text{with }\big(\Embed_{E,P}(x)\big)_{i,\cdot} := E_{x_i,\cdot} + P_{i,\cdot}, \quad \forall i \in [L]. \label{eq:input embedding}
  \end{align}
  Both matrices are trainable. We control the magnitude of the embedding vectors by some chosen parameter $R_{\embed}>0$, i.e., $\max_{i\in [L],v\in \overline{\bV}}\|E_{v,\cdot}+P_{i,\cdot}\|_2 \leq R_{\embed}$.

  \item \textit{$N$ transformer blocks.}
  For the $l$-th transformer block, collect its parameters as 
  $$\theta_{\tf}^{(l)}:= \big((Q_{l,m},K_{l,m},V_{l,m})_{m=1}^M,W_l,U_l\big),$$ 
  and denote the corresponding block by $ \mathsf{T}^{(l)} := \mathsf{T}_{\theta_{\tf}^{(l)}}$.
  We successively apply $N$ transformer blocks in \eqref{eq:single transformer block}:
  \begin{align*}
    \TF_{\theta_{\tf}}(Z):= \mathsf{T}^{(N)} \circ \cdots \circ \mathsf{T}^{(1)}(Z), \text{ where } \theta_{\tf}:= (\theta_{\tf}^{(1)},\cdots \theta_{\tf}^{(N)}). 
  \end{align*} 
  Each block preserves the output dimension. 

  \item \textit{Output embedding.}
  We map the final hidden representation in $\bR^{L\times D}$ to vocabulary logits $\bR^{L\times V}$ using the output embedding matrix $W_{\rm out}\in \bR^{D \times V}$:
  \begin{align*}
    \OutEmbed_{W_{\rm out}} (Z) := Z W_{\rm out} \in \bR^{L\times V}.
  \end{align*}

  \item \textit{Clipped rowwise softmax.}
  For a fixed chosen $G>0$, we clip each logit entrywise to $[-G,0]$, i.e., $\Clip_{[-G,0]}(x):= \max \{-G,\min\{x,0\}\}$, and apply softmax separately to each row:
  \begin{align}
    \big(\Softmax_{G}(Z)\big)_{i,\cdot} := \softmax \big( \Clip_{[-G,0]}(Z_{i,\cdot})\big), \quad \text{with } \softmax(x) := \bigg(\frac{e^{x_k}}{\sum_{j} e^{x_j}}\bigg)_{k}. \label{eq:rowwise softmax}
  \end{align}

  \item \textit{Parameter and function class.}
  Combining these operations gives our parameterized mask predictor:
  \begin{align}
    \widetilde{\TF}_{\theta}(x) = \Softmax_{G}\big( \OutEmbed_{W_{\rm out}} \big(\TF_{\theta_{\tf}}\big(\Embed_{E,P}(x))\big)\big). \label{eq:transformer predictor} 
  \end{align} 
   Let $\Theta := \Theta(D,N,M,D_{\ff},B,S;R_{\embed},G)$ consist of parameters $\theta := \big(E,P,\theta_{\tf},W_{\rm out}\big)$ satisfying the sizes determined by $D,N,M,D_{\ff}$, the embedding bound $R_{\embed}$ above, 
   with each trainable matrix having Frobenius norm at most $B$ and at most $S$ nonzero parameters in total.
\end{itemize}
The predictor returns a probability distribution at each masked position $i$ in a single forward pass:
\begin{align}
  p_{\theta,i}(\cdot \mid x) := \big(\widetilde{\TF}_{\theta}(x)\big)_{i,\cdot} \label{eq:mask token predictor}
\end{align}
We define its approximation error by $\varepsilon_{\mathsf{approx}}(\Theta):= \min_{\theta\in \Theta} \mathcal{L}(\theta)$.

\section{Problem setup: mixtures of hidden Markov trees}
\label{sec:problem setup}


Without structural assumptions, a distribution on $\bV^L$ has $V^L-1$ parameters, and learning it can require a number of samples that grows with $V^L$.
To exploit low-dimensional structure, we introduce mixtures of hidden Markov trees (MHMTs) as an approximation class for $p^\star$, enabling statistical guarantees governed by the complexity of its latent representation.


A \textit{hidden Markov tree} (HMT) is a latent-variable model that captures dependency structure \citep{crouse1998wavelet}, extending a hidden Markov model (HMM) \citep{baum1966statistical} from a chain to a tree.
An HMM associates each token position $i\in[L]$ with a hidden state $Z_i\in[J]$, where $J$ is the number of hidden states. The states form a Markov chain $Z_1\to \dots \to Z_L$, and, conditioned on these states, the tokens are generated independently, with each $X_i$ depending only on $Z_i$.
An HMT allows the latent states to evolve along a tree instead of a chain, supporting more general dependence structures. And a \emph{mixture of hidden Markov trees} (MHMT) further allows the tree topology and parameters to vary across components.
%


The MHMT class contains mixtures of HMMs, which capture sequential dependence and diversity in language models \citep{ypma2002automatic,chiu2020scaling,zekri2024large}.
In addition, they are widely used to capture biological dependencies in protein-DNA binding sites \citep{barash2003modeling}, gene modules \citep{li2019bayesian}, and heterogeneous cell lineages \citep{mohammadi2022lineage}. 


We now introduce the formal definition of MHMTs, following \citet{meila2000mixtures}.

\begin{definition}[Dependence trees and tree distributions]
    \label{def:tree}
    Let $\mathcal{T}$ be a rooted tree structure on the vertex set $[L]$, namely a connected, acyclic graph with a designated root $o\in [L]$.  
    For each $i\in [L]\setminus \{o\}$, denote $\pa(i)$ as its unique parent, which represents the edge $(i,\pa(i))$. 
    Then the tree structure $\mathcal{T}$ can be specified as
    \begin{align}
        \mathcal{T} := \big\{ \{o\}, \big(i,\pa(i)\big)_{i\in [L] \setminus \{o\}}\big\}. \label{eq:tree structure.}
    \end{align}
    In particular, repeatedly following the parent map from any nonroot vertex reaches $o$ after finite steps.
    We further denote $\mathfrak{T}$ as the set of all trees on $[L]$ and suppress its dependence on $L$.

    For any discrete random vector $Z = (Z_1,\cdots,Z_L)$, we say that $Z$ follows a tree distribution if there exists a dependence tree $\mathcal{T} \in \mathfrak{T}$ such that the law of $Z$ admits the following factorization
    \begin{align}
        p_Z(z_1,\cdots,z_L) = p_o(z_o) \prod_{i\neq o} p_i(z_i \mid z_{\pa(i)}). \label{eq:follow dependence tree}
    \end{align}
    Here the root $o$ and parent map $\pa(i)$ correspond to tree structure $\mathcal{T}$. 
    We further say that $p_Z$ factorizes according to $\mathcal{T}$. 
\end{definition}
\begin{definition}[Mixture of hidden Markov trees]
\label{def:tree-mixtures}
Let $\mathcal{T}$ be a rooted tree on the index set $[L]$, with root $o \in [L]$ and parent map $\pa(\cdot)$ for nonroot vertices as in Definition \ref{def:tree}.
A $J$-state \emph{hidden Markov tree} on $\mathcal{T}$ is a distribution $q$ on $\bV^L$ of the form
\begin{align}
    q(x) = \sum_{z\in[J]^L} q(x,z) \qquad \text{with}\qquad  q(x,z) := \nu(z_o) \prod_{i\neq o} A_i\big(z_i, z_{\pa(i)}\big) \prod_{i=1}^L B_i(x_i, z_i). \label{eq:hidden-tree-law}
\end{align}
Here, $\nu$ is the root-state distribution on $[J]$. For hidden state $b\in[J]$, the transition kernel $A_i(\cdot,b)$ is a distribution on $[J]$ for $i\neq o$, and the emission kernel $B_i(\cdot,b)$ is a distribution on $\bV$ for $i\in[L]$.

The class $\mathcal{M}_{J,R}$ consists of all mixtures of $R$ such models:
\[
    q(x)=\sum_{r=1}^R \pi_r q_r(x)
    \qquad \text{with} \qquad \pi_r\geq0,\quad \sum_{r=1}^R\pi_r=1,
\]
where each $q_r$ is a $J$-state hidden Markov tree with its own topology $\mathcal{T}_r$, root distribution, transition kernels, and emission kernels.
\end{definition}

The MHMT has a simple generative interpretation. 
We first draw a component label $C\sim \pi$. 
Using the parameters of the selected component, we sample the root state $Z_o\sim\nu$ and then traverse the tree from the root, drawing $Z_i\sim A_i(\cdot,b)$ given that its parent is in state $b$. 
Finally, each token is sampled independently as $X_i\sim B_i(\cdot,a)$ given $Z_i=a$.
%
An HMM is recovered when the tree is a chain with $o=1$ and $\pa(i) = i-1$.
More general trees allow direct dependencies between hidden states at nonadjacent sequence positions, providing a flexible representation of dependence among tokens beyond a sequential latent chain. 

\begin{remark}[Why mixtures?]

For fixed component topologies, a distribution in the class $\mathcal{M}_{J,R}$ has $O\big(RLJ(J+V)\big)$ parameters.
This representation complexity can be much smaller than the support size: even a single component with strictly positive emission kernel has full support on $\bV^L$.
Consequently, guarantees that scale with the support size may incur a dependence on $V^L$, even when the distribution admits a compact representation with small $J$ and $R$.
Mixtures enrich this representation by allowing components to differ in tree topology, transitions, and emissions \citep{meila2000mixtures}, while increasing the parameter count only linearly in $R$.
To illustrate this, suppose that all components share a common tree topology.
The mixture can then be represented as a single HMT with $RJ$ states with block diagonal transition matrices.
This structure requires only $O(LRJ^2)$ transition parameters, reduced from $O(LR^2J^2)$ for a generic $RJ$-state HMT with dense transitions.
    
\end{remark}

\paragraph{Approximation of low-dimensional data.}

We focus on the distribution $p^\star$ that is close to the MHMT class $\mathcal{M}_{J,R}$, allowing departures from exact tree-mixture structure.
Misspecification enters through the following approximation error.
\begin{definition}[KL approximation error]
    \label{def:projection error}
    Let $\mathcal{F}$ be a class of probability distributions on $\bV^L$. 
    For any distribution $p$ on $\bV^L$, define the KL approximation error w.r.t.~$\mathcal{F}$ by 
    \begin{align}\label{eq:projection error}
        \Proj_{\mathcal{F}}(p) := \inf_{q\in \mathcal{F}} \KL\big(p\|q\big). 
    \end{align}
    In particular, we write $\Proj_{J,R} := \Proj_{\mathcal{M}_{J,R}}$ for the MHMT class $\mathcal{M}_{J,R}$.
\end{definition}
The quantity $\Proj_{J,R}(p^\star)$ measures how well $p^\star$ can be approximated by a mixture of $R$ $J$-state hidden Markov tree models.
The pair $(J,R)$ indexes the approximation class used in our analysis and is not an input to the training algorithm once the transformer architecture is fixed.
%
Increasing $J$ or $R$ can reduce the approximation error at the cost of greater model complexity.

\section{Main results: statistical guarantees for MDMs}
\label{sec:main}

In this section, we first develop estimation error bounds for the transformer mask predictor learned by ERM, and then establish end-to-end sampling guarantees for the resulting MDM-based sampler.

\subsection{Learning the mask predictor}
\label{sec:estimation-main}

Fix integers $n,L,V\geq 3$ and $J,R\geq 1$. Let $s=(s_1,\cdots,s_T)$ be a fixed-cardinality unmasking schedule with $s_{\mathsf{max}} := \max s_t$. We use the same schedule $s$ for training and inference.

Recall from Section~\ref{sec:prelim} that the ERM $\widehat \theta$ minimizes the empirical masked-token negative log-likelihood in \eqref{eq:ERM estimator}, and the excess risk $\mathcal L(\theta) :=  \overline{R}(p_{\theta}) - \overline{R}(p^\star)$ in \eqref{eq:excess risk}, 
where $\overline{R}(q) = \bE_{X\sim p^\star}[\ell_q(X)]$ and $\ell_q$ is defined in \eqref{eq:per-sequence loss}, equals the average KL divergence between the true and predicted conditionals of masked tokens.

The following theorem controls this estimation error of the transformer mask predictor learned by ERM.
The proof can be found in Section \ref{sec:pf-of-thm1}.
\begin{theorem}[Estimation error of mask predictor]
\label{thm:tree-mixture-estimation}

There exists a transformer class $\Theta$ of the form \eqref{eq:transformer predictor}, with an architecture depending only on $(n,L,V,J,R)$ (specified in Proposition \ref{prop:appendix-parallel-mixture} with $\delta=\varepsilon=1/n$), such that for any distribution $p^\star$ on $\bV^L$, the ERM $\widehat{\theta}$ over $\Theta$ computed from $n$ i.i.d.~samples drawn from $p^\star$ satisfies
\begin{align}
    \bE[\mathcal{L}(\widehat{\theta})] \leq C_{1} \bigg(\frac{s_{\mathsf{max}}}{L} \Proj_{J,R}(p^\star) + \frac{K_{\rm par}}{n}\, \polylog(n,L,V,J,R) \bigg).  \label{eq:estimation error}
\end{align}
Here, $C_{1}>0$ is a universal constant, the expectation is over training samples, and 
\begin{align}\label{eq:Kpar}
K_{\rm par} := R[LJ(J+V)+J^3].
\end{align}

\end{theorem}

The two terms in the error bound \eqref{eq:estimation error} have distinct roles.
The first term captures the approximation error of $p^\star$ by the MHMT class $\mathcal{M}_{J,R}$, and the factor $s_{\mathsf{max}} /L$ converts the joint KL error into a per-token one.
The second term corresponds to the statistical error due to finite-sample training.
The network has a polylogarithmic number of transformer blocks, at most $2R$ attention heads, and $\widetilde O(K_{\rm par})$ nonzero parameters, so the statistical term is a parameter count over the sample size $n$.

When the number of hidden states is moderate in the sense that $J\lesssim \sqrt{LV}$, where $L$ is the sequence length and $V$ is the vocabulary size, $K_{\rm par} = O\big(RLJ(J+V)\big)$ is of the same order as the number of parameters of $\mathcal M_{J,R}$ (see Section~\ref{sec:problem setup}). 
Up to logarithmic factors, the statistical error is thus governed by this intrinsic complexity rather than by the ambient dimension $V^L-1$.
Since the architecture depends only on $(n,L,V,J,R)$, this holds without knowledge of the trees, parameters, or latent states of the approximating mixture. 
The transformer adapts to the unknown latent structure.

\begin{remark}
    Theorem \ref{thm:tree-mixture-estimation} applies to arbitrary tree structures.
    When latent dependencies are organized into only a few hierarchical levels, component trees of bounded height provide a natural model.
    For such structure, Appendix \ref{app:shallow tree} gives a complementary construction for approximation and the resulting estimation and sampling guarantees. 
    Its architectural budgets replace the $J^3$ complexity term with $J^2$, with depth proportional to the tree height.
\end{remark}

\subsection{End-to-end sampling guarantees}
\label{sec:end-to-end result}

Recall from Section~\ref{sec:prelim} a $T$-step MDM sampler with a fixed-cardinality unmasking schedule $s=(s_1,\ldots,s_T)$.
It outputs a sequence $Y$ in $\bV^L$ by iteratively unmasking tokens in the index set $S^t$ in parallel according to a mask predictor $p_\theta$ at step $t$, where $S^t$ with size $|S^t| = s_t$ is a random subset of the currently masked positions.

To isolate the effect of parallel sampling, let $Y^\star$ denote the output of the same MDM sampler when the exact conditional marginals of $p^\star$ are used as the mask predictor. Define the parallelization error 
\begin{align}
\varepsilon_{\mathsf{parall}}(p^\star;s) := \bE_S[\KL(p^\star \,\|\, p_{Y^\star\mid S})] \label{eq:parallelization error},
\end{align}
where the expectation is over the random unmasking sets $S=(S^1,\ldots,S^T)$ with $|S^t| = s_t$.

We now present an end-to-end sampling guarantee for the MDM-based sampling. The proof can be found in Appendix \ref{app:end-to-end}.

\begin{theorem}[End-to-end sampling error of MDM]
\label{thm:end-to-end}
Assume the conditions and adopt the notation of Theorem \ref{thm:tree-mixture-estimation}.
Let $p_{Y\mid S}$ denote the output distribution of the $T$-step MDM sampler using the ERM-based mask predictor $p_{\widehat{\theta}}$ and the realized unmasking sets $\mathcal{S}=(S^1,\ldots,S^T)$ with $|S^t| = s_t$.
Then for any $1\leq T \leq L$, one has 
\begin{align}
    & \bE_{\{X^{(i)}\}_{i=1}^n,S} \big[\KL \big( p^\star \,\|\, p_{Y\mid S}\big)\big] \notag \\
    &\quad \leq C_{2}\bigg( \varepsilon_{\mathsf{parall}}(p^\star;s) + s_{\mathsf{max}} \Proj_{J,R}(p^\star) + \frac{L K_{\rm par}}{n}\, \polylog (n,L,V,J,R)\bigg), \label{eq:sampling error}
\end{align}
where $C_{2}>0$ is a universal constant, and the expectation is over the training samples and random unmasking sets $S$. 
\end{theorem}

To the best of our knowledge, this is the first end-to-end guarantee for transformer-based MDMs that considers both parallelization error due to parallel sampling and finite-sample estimation error of the mask predictor.
Moreover, the guarantee is adaptive to the low-dimensional structure of the data distribution $p^\star$.

The last two terms in the bound \eqref{eq:sampling error} are inherited from the estimation error of the mask predictor in Theorem \ref{thm:tree-mixture-estimation}, multiplied by $L$ to convert the per-token KL error into a joint KL error.
They capture the impact of finite-sample mask predictor training on the end-to-end sampling error.
The first term $\varepsilon_{\mathsf{parall}}(p^\star;s)$ captures the parallelization error due to simultaneous sampling of multiple tokens that ignore their dependencies, which depends only on the data distribution $p^\star$ and the unmasking schedule $s$ used in the inference stage.

\begin{remark}[Parallelization error]
    \label{rem:parallel}
Recent sampling analyses \citep{li2026breaking,wainwright2026data} characterize the parallelization error $\varepsilon_{\mathsf{parall}}(p^\star;s)$ in terms of the information-theoretic structure of the data distribution $p^\star$ and the unmasking schedule $s$.
For instance, as shown in \citet{chen2025optimal}, for any $T=\Omega(\log L)$, there exists a schedule $s^{\mathsf{DTC}}$ depending only on $(L,T)$\footnote{The original construction in \citet{chen2025optimal} requires an estimate of DTC. A straightforward modification yields a schedule that depends only on $(L,T)$ with the same guarantees. For completeness, we present the construction $s^{\DTC}$ in Appendix \ref{app:TC and DTC parallel error}.} such that
    \begin{align}
        \varepsilon_{\mathsf{parall}}(p^\star;s^{\mathsf{DTC}})
        \le \frac{6\log(eL)}{T}\,\mathsf{DTC}(p^\star). \label{eq:TC and DTC control for parallelization error}
    \end{align}
    Here 
    \textit{dual total correlation} (DTC) is defined as
    \begin{align*}
        \DTC(p^\star):= H(X) - \sum_{i=1}^L H(X_i\mid X_{[L]\setminus \{i\}}),
    \end{align*}
    where $X\sim p^\star$ and $H$ denotes the entropy. 
    In particular, DTC can be substantially smaller than the sequence length $L$ for structured data distributions, 
    e.g., an MHMT with star-shaped components satisfies $\DTC(p^\star) \leq \log(RJ)$.
    We refer to \citet{chen2025optimal}, \citet{dmitriev2026efficient} and \citet{zhao2026adaptation} for further discussion of distributions with small $\TC$ and $\DTC$. 
\end{remark}

We now discuss several important implications of Theorem~\ref{thm:end-to-end}.

\begin{itemize}
\item \textbf{Fast and data-efficient sampling for structured data.}
Assume that $p^\star\in\mathcal M_{J,R}$, i.e., the data distribution is exactly a mixture of $R$ $J$-state hidden Markov trees.
Use $s^{\DTC}$ from Remark~\ref{rem:parallel} with $T$ steps. Then Theorem~\ref{thm:end-to-end} yields
\begin{align}
    \bE_{\{X^{(i)}\}_{i=1}^n,S}\big[\KL\big(p^\star\,\|\,p_{Y|S}\big)\big] \lesssim \frac{\log (eL) \DTC(p^\star)}{T} + \frac{LK_{\rm par}}{n} \,  \polylog (n,L,V,J,R), \label{eq:sampling-complexity}
\end{align}
for $T \gtrsim \log L$. 
As a result, when the latent space is small compared to the vocabulary size and length, i.e., $J \lesssim \min\{V,\sqrt{LV}\}$, to generate a sample that is $\varepsilon$-close to $p^\star$ in KL divergence, it suffices to take
\begin{align}
        \text{number of iterations: } T &= \min\Big\{ L, \widetilde{O} \Big( \frac{\DTC(p^\star)}{\varepsilon} \vee \log L\Big) \Big\}, \notag \\
         \text{number of samples: }n &= \widetilde{O}\Big(  \frac{L^2RJV}{\varepsilon} \Big).
\end{align}
This shows that the MDM can achieve sampling acceleration and avoid the curse of dimensionality in sample complexity when the data distribution has low-dimensional latent structure.
For instance, when $p^\star$ is a mixture of $R$ star-shaped HMTs with $J$ hidden states, $\DTC(p^\star) \leq \log(RJ)$, and the number of iterations scales logarithmically in $(R,J)$, up to logarithmic dependence on the sequence length $L$. 
The number of training samples scales polynomially with $(L,V,J,R)$.
This shows that the MDM can efficiently learn and sample from structured data distributions.

\item \textbf{Adaptation to low-complexity structure.}
The architecture in Theorems~\ref{thm:tree-mixture-estimation}--\ref{thm:end-to-end} depends only on $(n,L,V,J,R)$, not on $p^\star$ or on the latent variables, trees, and parameters, none of which are observed.
One network class approximates the masked-token conditionals of every $q\in\mathcal{M}_{J,R}$ uniformly.
In this sense, transformer-based MDMs adapt to the unknown latent structure, with $(J,R)$ acting as budget parameters that trade approximation against estimation.

\item \textbf{Handling non-exact MHMT structure.}
By Theorem~\ref{thm:end-to-end}, for general data distributions $p^\star$ that do not exactly belong to the MHMT class $\mathcal{M}_{J,R}$, to achieve $\varepsilon$-sampling accuracy, the sample complexity polynomially scales with the parameters $(J,R)$ that ensure $\Proj_{J,R}(p^\star) \lesssim \varepsilon/s_{\mathsf{max}}$.

The following two examples show that the required $(J,R)$ can be chosen to scale with low-dimensional structure of the data distribution, rather than the ambient state-space size $V^L$.
The detailed discussions and proofs are deferred to Appendix \ref{app:details of examples}. 


\begin{enumerate}
\item  \emph{Continuous latent representation.}
Suppose that the data distribution satisfies 
$$p^{\star}(x) = \int_{[0,1]^k} \prod_{i=1}^L p_i(x_i \mid z) \mu(\diff z),$$ 
for a latent distribution $\mu$ on $[0,1]^k$, where the kernels $p_i(\cdot \mid z)$ are smooth in $z$ in the sense that for some $\alpha\in (0,1]$,
$\KL\big(p_i(\cdot \mid  z) \,\|\, p_i(\cdot \mid z') \big) \le A^2\| z-z'\|_2^{2\alpha}$
for $i\in[L]$ and $z,z'\in[0,1]^k$. 
This covers  quantized Lipschitz decoders with independent Gaussian observation noise \citep{dmitriev2026efficient}.
Discretizing $\mu$ onto a grid of size $R$ gives a mixture of $R$ product distributions with
$\Proj_{J,R} (p^\star) \le k^{\alpha} L A^2 R^{-2\alpha/k}$ for any $J\geq 1$, so 
$$R \gtrsim (\varepsilon^{-1} k^{\alpha} A^2 s_{\max} L)^{k/(2\alpha)} \Longrightarrow \Proj_{J,R} (p^\star) \leq \varepsilon/s_{\max}.$$ 
So the component count is governed by the latent dimension $k$ and independent of $V$.

\item  \emph{Low R\'enyi entropy.}
For $\alpha\in(0,1)$, the R\'enyi entropy $H_\alpha(p):=\frac{1}{1-\alpha}\log\sum_{x\in\bV^L}p^\alpha(x)$ is small when $p$ concentrates its mass on few points.
In particular, $H_\alpha(p) \uparrow \log|\supp(p)|$ as $\alpha\to 0$.
For any integer $R\geq 2$ and $J\ge 1$, one can show that
$    \Proj_{J,R} (p^\star) \leq L\log V \big({e^{H_{\alpha}(p^\star)}}/{R}\big)^{(1-\alpha)/\alpha},$
that is, $$ R\ge e^{H_{\alpha}(p^\star)} ({\varepsilon^{-1}s_{\max}L\log V})^{\alpha/(1-\alpha)} \Longrightarrow \Proj_{J,R} (p^\star) \le \varepsilon/s_{\max}.$$ 
So the component count is controlled by the R\'enyi entropy and $L\log V$. 
\end{enumerate}
\end{itemize}
\section{Analysis}
\label{sec:proof sketch}
In this section, we prove Theorem \ref{thm:tree-mixture-estimation}.
%
Recall the definition of $\mathcal{L}(\theta) = \overline{R}(p_{\theta}) - \overline{R}(p^\star)$ in \eqref{eq:excess risk}, where $\overline{R}(q) := \bE_{X\sim p^\star}[\ell_{q}(X)]$ and $\ell_{q}$ is defined in \eqref{eq:per-sequence loss}.
The estimation error can be decomposed as
\begin{align}
 \mathcal L(\widehat\theta)
    &=
    \underbrace{
        \overline{R}(p_{\widehat\theta})-\inf_{\theta\in\Theta}\overline{R}(p_{\theta})
    }_{\text{statistical error}}
    +
    \underbrace{
        \inf_{\theta\in\Theta}\overline{R}(p_{\theta})-\overline{R}(p^\star).
    }_{\text{approximation error:
      }\varepsilon_{\mathsf{approx}}(\Theta)}
    \label{eq:estimation-error-decomposition}
\end{align}
We control these two terms as follows.
\begin{itemize}
    \item 
    \textit{Statistical error.}
    The statistical error is governed by the properties of the loss and the capacity of the mask predictor class.
    Using localization techniques \citep{bartlett2005local,koltchinskii2006local}, the following proposition bounds the total excess risk by the approximation error and a statistical term determined by the capacity of the predictor class.

    \begin{proposition}[Statistical error]
        \label{prop:gen error}
        Assume $n,V\ge 3$. 
        Let $\Theta = \Theta (D,N,M,D_{\ff},B,S;R_{\embed},G)$ be the parameter class in Section~\ref{sec:estimator} with $D,N,M,D_{\ff} \ge 1$, and $B,R_{\embed},G>0$.
        Define 
        \begin{align*}
            P_{\mathrm{tr}} :=(V+1)D+LD+N(3MD^2+2DD_{\mathrm{ff}})+DV = O(DV+LD+NMD^2+NDD_{\ff}),
        \end{align*}
        as the number of parameters in this function class and suppose the sparsity $1\le S\le P_{\mathrm{tr}}$ is an integer.
        There exists a constant $A_{\mathrm{TF}}\geq e$, depending only on $N$, $M$, $B$, $R_{\embed}$, satisfying
        \[
        \log A_{\mathrm{TF}}
        \leq CN^2\log\big(eNM(B\vee1)(R_{\mathrm{embed}}\vee1)\big),
        \]
        such that
        \begin{align}
            \bE[\mathcal{L}(\widehat{\theta})] \lesssim \varepsilon_{\mathsf{approx}}(\Theta) + (G+\log V) \cdot \frac{C(\Theta)\log (en) + 1 }{n}.
        \end{align}
        Here the capacity of $\Theta$ satisfies
        \begin{align*}
            C(\Theta):=S\bigg[\log\frac{eP_{\mathrm{tr}}}{S}
            +\log A_{\mathrm{TF}}\bigg],
        \end{align*}
        $\widehat{\theta}$ is the ERM estimator in \eqref{eq:ERM estimator} and the expectation is taken over training data $\{X^{(j)}\}_{j=1}^n$. 
    \end{proposition}
    Its proof can be found in Appendix~\ref{app:proof of prop: gen err}.

    \item  \textit{Approximation error.}
    We next control the approximation error by explicitly constructing sparse transformer predictors.
    \begin{proposition}[Tree-mixture approximation]
    \label{prop:tree-mixture-approximation}
    Fix integers $L,V\geq2$, $J,R\geq1$,
    $0<\delta,\varepsilon\leq1/2$,  and a masking schedule $s=(s_1,\cdots,s_T)$ with $s_{\max}:= \max_{t\in [T]} s_t$, as in Section \ref{sec:prelim}. 
    There exists a sparse transformer class $\Theta$, with architectural budgets
    depending only on $(L,V,J,R,\delta,\varepsilon)$, such that
    \begin{align}
        \varepsilon_{\mathsf{approx}}(\Theta)
        \leq
        \frac{s_{\max}}{L}\Proj_{J,R}(p^\star)
        +s_{\max}\log\frac{1}{1-\delta}
        +2\varepsilon
        \label{eq:tree-mixture-approximation}
    \end{align}
    for every distribution $p^\star$ on $\bV^L$.
    \end{proposition}
    Its architectural budgets are given in Proposition \ref{prop:appendix-parallel-mixture}.
    The proof of Proposition~\ref{prop:tree-mixture-approximation} is given in Appendix~\ref{app:pf-of-prop:tree-mixture-approx}.
    The key idea is to first construct a transformer-based mask predictor for a single hidden Markov tree structure, where the contraction idea \citep{miller1985contraction} enables us to build it within $\widetilde{O}(\log L)$ transformer blocks.
    Then we extend it to mixtures by using the posterior mixture weights computed from the context. 

\end{itemize}

\subsection{Proof of Theorem \ref{thm:tree-mixture-estimation}}
\label{sec:pf-of-thm1}
Now we are ready to prove Theorem \ref{thm:tree-mixture-estimation}.
Take the parameter class $\Theta$ in Proposition \ref{prop:appendix-parallel-mixture} with $\delta=\varepsilon=\frac1n$, $R_{\embed} = \sqrt{3}$ and $G = G_{\delta}=\log(nV)$. 
Let 
\[
 \beta:=\log(enLVJR),\qquad
 K_{\mathrm{par}}:=R\big[L(J^2+JV)+J^3\big].
\]
The architectural bounds in Proposition \ref{prop:appendix-parallel-mixture} imply
\begin{align*}
    N\le C\beta^3,\qquad M\le2R,&\qquad
 S\le CK_{\mathrm{par}}\beta^{10},\qquad \log B\le C\beta^2 \\
 D \le C[L+V+R(J^3+JV)\beta^3], &\qquad D_{\ff} \le C[LV+R(J^3+JV)\beta^3] 
\end{align*}
Recall from Section \ref{sec:estimator} that $N$ is the number of transformer blocks, $M$ is the number of attention heads, 
and $D,D_{\ff}$ are the embedding dimension and feed-forward hidden width, respectively. 
The sparsity budget $S$ bounds the total number of nonzero parameters,
while $B$ bounds the Frobenius norm of each parameter matrix. 
Finally, $R_{\embed}$ bounds the row norms of the embedded inputs,
and $G$ determines the logit clipping interval $[-G,0]$.

\begin{itemize}
\item  \textit{Approximation error.}
By Proposition~\ref{prop:tree-mixture-approximation}, 
\begin{align}
 \varepsilon_{\mathsf{approx}}(\Theta)
 &\le\frac{s_{\max}}L\Proj_{J,R}(p^\star)
      +s_{\max}\log\frac1{1-1/n}+\frac2n\notag\\
 &\le\frac{s_{\max}}L\Proj_{J,R}(p^\star)
      +\frac{2s_{\max}+2}{n}. \label{eq:estimation-proof-approximation}
\end{align}
Here we use $-\log(1-x)\le2x$ for $0\le x\le1/2$. 

\item  \textit{Capacity and estimation error.}
Recall the definition of $P_{\rm tr}$ and $A_{\TF}$ in Proposition \ref{prop:gen error}.
The architectural bounds above give
\begin{align*}
    \log P_{\rm tr}&=\log \big((V+1)D+LD+N(3MD^2+2DD_{\rm ff})+DV\big) \lesssim \beta, \\
    \log A_{\TF} &\le CN^2\log\big(eNM(B\vee1)(R_{\embed}\vee1)\big) \le C\beta^8.
\end{align*}
The capacity of $\Theta$ is therefore controlled by
\begin{align}
 C(\Theta)
  = S\bigg[\log\frac{eP_{\rm tr}}{S}+\log A_{\TF}\bigg]
 \le CK_{\mathrm{par}}\beta^{18}.
 \label{eq:estimation-proof-capacity}
\end{align}

Applying Proposition \ref{prop:gen error} then yields
\begin{align*}
 \bE\big[\loss(\widehat\theta)\big]
 &\lesssim \varepsilon_{\mathsf{approx}}(\Theta)
       + (\log nV + \log V)\frac{C(\Theta)\log(en)+1}{n}\\
 &\lesssim \frac{s_{\max}}L\Proj_{J,R}(p^\star)
       +\frac{K_{\mathrm{par}}\beta^{20}}{n}.
\end{align*}
The expectation is over the i.i.d. training data $\{X^{(i)}\}_{i=1}^n \sim p^\star$. 
In the last inequality, we use \eqref{eq:estimation-proof-approximation}--\eqref{eq:estimation-proof-capacity}, $\log(en)\le\beta$, and $s_{\max}\le L\le K_{\mathrm{par}}$. 
This proves Theorem \ref{thm:tree-mixture-estimation}.

\end{itemize}
\section{Discussion}
This paper studies the sample complexity of masked diffusion models (MDMs) for learning distributions with low-complexity structure, measured by their approximation through mixtures of hidden Markov tree models.
We establish end-to-end sampling guarantees whose statistical error depends on the structural complexity $\mathrm{poly}(L,V,J,R)$, rather than the size of the sequence space $V^L$.
Using the prescribed architectural parameters, our transformer predictor adapts to the unknown latent structure through the training objective on observed data.  
These results help explain why transformer-based MDMs perform well in learning distributions with structural dependence.

Several directions remain open.
First, extending the analysis beyond mixtures of hidden Markov trees would accommodate more general dependence structures, including latent graphic models with cycles. 
A key challenge is to identify structures that capture dependencies arising in practice while admitting efficient approximation of conditional marginals by transformers. 
Second, confidence-based unmasking has shown strong empirical performance in diffusion language models, improving generation quality and enabling efficient parallel decoding \citep{wu2026fast}.
Establishing statistical guarantees for such decoding strategies would clarify how predictor estimation errors affect adaptive token selection and the resulting sampling accuracy.
Such an analysis must account for both the effect of prediction errors on the unmasking decisions and the mismatch between the masking patterns used during training and those encountered during adaptive sampling.
Third, incorporating optimization error would further help bridge the gap between our theoretical guarantees and practical training. 
Our analysis assumes access to the ERM estimator, whereas practical training performs finitely many parameter updates using algorithms such as AdamW \citep{loshchilov2017decoupled} and Muon \citep{liu2025muon}. 
Extending the guarantees to approximately optimized predictors would clarify how optimization accuracy affects the final sampling error and inform the allocation of computational resources between training and sampling. 

\section*{Acknowledgements}
C.\ Cai is supported by the NSF CAREER Award CCF-2541600 and the NSF grant DMS-2515333.

\bibliographystyle{apalike}
\bibliography{references}

\appendix
\section{Examples of low-complexity latent structure}
\label{app:details of examples}
In this appendix, we derive the bounds on $\Proj_{J,R}(p^\star)$ for the continuous latent representation and low R\'enyi entropy examples in Section \ref{sec:end-to-end result}.

\subsection{Continuous latent representation}
\begin{lemma}
      \label{lem:continuous latent representation}
      Suppose that there exists a probability distribution $\mu$ on $[0,1]^k$ such that, 
      \begin{align}
            p^{\star}(x) = \int_{[0,1]^k} \prod_{i=1}^L p_i(x_i \mid z) \mu(\diff z), \quad \forall \, x\in \bV^L  \label{eq:continuous latent repre}
      \end{align}
      We further assume that for some $A\ge 0$ and $\alpha\in (0,1]$, the kernel admits the following continuity condition:
      \begin{align}
            D_{\mathrm{KL}}\big(p_i(\cdot \mid z) \,\|\, p_i(\cdot \mid z')\big) \leq A^2 \|z-z'\|_2^{2\alpha}, \quad \forall i \in [L],\quad \forall z,z' \in [0,1]^k.  \label{eq:continuour latent repre KL condition}
      \end{align}
      Then for any integers $J,R\geq 1$, it holds that, 
      \begin{align}
            \Proj_{J,R}(p^\star)\leq \Proj_{1,R}(p^\star) \leq C_{k,\alpha} L A^2 R^{-2\alpha/k} \label{eq:cont latent repre proj error}.
      \end{align}
      Here constant $C_{k,\alpha}$ depends only on $k,\alpha$. In particular, condition \eqref{eq:continuour latent repre KL condition} holds if
      $p_i(\cdot \mid z)=\softmax(g_i(z))$ with logits
            $g_i:[0,1]^k \to \bR^V$ satisfying
      \begin{align}
      \|g_i(z)-g_i(z')\|_2 \leq \sqrt{2} A\|z-z'\|_2^\alpha, \quad \forall i\in [L], \quad z,z' \in [0,1]^k \label{eq:continuous latent repre sufficient for KL}
      \end{align}
\end{lemma}

\paragraph{Proof of Lemma \ref{lem:continuous latent representation}.}

We approximate the latent variable by a finite grid. Let $s=\lfloor R^{1/k}\rfloor\geq1$, and partition $[0,1]^k$ into $s^k\leq R$ cubes $\{C_j\}_{j=1}^{s^k}$ of side length $1/s$, assigning their boundaries so that the cells are disjoint. Denote by $z_j$ the midpoint of $C_j$, and define
\[
    w_j:=\mu(C_j),\qquad
    q_R(x):=\sum_{j=1}^{s^k}w_j\prod_{i=1}^L p_i(x_i\mid z_j).
\]
Since each component is a product distribution, $q_R\in\mathcal M_{1,R}$. For $z\in[0,1]^k$, let $j(z)$ be the index of its cell. Consider the joint distributions
\[
    P(\diff z,x):=\mu(\diff z)\prod_{i=1}^L p_i(x_i\mid z),
    \qquad
    Q(\diff z,x):=\mu(\diff z)\prod_{i=1}^L p_i(x_i\mid z_{j(z)}).
\]
Their marginals on $x$ are $p^\star$ and $q_R$, respectively:
\begin{align*}
    \int_{[0,1]^k} Q(\diff z,x) = \int_{[0,1]^k} \prod_{i=1}^L p_i(x_i\mid z_{j(z)})\,\mu(\diff z) = \sum_{j=1}^{s^k} \int_{C_j} \prod_{i=1}^L p_i(x_i\mid z_{j})\,\mu(\diff z) = q_R(x).
\end{align*}
By the data-processing inequality and the chain rule for KL divergence,
\begin{align*}
    D_{\mathrm{KL}}(p^\star\, \| \, q_R)
    &\leq D_{\mathrm{KL}}(P\, \|\, Q)\\
    &=\int_{[0,1]^k}\sum_{i=1}^L
      D_{\mathrm{KL}}\big(p_i(\cdot\mid z)\,\|\,p_i(\cdot\mid z_{j(z)})\big)\,\mu(\diff z)\\
    &\leq LA^2\int_{[0,1]^k}\|z-z_{j(z)}\|_2^{2\alpha}\,\mu(\diff z)\\
    &\leq LA^2\bigg(\frac{\sqrt{k}}{2s}\bigg)^{2\alpha}
     \leq k^\alpha LA^2R^{-2\alpha/k}.
\end{align*}
Here the last two inequalities control the maximum distance to the cell midpoint and $s\geq R^{1/k}/2$. Together with $\mathcal M_{1,R}\subseteq\mathcal M_{J,R}$, this proves \eqref{eq:cont latent repre proj error} with $C_{k,\alpha}=k^\alpha$.

It remains to verify the sufficient condition \eqref{eq:continuous latent repre sufficient for KL}. Let $F(a):=\log\sum_{v=1}^V e^{a_v}$. For $u=\operatorname{softmax}(a)$, we have
\[
    \nabla F(a)=u,\qquad
    \nabla^2F(a)=\operatorname{diag}(u)-uu^\top,
    \qquad \|\nabla^2F(a)\|_{\mathrm{op}}\leq1.
\]
For any $a,b\in\mathbb R^V$,
\begin{align*}
    D_{\mathrm{KL}}\big(\softmax(a)\,\|\,\operatorname{softmax}(b)\big)
    &= \sum_{v=1}^V \big(\softmax(a)_v\log\softmax(a)_v - \softmax(a)_v\log\softmax(b)_v\big)\\
    &=F(b)-F(a)-\nabla F(a)^\top(b-a)\\
    &\leq\frac12\|b-a\|_2^2.
\end{align*}
Taking $a=g_i(z)$ and $b=g_i(z')$, and applying \eqref{eq:continuous latent repre sufficient for KL} gives
\[
    D_{\mathrm{KL}}\big(p_i(\cdot\mid z)\,\|\,p_i(\cdot\mid z')\big)
    \leq\frac12\|g_i(z)-g_i(z')\|_2^2
    \leq A^2\|z-z'\|_2^{2\alpha}.
\]
This implies \eqref{eq:continuour latent repre KL condition} and completes the proof.

\subsection{Low R\'enyi entropy}
For a distribution $p$ on $\mathbb V^L$ and $\alpha\in(0,1)$, recall that its R\'enyi entropy is defined by
\[
    H_\alpha(p):=\frac{1}{1-\alpha}\log\sum_{x\in\mathbb V^L}p(x)^\alpha.
\]
\begin{lemma}[Approximation under low R\'enyi entropy]
\label{lem:renyi-compression}
Let $p^\star$ be a distribution on $\mathbb V^L$ and $\alpha\in(0,1)$. For any integers $J\geq1$ and $R\geq2$, it holds that
\begin{align}
    \Proj_{J,R}(p^\star)\leq\Proj_{1,R}(p^\star)
    \leq L\log V\min\bigg\{1,
    \bigg(\frac{e^{H_\alpha(p^\star)}}{R}\bigg)^{(1-\alpha)/\alpha}\bigg\}.
    \label{eq:renyi-error-bound}
\end{align}
\end{lemma}

\paragraph{Proof of Lemma \ref{lem:renyi-compression}.}
We maintain the $R-1$ largest atoms and approximate the remaining mass by a uniform distribution. Enumerate the points of $\mathbb V^L$ as $x_{(1)},\ldots,x_{(V^L)}$, with probabilities $p_{(j)}:=p^\star(x_{(j)})$ in decreasing order. Set $p_{(j)}=0$ for $j>V^L$, and let
\[
    S_\alpha:=\sum_{j\geq1}p_{(j)}^\alpha
    =e^{(1-\alpha)H_\alpha(p^\star)},\quad
    \tau_{R-1}:=\sum_{j\geq R}p_{(j)}.
\]
If $\tau_{R-1}=0$, then $p^\star$ is a mixture of at most $R-1$ point masses. Each point mass is a product distribution, so $\Proj_{1,R}(p^\star)=0$.

For $\tau_{R-1}>0$, let $u$ be the uniform distribution on $\mathbb V^L$, and define
\[
    q_R:=\sum_{j=1}^{R-1}p_{(j)}\delta_{x_{(j)}}+\tau_{R-1}u,
    \qquad
    v:=\frac{1}{\tau_{R-1}}\sum_{j=R}^{V^L}p_{(j)}\delta_{x_{(j)}}.
\]
Since Dirac measures and the uniform distribution are product distributions, $q_R\in\mathcal M_{1,R}$.
Furthermore, $p^\star=\sum_{j=1}^{R-1}p_{(j)}\delta_{x_{(j)}}+\tau_{R-1}v$. 
By the convexity of KL divergence,
\begin{align*}
    \Proj_{1,R}(p^\star)
    &\leq D_{\mathrm{KL}}(p^\star\|q_R)
     \leq\tau_{R-1}D_{\mathrm{KL}}(v\|u)\\
    &=\tau_{R-1}\big(L\log V-H(v)\big)
     \leq\tau_{R-1}L\log V.
\end{align*}
To bound the remaining mass $\tau_{R-1}$, notice that
\[
    \tau_{R-1}
    =\sum_{j\geq R}p_{(j)}^\alpha p_{(j)}^{1-\alpha}
    \leq p_{(R)}^{1-\alpha}S_\alpha,
    \quad \text{and }
    R\, p_{(R)}^\alpha\leq\sum_{j=1}^R p_{(j)}^\alpha\leq S_\alpha.
\]
Therefore,
\[
    \tau_{R-1}\leq
    \bigg(\frac{S_\alpha}{R}\bigg)^{(1-\alpha)/\alpha}S_\alpha
    =\bigg(\frac{e^{H_\alpha(p^\star)}}{R}\bigg)^{(1-\alpha)/\alpha}.
\]
Combining this bound with $\tau_{R-1}\leq1$ and $\mathcal M_{1,R}\subseteq\mathcal M_{J,R}$ proves \eqref{eq:renyi-error-bound}.


\section{Proof of Theorem~\ref{thm:end-to-end}}
\label{app:end-to-end}
In this section, we decompose the sampling error into parallelization error and conditional prediction error following \citet{li2026breaking}, and then apply Theorem~\ref{thm:tree-mixture-estimation} to derive an end-to-end bound.

Fix the training samples and write $\theta=\widehat\theta$. 
Throughout this proof, $p_{Y\mid S}$ denotes the sampler distribution conditional on both the training sample and $S$, with the dependence on the training sample suppressed.
Let $p_{Y^\star \mid S}$ denote the analogous sampler distribution using the exact conditional marginals $p_i^\star$.
Recall the unmasking partition $S=(S^1,\ldots,S^T)$ and the unmasked index sets $W^t$ in Section \ref{sec:prelim}. 
Then it holds that
\[
    p_{Y\mid S}(x)
    :=\prod_{t=1}^T\prod_{i\in S^t}p_{\theta,i}(x_i\mid x_{W^t}).
\]
For a fixed context set $W\subseteq[L]$ and $i\notin W$, write
\[
    \Delta_{\theta,i}(W)
    :=\bE_{X\sim p^\star}\Big[
    \KL(p_i^\star(\cdot\mid X_W)
    \,\|\,p_{\theta,i}(\cdot\mid X_W)\big)\Big].
\]

By the KL chain rule, we have
\begin{align*}
    \KL(p^\star\|p_{Y\mid S})
    &= \bE_{X\sim p^\star} \Big[\log \frac{\prod_{t=1}^T p^\star(X_{S^t}\mid X_{W^t})}{\prod_{t=1}^T \prod_{i\in S^t} p_{\theta,i}(X_i\mid X_{W^t})}\Big]\\
    & = \KL(p^\star\|p_{Y^\star\mid S}) + \bE_{X\sim p^\star}\Big[\sum_{t=1}^T\sum_{i\in S^t} \log \frac{p_i^\star(X_i \mid X_{W^t})}{p_{\theta,i}(X_i \mid X_{W^t})}\Big]\\
    &=\KL(p^\star\|p_{Y^\star\mid S})
    +\sum_{t=1}^T\sum_{i\in S^t}\Delta_{\theta,i}(W^t).
\end{align*}
Averaging this first term over $S$ gives $\varepsilon_{\parall}(p^\star;s) $. 
For the second term, let $M^t:=[L]\setminus W^t$ and $m_t:=\sum_{u=1}^t s_u$. 
Then $M^t$ is uniform among subsets of $[L]$ of size $m_t$, and, conditioned on $M^t$, the set $S^t$ is a uniform $s_t$-subset of $M^t$. Hence,
\begin{align*}
    \bE_{ S}\bigg[\sum_{t=1}^T\sum_{i\in S^t}
    \Delta_{\theta,i}(W^t)\bigg]
    &=\sum_{t=1}^T\frac{s_t}{m_t}\,
    \bE_{M^t}\bigg[\sum_{i\in M^t}
    \Delta_{\theta,i}([L]\setminus M^t)\bigg]\\
    &=L\mathcal L(\theta).
\end{align*}
The last equality uses the conditional-KL representation of the excess risk $\mathcal L$ in Lemma~\ref{lem:kl}, with training time distribution $\mathbb P[\tau=t]=s_t/L$. We therefore obtain the decomposition
\begin{align}
    \bE_{S}\big[\KL(p^\star\|p_{Y\mid S})\big]
    =\varepsilon_{\parall}(p^\star; s)
    +L\mathcal L(\theta).
    \label{eq:sampling-decomposition}
\end{align}

Finally, taking expectation over the training sample and
applying Theorem~\ref{thm:tree-mixture-estimation} gives
\begin{align*}
    \bE_{\{X^{(j)}\}, S}\big[
        \KL(p^\star\|p_{Y\mid S})
    \big]
    &=\varepsilon_{\parall}(p^\star; s)
      +L\bE_{ \{X^{(j)}\}}\big[\mathcal L(\widehat\theta)\big]\\
    &\lesssim
      \varepsilon_{\parall}(p^\star; s)
      +s_{\max}\Proj_{J,R}(p^\star)
      +\frac{LK_{\mathrm{par}}}{n} \polylog (n,L,V,J,R).
\end{align*}
This proves Theorem \ref{thm:end-to-end}.


\section{Generalization analysis}
In this appendix, we prove Proposition~\ref{prop:gen error} by combining a localized oracle inequality with a covering-number bound for the transformer logit class. 
We first establish the localized oracle inequality for a general class of bounded logits under the masking loss from Section \ref{sec:prelim}.
We then control the capacity of the transformer class and prove Proposition~\ref{prop:gen error}.

To state the result beyond the transformer setting, we consider a general logit class $\mathcal{F}(\Theta)$ of functions from $\overline{\bV}^L$ to $\bR^{L\times \bV} $, parametrized by $\theta \in \Theta$ and define its induced mask predictor class by:
    \begin{align}
        \mathcal{P}_{\mathcal{F}}(\Theta) := \Big\{ p_{\theta}\,  \big|\,  p_{\theta,i,v} (x):= p_{\theta,i}(v\mid x) = \frac{\exp (F_{\theta,i,v}(x))}{\sum_{w\in \bV} \exp (F_{\theta,i,w}(x))}, \quad \forall i\in [L], \, v \in \bV, \, x\in \overline{\bV}^L, \, F_{\theta} \in \mathcal{F}(\Theta)\Big\}. \label{eq:general mask predictor class}
    \end{align}
    Here, softmax is applied separately at each position $i \in [L]$. The empirical and population risks defined in Section \ref{sec:prelim} are evaluated at $p_{\theta} \in \mathcal{P}_{\mathcal{F}}(\Theta)$. 
    To characterize the capacity of $\mathcal{F}(\Theta)$, we introduce the following metric.
    \begin{align}
        d_\infty(F,F'):=\sup_{x\in\overline{\bV}^L}\max_{i\in [L],v\in \bV}|F_{i,v}(x)-F'_{i,v}(x)|, \quad \forall \, F,F': \overline{\bV}^L \rightarrow \bR^{L\times \bV} \label{eq:infinity-norm difference}
    \end{align}
    Let $\mathcal N(\eta,\F,d_\infty)$ denote the minimum number of radius-$\eta$ balls needed to cover function class $\F$ under metric $d_{\infty}$. 
    We impose the following boundedness and capacity conditions on the logit class to control the generalization error. 
    \begin{assumption}[Logit class]
    \label{ass:class}
    Let $\mathcal{F}(\Theta)$ be a logit function class on $\overline{\bV}^L$. 
    Suppose that there exist constants $G\ge 0$ and $C(\Theta)\ge 1$ such that 
    \begin{enumerate}[label=\textup{(\roman*)}]
        \item Boundedness. For every $\theta \in \Theta$ and any position $i\in [L]$, token $v\in \bV$, it holds that 
        $-G\le F_{\theta,i,v}(x)\le0$ for any $F_{\theta} \in \mathcal{F}(\Theta)$ and any input $x\in \overline{\bV}^L$. 
        \item Capacity. For any $\eta\in (0,1]$, the covering number satisfies  
        \[
        \log\mathcal N(\eta,\F(\Theta),d_\infty)\le C(\Theta)\log(e/\eta).
        \]
    \end{enumerate}
    \end{assumption}
    The boundedness condition implies that $p_{\theta} \in \mathcal{P}_{\mathcal{F}}(\Theta)$ satisfies $p_{\theta,i,v} \ge e^{-G}/V$, so its tokenwise negative log-likelihood is bounded by $M_G := G + \log V$. 
    Together with the capacity bound, this allows us to apply the localization techniques \citep{bartlett2005local,koltchinskii2006local} to obtain the following statistical guarantee. 

    \begin{proposition}[Localized oracle inequality for a logit class]
    \label{prop:oracle}
    Fix integers $n,V\ge3$, and a masking schedule $s$ in Section~\ref{sec:mdm}. 
    Suppose that $\mathcal{F}(\Theta)$ satisfies Assumption~\ref{ass:class} with parameters $G$ and $C(\Theta)$, 
    and let $\widehat{\theta} \in \arg \min_{\theta \in \Theta} \widehat{R}_n(p_{\theta})$ be an exact ERM estimator over the induced predictor class $\mathcal{P}_{\mathcal{F}}(\Theta)$.
    Here the population excess risk is defined as $\mathcal{L}(\theta) := \overline{R}(p_{\theta}) - \overline{R}(p^\star)$ for $p_{\theta} \in \mathcal{P}_{\mathcal{F}}(\Theta)$.
    Let $M_G:=G+\log V$. 
    Then
    \begin{equation}
    \bE\big[\loss(\widehat\theta)\big]
    \lesssim  \inf_{\theta \in \Theta} \mathcal{L} (\theta)
        +M_G \cdot \frac{C(\Theta)\log(en)+1}{n}.
    \label{eq:prop oracle}
    \end{equation}
    Here the expectation is taken over training data $\{X^{(j)}\}_{j=1}^n$. 
    \end{proposition}

\subsection{Proof of Proposition~\ref{prop:oracle}}
Throughout the proof, the masking schedule $s$ is fixed, and $C$ denotes a universal constant that may change from line to line.
We write $\varepsilon_{\mathsf{approx}}(\Theta):= \inf_{\theta\in\Theta} \mathcal{L}(\theta)$ for the approximation error of the predictor class $\mathcal{P}_{\mathcal{F}}(\Theta)$.

For a function $g$ of a token sequence, write
\[
    Pg:=\bE_{X\sim p^\star}[g(X)],
    \qquad
    P_ng:=\frac1n\sum_{j=1}^n g(X^{(j)}).
\]
Recall that $\ell_{p_\theta}$ already averages over the masking randomness. Hence, $P_n$ averages over the $n$ independent training sequences. For $\zeta>0$, choose $\theta^\circ\in\Theta$ such that
\[
    a:=\mathcal L(\theta^\circ)
    \leq\varepsilon_{\mathsf{approx}}(\Theta)+\zeta,
    \qquad
    h_\theta:=\ell_{p_{\theta}}-\ell_{p_{\theta^\circ}}.
\]
By the definition of $h_{\theta}$ and the ERM property,
\[
    Ph_\theta=\mathcal L(\theta)-a,
    \qquad P_nh_{\widehat\theta}\leq0.
\]
We first bound the second moment of $h_\theta$ and then control its empirical deviation.

\paragraph{Second-moment bound.}
Let $M_G=G+\log V$ and $\rho=e^{-M_G}$. The bounded logits imply $p_{\theta,i}(v\mid x)\geq\rho$ for every masked position and token. We claim that, for any probability vectors $p,q$ on $[V]$ with $q_v\geq\rho$,
\begin{align}
    \sum_{v:p_v>0}p_v\log^2\frac{p_v}{q_v}
    \leq (M_G+2)D_{\mathrm{KL}}(p\|q)
    \leq 3M_GD_{\mathrm{KL}}(p\|q).
    \label{eq:appendix-logloss-curvature}
\end{align}
Here the last inequality uses $M_G\geq\log3>1$.
To prove the first inequality, notice that
\[
    t\log^2t\leq(M_G+2)(t\log t-t+1),
    \qquad \forall 0\leq t\leq e^{M_G}.
\]
Indeed, for $t=e^{-x}\leq1$, the inequality $e^x\geq1+x+x^2/2$ gives
\[
    e^{-x}x^2\leq2\cdot \big(1-(x+1)e^{-x}\big).
\]
For $t=e^x\geq1$, with $0\leq x\leq M_G$, we have
\[
    (x+2)(x-1+e^{-x})-x^2
    =x-2+(x+2)e^{-x}\geq0,
\]
since the last expression is zero at $x=0$ and has derivative $1-(x+1)e^{-x}\geq0$. The case $t=0$ follows by continuity. Taking $t=p_v/q_v$, multiplying by $q_v$, and summing over $v$ proves \eqref{eq:appendix-logloss-curvature}.

Since $0\leq\ell_{p_{\theta}}\leq M_G$, we have $\|h_\theta\|_\infty\leq M_G$. Let $\tau$ and $M^\tau$ follow the training masking distribution, set $W^\tau=[L]\setminus M^\tau$, and let $I$ be uniform in $M^\tau$. Jensen's inequality gives
\begin{align*}
    Ph_\theta^2
    &\leq\bE\bigg[
        \log^2\frac{p_{\theta^\circ,I}(X_I\mid X_{W^\tau})}
                       {p_{\theta,I}(X_I\mid X_{W^\tau})}
    \bigg]\\
    &\leq2\bE\bigg[
        \log^2\frac{p_I^\star(X_I\mid X_{W^\tau})}
                       {p_{\theta,I}(X_I\mid X_{W^\tau})}
    \bigg]
    +2\bE\bigg[
        \log^2\frac{p_I^\star(X_I\mid X_{W^\tau})}
                       {p_{\theta^\circ,I}(X_I\mid X_{W^\tau})}
    \bigg].
\end{align*}
Here the expectation includes $X\sim p^\star$ and $(\tau,M^\tau,I)$. Conditioned on the context and $I$, the token $X_I$ has law $p_I^\star(\cdot\mid X_{W^\tau})$. Applying \eqref{eq:appendix-logloss-curvature} and the conditional KL expression for $\mathcal L$ therefore yields
\begin{align}
    Ph_\theta^2\leq6M_G\big(\mathcal L(\theta)+a\big).
    \label{eq:appendix-bernstein}
\end{align}

\paragraph{Control of the empirical deviation.}
Let $F_\theta \in \mathcal{F}(\Theta)$ be the logit function associated with $p_{\theta}$ through the rowwise softmax in \eqref{eq:general mask predictor class}. By the Lipschitz property of log-softmax,
\[
    \|h_\theta-h_{\theta'}\|_\infty
    =\|\ell_{p_{\theta}}-\ell_{p_{\theta'}}\|_\infty
    \leq2d_\infty(F_\theta,F_{\theta'}).
\]
For $r>0$, define
\[
    \mathcal H(r):=\{h_\theta:\theta\in\Theta,\ \mathcal L(\theta)\leq r\}.
\]
The covering bound in Assumption~\ref{ass:class} then implies
\[
    \log\mathcal N(\eta,\mathcal H(r),\|\cdot\|_\infty)
    \leq C(\Theta)\log\frac{8e}{\eta},
    \qquad 0<\eta\leq1,
\]
whenever $\mathcal H(r)$ is nonempty. To take the covering centers in $\mathcal H(r)$, choose one member from each ball intersecting this class and double the radius.

Let $a_n:=C(\Theta)\log(cn)$ for a sufficiently large universal constant $c\geq2$. Applying Lemma~\ref{lem:localized-entropy} with $b=M_G$ and $\sigma^2=6M_G(r+a)$ gives, for fixed $r,t>0$,
\begin{align}
    \sup_{h\in\mathcal H(r)}(P-P_n)h
    \leq C\bigg[
        \sqrt{\frac{M_G(r+a)(a_n+t)}{n}}
        +\frac{M_G(a_n+t)}{n}
    \bigg]
    \label{eq:oracle-local-deviation}
\end{align}
with probability at least $1-e^{-t}$.

We apply this bound at a sequence of increasing risk levels. Fix $u>0$ and set
\[
    r_0:=4a+A M_G\frac{a_n+u}{n},
    \qquad r_j:=2^jr_0,\quad j\geq1,
\]
where $A$ is a sufficiently large universal constant. Taking $r=r_j$ and $t=u+j\log2$ in \eqref{eq:oracle-local-deviation}, and using $\sum_{j\geq1}e^{-u-j\log2}=e^{-u}$, all these bounds hold simultaneously with probability at least $1-e^{-u}$. Moreover,
\[
    r_j+a\leq3r_{j-1},\quad
    a_n+u+j\log2\leq C2^{j-1}(a_n+u),\quad
    \frac{M_G(a_n+u+j\log2)}{n}\leq\frac{C}{A}r_{j-1}.
\]
Consequently, on the same event,
\[
    \sup_{h\in\mathcal H(r_j)}(P-P_n)h
    \leq C(A^{-1/2}+A^{-1})r_{j-1}
    \leq\frac12 \, r_{j-1},
\]
after choosing $A$ large enough.

Suppose that $\mathcal L(\widehat\theta)>r_0$, and choose $j\geq1$ such that $r_{j-1}<\mathcal L(\widehat\theta)\leq r_j$. Since $a\leq r_{j-1}/4$, the ERM property gives
\[
    \frac34r_{j-1}
    <\mathcal L(\widehat\theta)-a
    =Ph_{\widehat\theta}
    \leq(P-P_n)h_{\widehat\theta}
    \leq\frac12r_{j-1},
\]
a contradiction. Thus, with probability at least $1-e^{-u}$,
\[
    \mathcal L(\widehat\theta)
    \leq4a+C M_G\frac{C(\Theta)\log(en)+u}{n}.
\]
Using $a\leq\varepsilon_{\mathsf{approx}}(\Theta)+\zeta$ and letting $\zeta\downarrow0$, we obtain
\begin{align}
    \bP\, \bigg[
        \mathcal L(\widehat\theta)
        >4\varepsilon_{\mathsf{approx}}(\Theta)
        +C M_G\frac{C(\Theta)\log(en)+u}{n}
    \bigg]\leq e^{-u}.
    \label{eq:oracle-proof-tail}
\end{align}
Finally, integrating \eqref{eq:oracle-proof-tail} over $u>0$ yields
\begin{align*}
    \bE\big[\mathcal L(\widehat\theta)\big]
    &\leq4\varepsilon_{\mathsf{approx}}(\Theta)
      +\frac{C M_G}{n}
       \bigg[C(\Theta)\log(en)+\int_0^\infty e^{-u}\,\diff u\bigg]\\
    &=4\varepsilon_{\mathsf{approx}}(\Theta)
      +C M_G \cdot \frac{C(\Theta)\log(en)+1}{n}.
\end{align*}
This proves Proposition~\ref{prop:oracle}.

\subsection{Capacity of transformer logit class}
\label{app:cover}

In this section, we bound the capacity of the clipped logit class associated with the predictor in \eqref{eq:transformer predictor}. 
For $\theta=(E,P,\theta_{\mathrm{tf}},W_{\mathrm{out}})\in\Theta$, denote
\[
    F_\theta(x):=\Clip_{[-G,0]}\big(
        \TF_{\theta_{\mathrm{tf}}}
        (\Embed_{E,P}(x))W_{\mathrm{out}}\big),
    \qquad \F(\Theta):=\{F_\theta:\theta\in\Theta\}.
\]
Thus, $p_{\theta,i}(\cdot\mid x)=\operatorname{softmax}(F_{\theta,i,\cdot}(x))$. 
Recall the definition of $d_\infty(F,F')$ in \eqref{eq:infinity-norm difference}.

\begin{lemma}[Capacity of the sparse transformer class]
\label{lem:cover}
Let $\Theta=\Theta(D,N,M,D_{\mathrm{ff}},B,S;R_{\embed},G)$ be the parameter class in Section \ref{sec:estimator}, with $D,N,M,D_{\mathrm{ff}}\geq1$ and $B,R_{\mathrm{embed}},G>0$. Define
\[
    P_{\mathrm{tr}}
    :=(V+1)D+LD+N(3MD^2+2DD_{\mathrm{ff}})+DV = O(DV+LD+NMD^2+NDD_{\ff}),
\]
and suppose the sparsity $1\leq S\leq P_{\mathrm{tr}}$ is an integer. 
There exists a constant $A_{\mathrm{TF}}\geq e$, depending only on $N$, $M$, $B$, $R_{\embed}$, satisfying
\[
    \log A_{\mathrm{TF}}
    \leq CN^2\log\big(eNM(B\vee1)(R_{\mathrm{embed}}\vee1)\big),
\]
such that, for every $0<\eta\leq1$,
\begin{align}
    \log\mathcal N(\eta,\F(\Theta),d_\infty)
    \leq S\log\frac{eP_{\mathrm{tr}}}{S}
        +S\log\frac{A_{\mathrm{TF}}}{\eta}.
    \label{eq:sparse-cover}
\end{align}
Consequently, Assumption~\ref{ass:class} holds with
\[
    C(\Theta)
    =S\bigg[\log\frac{eP_{\mathrm{tr}}}{S}
            +\log A_{\mathrm{TF}}\bigg].
\]
The same covering bound holds for the probability-valued predictor class in the uniform distance: 
$$\max_{x\in \overline{\bV}^L} \max_{i,v} |p_{\theta,i}(v|x)-p_{\theta',i}(v|x)|.$$
If $S>P_{\mathrm{tr}}$, replace $S$ by $P_{\mathrm{tr}}$ in these bounds.
\end{lemma}

\paragraph{Proof of Lemma~\ref{lem:cover}.}
We first control the change in the logits under a perturbation of the parameters, and then cover the sparse parameter space. Write $\|Z\|_{2,\infty}:=\max_i\|Z_{i,\cdot}\|_2$, and let $d_\Theta(\theta,\theta')$ be the largest Frobenius distance between corresponding parameter matrices, including $E,P$, and $W_{\mathrm{out}}$. Set
\[
    \overline B:=B\vee1,\qquad
    \overline R:=R_{\embed}\vee1,\qquad
    a:=eNM\overline B\,\overline R.
\]

For $\|Z\|_{2,\infty}\leq r$, the bounded attention activation gives
\begin{align*}
    \|\head_\theta(Z)_{i,\cdot}\|_2
    &\leq\frac1L\sum_{j=1}^L
      \big|\sigma_{\attn}
        (\langle Z_{i,\cdot}Q,Z_{j,\cdot}K\rangle)\big|
      \|Z_{j,\cdot}V\|_2
    \leq Br.
\end{align*}
Hence the attention layer maps the row radius $r$ to at most $(1+MB)r$. The feed-forward layer multiplies this radius by at most $1+B^2$, since ReLU is one-Lipschitz and vanishes at zero. Denoting the hidden states by $H_\theta^{(\ell)}(x)$, with $H_\theta^{(0)}(x)=\Embed_{E,P}(x)$, we obtain
\[
    \|H_\theta^{(\ell)}(x)\|_{2,\infty}
    \leq R_{\embed}\big[(1+MB)(1+B^2)\big]^\ell
    \leq a^{cN},\qquad 0\leq\ell\leq N,
\]
uniformly over $x$ and $\theta\in\Theta$.

Next, let $\Delta:=d_\Theta(\theta,\theta')$. The trainable embeddings satisfy
\[
    \|H_\theta^{(0)}(x)-H_{\theta'}^{(0)}(x)\|_{2,\infty}
    \leq\|E-E'\|_{\mathrm F}+\|P-P'\|_{\mathrm F}
    \leq2\Delta.
\]
For one attention head, the sigmoid activation is $1/4$-Lipschitz. For inputs $Z,Z'$ with row norms at most $r$, we have
\begin{align*}
    \|\head_\theta(Z)
      -\head_\theta(Z')\|_{2,\infty}
    &\leq\big(B+\tfrac12 B^3r^2\big)
           \|Z-Z'\|_{2,\infty},\\
    \|\head_\theta(Z)
      -\head_{\theta'}(Z)\|_{2,\infty}
    &\leq\big(r+\tfrac12 B^2r^3\big)\Delta.
\end{align*}
For the feed-forward layer, the corresponding bounds are
\begin{align*}
    \|\FFN_\theta(Z)
      -\FFN_\theta(Z')\|_{2,\infty}
    &\leq(1+B^2)\|Z-Z'\|_{2,\infty},\\
    \|\FFN_\theta(Z)
      -\FFN_{\theta'}(Z)\|_{2,\infty}
    &\leq2Br\Delta.
\end{align*}
Combining these bounds with the hidden-state bound gives
\[
    \|H_\theta^{(\ell)}(x)-H_{\theta'}^{(\ell)}(x)\|_{2,\infty}
    \leq a^{cN}\big(
       \|H_\theta^{(\ell-1)}(x)-H_{\theta'}^{(\ell-1)}(x)\|_{2,\infty}
       +\Delta\big).
\]
Starting from the embedding perturbation bound and iterating over $N$ blocks, we obtain
\[
    \|H_\theta^{(N)}(x)-H_{\theta'}^{(N)}(x)\|_{2,\infty}
    \leq a^{cN^2}\Delta,
\]
after increasing the universal constant $c$.

For the output embedding, let $\|\cdot\|_{\infty,\infty}$ denote the largest absolute matrix entry. Then
\begin{align*}
    &\|H_\theta^{(N)}(x)W_{\mathrm{out}}
       -H_{\theta'}^{(N)}(x)W'_{\mathrm{out}}\|_{\infty,\infty}\\
    &\qquad\leq
       B\|H_\theta^{(N)}(x)-H_{\theta'}^{(N)}(x)\|_{2,\infty}
       +\|H_{\theta'}^{(N)}(x)\|_{2,\infty}
        \|W_{\mathrm{out}}-W'_{\mathrm{out}}\|_{\mathrm F}\\
    &\qquad\leq a^{cN^2}\Delta.
\end{align*}
Since clipping is one-Lipschitz, taking the supremum over masked sequences yields
\[
    d_\infty(F_\theta,F_{\theta'})
    \leq\Lambda d_\Theta(\theta,\theta'),
    \qquad \Lambda:=a^{cN^2}.
\]

It remains to cover the parameters. The input and positional embeddings have $(V+1)D+LD$ entries, each transformer block has $3MD^2+2DD_{\ff}$ entries, and the output embedding has $DV$ entries. This gives the stated value of $P_{\mathrm{tr}}$.

For a fixed support $\mathcal I\subseteq[P_{\mathrm{tr}}]$ of size $S$, denote
\[
    \Theta_{\mathcal I}
    :=\{\theta\in\Theta:\supp(\theta)\subseteq\mathcal I\}.
\]
Ignoring the embedded-input constraint temporarily, the parameters on this support lie in a product of Frobenius balls of radius $B$, with total dimension $S$. Covering each ball at radius $\zeta/2$ gives a product cover of size at most $(1+4B/\zeta)^S$ in the maximum Frobenius distance. For each ball intersecting $\Theta_{\mathcal I}$, choose one parameter from the intersection as its new center. The radius is then at most $\zeta$, and every center satisfies all constraints, including the embedded-input bound. Consequently,
\[
    \log\mathcal N(\zeta,\Theta_{\mathcal I},d_\Theta)
    \leq S\log\bigg(1+\frac{4B}{\zeta}\bigg).
\]
Every parameter with at most $S$ nonzero entries belongs to one of these support classes. Taking the union over supports gives
\begin{align*}
    \log\mathcal N(\zeta,\Theta,d_\Theta)
    &\leq\log\binom{P_{\mathrm{tr}}}{S}
       +S\log\bigg(1+\frac{4B}{\zeta}\bigg)\\
    &\leq S\log\frac{eP_{\mathrm{tr}}}{S}
       +S\log\bigg(1+\frac{4B}{\zeta}\bigg).
\end{align*}
Take $\zeta=\eta/\Lambda$ and $A_{\mathrm{TF}}:=e(1+4B\Lambda)$. Transferring the parameter cover to the logits proves \eqref{eq:sparse-cover}, and
\[
    \log A_{\mathrm{TF}}
    \leq CN^2\log\big(eNM(B\vee1)(R_{\embed}\vee1)\big).
\]
Since $A_{\mathrm{TF}}\geq e$ and $0<\eta\leq1$, the covering bound is at most $C(\Theta)\log(e/\eta)$. The logits already lie in $[-G,0]$, so both parts of Assumption~\ref{ass:class} hold.

Finally, softmax is one-Lipschitz in the maximum norm. Applying it rowwise therefore transfers the same cover to the predictor class. This completes the proof.

\subsection{Proof of Proposition~\ref{prop:gen error}}
\label{app:proof of prop: gen err}
Let $\Theta = \Theta (D,N,M,D_{\ff},B,S;R_{\embed},G)$ be the parameter class in Section \ref{sec:estimator}, 
and let $\mathcal{F}(\Theta)$ be its clipped logit class. Every logit entry lies in $[-G,0]$, and Lemma \ref{lem:cover} verifies Assumption \ref{ass:class} with
\begin{align*}
    C(\Theta) \le S\bigg[\log\frac{eP_{\mathrm{tr}}}{S}
            +\log A_{\mathrm{TF}}\bigg]. 
\end{align*}
The induced predictor class is exactly the class used in \eqref{eq:ERM estimator}, so the corresponding ERM is $\widehat{\theta}$ and 
$\inf_{\theta \in \Theta} \mathcal{L}(\theta) = \varepsilon_{\mathsf{approx}} (\Theta)$.
Applying Proposition~\ref{prop:oracle} proves Proposition~\ref{prop:gen error}:
\begin{align*}
    \bE[\loss(\widehat{\theta})] \lesssim \varepsilon_{\mathsf{approx}}(\Theta) + (G+\log V) \frac{C(\Theta)\log (en)+1}{n}.
\end{align*}
Here the expectation is taken over training data $\{X^{(j)}\}_{j=1}^n$.

\subsection{Localized empirical-process lemma}
\label{app:localized-entropy}
\begin{lemma}[Localized bounds from a uniform cover]
\label{lem:localized-entropy}
Let $X^{(1)},\ldots,X^{(n)}$ be independent samples from a distribution $P$, and write $Pg:= \bE[g(X)]$ and $P_n g:=n^{-1}\sum_{i=1}^n g(X^{(i)})$ for any function $g$.
Let $\mathcal G$ be a nonempty function class with $\|g\|_\infty\le b$ and $Pg^2\le\sigma^2$ for every $g\in\mathcal G$, where $b\ge1$. Suppose $v\ge1$, $A\ge e$, and 
\[
 \log\mathcal N(\eta,\mathcal G,\|\cdot\|_\infty)
 \le v\log(A/\eta),\quad \forall \, 0<\eta\le1.
\]
For $n\ge2$, take $d_n:=v\log(cAn)$, for a universal $c\ge2$. Then 
for $t>0$, with probability at least $1-e^{-t}$,
\[
 \sup_{g\in\mathcal G}(P-P_n)g
 \le C\Big(\sigma\sqrt{\frac{d_n+t}{n}}+\frac{b(d_n+t)}n\Big).
\]
\end{lemma}
\paragraph{Proof of Lemma \ref{lem:localized-entropy}.}
Take an internal $1/n$-net $\{g_1,\ldots,g_{J_{\rm net}}\}\subseteq\mathcal G$. 
If covering centers are initially outside the class, start with a $1/(2n)$-cover and choose one element of each intersecting ball. 
Thus $\log(2J_{\rm net})\le v\log(4An)\le d_n$ after increasing $c$. All centers retain the variance and envelope bounds.

For the deviation assertion, the independent centered variables $Pg_j-g_j(X^{(i)})$ have magnitude at most $2b$ and variance at most $\sigma^2$. Bernstein's inequality and a union bound then give
\[
 \max_{1\le j\le J_{\rm net}}(P-P_n)g_j
 \le \sqrt{\frac{2\sigma^2(\log J_{\rm net}+t)}n}
      +\frac{2b(\log J_{\rm net}+t)}{3n}
\]
with probability at least $1-e^{-t}$. Replacing a function by its net center changes $(P-P_n)g$ by at most $2/n$, which is absorbed in $b(d_n+t)/n$.


\subsection{Conditional-KL identity for excess risk}
\label{app:kl}
\begin{lemma}[Per-token denoising loss is average conditional KL]
\label{lem:kl}
Recall the definition of $\tau$, $M^{\tau}$ and $W^{\tau}$ in Section~\ref{sec:mdm}.
Conditioned on $(\tau,M^{\tau})$, draw $I$ uniformly from $M^\tau$.
Then the excess risk $\mathcal{L}(\theta)$ satisfies
\begin{align}
 \loss(\theta)&:=\overline{ R}(p_{\theta})-\overline{ R}(p^\star)\notag\\
 &=\bE_{\tau,M^\tau,I,\,X\sim p^\star}
   \Big[\KL \big(p_I^\star(\cdot\mid X_{W^\tau})\, \|\,
                        p_{\theta,I}(\cdot\mid X_{W^\tau})\big)\Big].
 \label{eq:kl-risk}
\end{align}
Here $\overline{R}(p_{\theta}):= \bE_{X\sim p^\star}[\ell_{p_{\theta}}(X)]$ is the risk of the conditional predictor $p_{\theta}$.
\end{lemma}

\paragraph{Proof of Lemma \ref{lem:kl}.}
Subtract the cross-entropy of the true conditional predictor from that of the network and introduce a conditionally uniform $I \sim \mathrm{Unif}(M^\tau)$. Then
\begin{align*}
 \overline{R}(p_{\theta})-\overline{R}(p^\star)
&=\bE_{\tau,M^{\tau},X\sim p^\star}\Big[\frac1{|M^\tau|}\sum_{i\in M^\tau}
       \log\frac{p_i^\star(X_i\mid X_{W^\tau})}{p_{\theta,i}(X_i\mid X_{W^\tau})}\Big]\\
       &=\bE_{\tau,M^{\tau},I,X\sim p^\star}\Big[\log\frac{p_I^\star(X_I\mid X_{W^\tau})}
                          {p_{\theta,I}(X_I\mid X_{W^\tau})}\Big].
\end{align*}
Conditioned on $(\tau,M^{\tau},X_{W^\tau},I)$, the token distribution $X_I$ follows the law $p_I^\star(\cdot\mid X_{W^\tau})$. 
Taking the conditional expectation therefore gives the KL divergence in \eqref{eq:kl-risk}.

\section{Transformer approximation of conditionals}
In this appendix, we prove Proposition \ref{prop:tree-mixture-approximation} by constructing sparse transformers that approximate the conditional probabilities of mixtures of hidden Markov trees.
Appendix \ref{app:regularization} introduces a regularization that preserves the tree-mixture structure and ensures positive conditional probabilities.
Appendix \ref{app:decomp} further reduces the approximation error to uniform approximation of the regularized conditional logits.
Appendix \ref{app:single-tree-parallel} then constructs a transformer-based approximation under a single tree structure using parallel contraction and reconstruction, 
and Appendix \ref{app:mixture-parallel-construction} combines these using posterior mixture weights. 
Appendix \ref{app:pf-of-prop:tree-mixture-approx} completes the proof of Proposition \ref{prop:tree-mixture-approximation}.

\subsection{Regularization}
\label{app:regularization}
Here we regularize distributions while preserving their tree-mixture structure. 
This gives positive conditional probabilities and bounded conditional log probabilities for the following transformer approximation.

\begin{definition}[$\delta$-regularization]
\label{def:regularization}
    For $0<\delta<1$, define the regularization kernel $Q_\delta(a,b):=(1-\delta)\mathbf1\{a=b\}+\delta/V$.
    For any distribution $q$ on $\bV^L$, define its $\delta$-regularization by
      \begin{align}
      q^\delta(y):=(P_\delta q)(y)
      :=\sum_{x\in\bV^L}q(x)\prod_{i=1}^LQ_\delta(x_i,y_i),
      \qquad y\in\bV^L.
    \label{eq:refresh}
      \end{align}
      This transition independently replaces each token by a uniform token with probability $\delta$.
\end{definition}

\begin{lemma}[Properties of $\delta$-regularization]
\label{lem:regularization}
Let $0<\delta<1$, $\rho_\delta:=\delta/V$, and $G_\delta:=\log(V/\delta)$. The following properties hold.
\begin{enumerate}[label=\textup{(\roman*)}]
    \item If $q=\sum_{r=1}^R\pi_rq_r\in\mathcal M_{J,R}$, then
    \begin{align}
        q^\delta=\sum_{r=1}^R\pi_rq_r^\delta\in\mathcal M_{J,R},
        \quad
        B_{i,r}^\delta(v,z)
        =(1-\delta)B_{i,r}(v,z)+\frac{\delta}{V}.
        \label{eq:refresh-mixture}
    \end{align}
    Here $q_r^\delta=P_\delta q_r$ is obtained by replacing the emission tables $B_{i,r}$ with $B_{i,r}^\delta$ as in \eqref{eq:hidden-tree-law}. 
    The root distributions, transitions, component trees, and mixture weights are unchanged.

    \item For any distribution $q$ on $\bV^L$, any $y\in\bV^L$, and any masked sequence $c\in\overline{\bV}^L$,
    it holds that:
    \begin{align}
        q^\delta(y) \geq\rho_\delta^L,\quad  q^\delta(y)\geq(1-\delta)^Lq(y),\quad q_i^\delta(v\mid c)\geq\rho_\delta, \quad \forall i\notin W(c),\text{ and } v\in\bV.
        \label{eq:refresh-positive}
    \end{align}
    Here $W(c):=\{j\in[L]:c_j\ne\mask\}$ and
    $q_i^\delta(v\mid c):=\bP_{X\sim q^\delta}[X_i=v\mid X_{W(c)}=c_{W(c)}]$.
    Hence, $-G_{\delta}\le \log q_i^{\delta}(v\mid c) \le 0$. 

    \item For any distributions $p,q$ on $\bV^L$,
    \begin{align}
        \KL(p\|q^\delta)
        \leq\KL(p\|q)+L\log\frac1{1-\delta}.
        \label{eq:refresh-kl}
    \end{align}
\end{enumerate}
\end{lemma}

\paragraph{Proof of Lemma~\ref{lem:regularization}.}
For part (i), fix a representation $q=\sum_{r=1}^R\pi_rq_r$ as in Section~\ref{sec:problem setup}. Notice that
\[
    \sum_{a\in\bV}B_{i,r}(a,z)Q_\delta(a,v)
    =(1-\delta)B_{i,r}(v,z)+\frac\delta V
    =B_{i,r}^\delta(v,z).
\]
Substituting the hidden-tree factorization into \eqref{eq:refresh} gives
\begin{align}
    (P_\delta q_r)(y)
    &=\sum_{z\in[J]^L}\nu_r(z_{o_r})
      \prod_{i\ne o_r}A_{i,r}(z_i,z_{\pa_r(i)})
      \prod_{i=1}^L\bigg[\sum_{a\in\bV}
          B_{i,r}(a,z_i)Q_\delta(a,y_i)\bigg]\notag\\
    &=\sum_{z\in[J]^L}\nu_r(z_{o_r})
      \prod_{i\ne o_r}A_{i,r}(z_i,z_{\pa_r(i)})
      \prod_{i=1}^L B_{i,r}^\delta(y_i,z_i)
      =q_r^\delta(y).
    \label{eq:proof-refresh-factorization}
\end{align}
Thus, regularization changes only the emission tables. By linearity, $q^\delta=\sum_{r=1}^R\pi_rq_r^\delta\in\mathcal M_{J,R}$. Since \eqref{eq:refresh} is defined using the observed law $q$, the resulting law does not depend on the chosen hidden-tree representation.

For part (ii), the bounds $Q_\delta(a,b)\geq\rho_\delta$ and $Q_\delta(a,a)\geq1-\delta$ imply
\begin{align}
    q^\delta(y)
    &\geq\sum_{x\in\bV^L}q(x)\rho_\delta^L
     =\rho_\delta^L,
    \label{eq:proof-refresh-full-support}\\
    q^\delta(y)
    &\geq q(y)\prod_{i=1}^LQ_\delta(y_i,y_i)
     \geq(1-\delta)^Lq(y).
    \label{eq:proof-refresh-domination}
\end{align}
Fix a masked sequence $c$ and write $W=W(c)$. Summing over the coordinates outside $W$ gives
\[
    \bP_{X\sim q^\delta}[X_W=c_W]
    =\sum_{x\in\bV^L}q(x)\prod_{j\in W}Q_\delta(x_j,c_j)
    \geq\rho_\delta^{|W|}>0.
\]
Hence, for $i\notin W$ and $v\in\bV$,
\[
    q_i^\delta(v\mid c)
    =\frac{\displaystyle\sum_{x\in\bV^L}q(x)
        Q_\delta(x_i,v)\prod_{j\in W}Q_\delta(x_j,c_j)}
       {\displaystyle\sum_{x\in\bV^L}q(x)
        \prod_{j\in W}Q_\delta(x_j,c_j)}
    \geq\rho_\delta.
\]

For part (iii), the result is immediate if $\KL(p\|q)=+\infty$. Otherwise, applying \eqref{eq:proof-refresh-domination} yields
\begin{align*}
    \KL(p\|q^\delta)-\KL(p\|q)
    &=\sum_{y:p(y)>0}p(y)\log\frac{q(y)}{q^\delta(y)}\\
    &\leq L\log\frac1{1-\delta}.
\end{align*}
This proves the lemma. 
In particular, regularization also preserves any restriction on the component tree topologies, including a prescribed height bound.

\subsection{Decomposition of the approximation error}
\label{app:decomp}
In this section, we bound the approximation error by the joint KL error of a comparison distribution, the regularization error, and the network approximation error.

Fix $0<\delta<1$ and a distribution $q$ on $\bV^L$. 
For a masked sequence $c\in\overline{\bV}^L$, define the target logits of its $\delta$-regularized distribution as in Definition \ref{def:regularization} by
\begin{align}
    F_{i,v}^{\delta,q}(c)
    :=\begin{cases}
        \log q_i^\delta(v\mid c),&i\notin W(c),\\
        0,&i\in W(c),
    \end{cases}
    \quad \forall i\in[L], \, \forall v\in\bV.
    \label{eq:regularized-target-logits}
\end{align}
By Lemma~\ref{lem:regularization}, these logits lie in $[-G_\delta,0]$, and
\[
    \softmax(F_{i,\cdot}^{\delta,q}(c))
    =q_i^\delta(\cdot\mid c),\qquad i\notin W(c).
\]
The values at unmasked positions are set to zero only to define a full $L\times V$ matrix; they do not enter the loss or the approximation distance below.

Recall the clipped logit class $\F(\Theta)$ in Appendix~\ref{app:cover}. 
Define its uniform approximation error on masked positions by
\begin{align}
    d_{\infty,\mathrm{mask}}(F^{\delta,q},\F(\Theta))
    :=\inf_{\theta\in\Theta}
      \sup_{\substack{c\in\overline{\bV}^L\\W(c)\ne[L]}}
      \max_{\substack{i\notin W(c)\\v\in\bV}}
      |F_{\theta,i,v}(c)-F_{i,v}^{\delta,q}(c)|.
    \label{eq:masked-logit-distance}
\end{align}

\begin{lemma}[Decomposition of the approximation error]
\label{lem:decomp}
Fix $0<\delta<1$ and a fixed-cardinality masking schedule $s=(s_1,\ldots,s_T)$, with $s_t\in\bN_+$ and $\sum_{t=1}^T s_t=L$. For any distribution $q$ on $\bV^L$,
\begin{align}
    \varepsilon_{\mathsf{approx}}(\Theta)
    \leq\frac{s_{\max}}{L}\KL(p^\star\|q^\delta)
       +2d_{\infty,\mathrm{mask}}(F^{\delta,q},\F(\Theta)),
    \label{eq:dist-net}
\end{align}
where $s_{\max}:=\max_t s_t$ and $\varepsilon_{\mathsf{approx}}(\Theta):=\inf_{\theta\in\Theta}\loss(\theta)$. 
In particular, if function class $\F(\Theta)$ satisfies
\[
    \sup_{q\in\mathcal M_{J,R}}
    d_{\infty,\mathrm{mask}}(F^{\delta,q},\F(\Theta))
    \leq\epsilon,
\]
then
\begin{align}
    \varepsilon_{\mathsf{approx}}(\Theta)
    \leq\frac{s_{\max}}{L}\Proj_{J,R}(p^\star)
       +s_{\max}\log\frac1{1-\delta}+2\epsilon.
    \label{eq:approx-oracle}
\end{align}
The same conclusion holds for any nonempty subclass $\mathcal Q\subseteq\mathcal M_{J,R}$, with the uniform approximation condition imposed on $\mathcal Q$ and $\Proj_{J,R}$ replaced by $\Proj_{\mathcal Q}(p^\star) := \inf_{q\in \mathcal{Q}} \KL(p^\star \| q)$.
\end{lemma}

\paragraph{Proof of Lemma~\ref{lem:decomp}.}
Fix $q$ and write $r=q^\delta$. Let $r_{\mathrm{cond}}$ denote its conditional predictor on masked positions. 
We use the same training experiment as in Section~\ref{sec:prelim}: $\bP[\tau=t]=s_t/L$, $M^\tau$ is uniform among subsets of size $m_\tau:=\sum_{u=1}^\tau s_u$, and $I$ is uniform in $M^\tau$. 
Write $W^\tau=[L]\setminus M^\tau$, and let $X\sim p^\star$. 
As in Section \ref{sec:prelim}, $X_{W^\tau}$ denotes the sequence obtained by replacing all tokens outside $W^\tau$ with $\mask$.
Define
\[
    \loss(r_{\mathrm{cond}})
    :=\bE\big[\KL\big(p_I^\star(\cdot\mid X_{W^\tau})
         \,\|\,r_I(\cdot\mid X_{W^\tau})\big)\big].
\]
By Lemma~\ref{lem:kl}, adding and subtracting the log conditional probability under $r$ gives
\begin{align}
    \loss(\theta)
    &=\bE\bigg[\log\frac{p_I^\star(X_I\mid X_{W^\tau})}
                                 {p_{\theta,I}(X_I\mid X_{W^\tau})}\bigg]\notag\\
    &=\loss(r_{\mathrm{cond}})
      +\bE\bigg[\log\frac{r_I(X_I\mid X_{W^\tau})}
                              {p_{\theta,I}(X_I\mid X_{W^\tau})}\bigg].
    \label{eq:proof-exact-mixture-network-split}
\end{align}
For any $f,g\in\bR^V$, the log-sum-exp function is one-Lipschitz in the maximum norm. Hence,
\begin{align}
    |\log\softmax_v(f)-\log\softmax_v(g)|
    &\leq |f_v-g_v|
       +\bigg|\log\sum_{a=1}^V e^{f_a}
              -\log\sum_{a=1}^V e^{g_a}\bigg|\notag\\
    &\leq2\|f-g\|_\infty.
    \label{eq:proof-logsoftmax-lipschitz}
\end{align}
Since $r_i(\cdot\mid x)=\softmax(F_{i,\cdot}^{\delta,q}(x))$ for $i\notin W(x)$, applying this bound in \eqref{eq:proof-exact-mixture-network-split} and taking the infimum over $\theta\in\Theta$ yields
\[
    \varepsilon_{\mathsf{approx}}(\Theta)
    \leq\loss(r_{\mathrm{cond}})
       +2d_{\infty,\mathrm{mask}}(F^{\delta,q},\F(\Theta)).
\]

It remains to control $\loss(r_{\mathrm{cond}})$ by $\KL(p^\star\|r)$. We use a subset-averaging argument, analogous to the entropy calculation in \citet[proof of Theorem~3.3]{chen2025optimal}. For $W\subseteq[L]$, denote by $p_W^\star$ and $r_W$ the corresponding marginals. Let
\[
    \overline d_j
    :=\binom Lj^{-1}\sum_{\substack{W\subseteq[L]\\|W|=j}}
       \KL(p_W^\star\|r_W),\quad \forall \, 0\leq j\leq L,
\]
and write $\Delta_j:=\overline d_{j+1}-\overline d_j$ for $0\leq j<L$. By the KL chain rule, for $i\notin W$,
\begin{align*}
    \KL(p_{W\cup\{i\}}^\star\|r_{W\cup\{i\}})
    -\KL(p_W^\star\|r_W)
    =\bE_{X_W\sim p_W^\star}
      \big[\KL\big(p_i^\star(\cdot\mid X_W)
           \,\|\,r_i(\cdot\mid X_W)\big)\big].
\end{align*}
If $W$ is a uniform $j$-subset and $I$ is uniform in $[L]\setminus W$, then $W\cup\{I\}$ is a uniform $(j+1)$-subset. Averaging the preceding identity gives
\begin{align}
    \bE_{W,I,X_W}\big[
        \KL\big(p_I^\star(\cdot\mid X_W)
           \,\|\,r_I(\cdot\mid X_W)\big)\big]
    =\Delta_j\geq0.
    \label{eq:proof-kl-level-increment}
\end{align}
All these divergences are finite because $r$ has full support. Since $\overline d_0=0$ and $\overline d_L=\KL(p^\star\|r)$,
\begin{align}
    \sum_{j=0}^{L-1}\Delta_j=\KL(p^\star\|r).
    \label{eq:proof-kl-increments-telescope}
\end{align}
At training time $t$, the context set $W^t$ is uniform of size $L-m_t$, where $m_t=\sum_{u=1}^t s_u$. 
Therefore,
\begin{align}
    \loss(r_{\mathrm{cond}})
    =\frac1L\sum_{t=1}^T s_t\Delta_{L-m_t}
    \leq\frac{s_{\max}}{L}\sum_{j=0}^{L-1}\Delta_j
     =\frac{s_{\max}}{L}\KL(p^\star\|r).
    \label{eq:proof-general-schedule-kl}
\end{align}
Here we use $\Delta_j\geq0$ and the fact that the indices $L-m_t$ are distinct, since $s_t\geq1$. Combining this with the network approximation bound proves \eqref{eq:dist-net}.

Under the uniform approximation condition, Lemma~\ref{lem:regularization}(iii) further gives, for every $q\in\mathcal M_{J,R}$,
\begin{align*}
    \varepsilon_{\mathsf{approx}}(\Theta)
    &\leq\frac{s_{\max}}{L}\KL(p^\star\|q^\delta)+2\epsilon\\
    &\leq\frac{s_{\max}}{L}\KL(p^\star\|q)
       +s_{\max}\log\frac1{1-\delta}+2\epsilon.
\end{align*}
Taking the infimum over $q\in\mathcal M_{J,R}$ proves \eqref{eq:approx-oracle}. The same argument applies when this infimum is restricted to $\mathcal Q$.

Lemma \ref{lem:decomp} reduces Proposition \ref{prop:tree-mixture-approximation} to uniformly approximating the regularized conditional logits $F^{\delta,q}$ over $q \in \mathcal{M}_{J,R}$ using a transformer class.
We establish this approximation first for a single tree in Appendix \ref{app:single-tree-parallel} and then for mixtures in Appendix \ref{app:mixture-parallel-construction}.

\subsection{Single-tree predictor construction}
\label{app:single-tree-parallel}
In this subsection, we construct a sparse transformer that approximates the regularized conditional logits under a single hidden Markov tree structure. 
Parallel contraction computes the context likelihood, and reconstruction recovers the hidden-state posteriors needed to evaluate the conditional probabilities.
We maintain the context likelihood approximation for the mixture construction in Appendix \ref{app:mixture-parallel-construction}. 


Throughout Appendices \ref{app:single-tree-parallel} and \ref{app:mixture-parallel-construction}, 
let $L,V\ge2$, $J,R\ge1$, $0<\delta<1$, and $0<\varepsilon\le1/2$. Write
\[
 \rho_\delta:=\delta/V,\quad G_\delta:=\log(V/\delta),\quad
 H:=(L+1)G_\delta+\log(64LJ/\varepsilon),\quad
 b_*:=\log(eLJVH/\varepsilon).
\]
For a masked sequence $c\in\overline{\bV}^L$, let $W(c):=\{i:c_i\ne\mask\}$ and write $q(c):=\bP_q[X_{W(c)}=c_{W(c)}]$ for its context likelihood. 
We use the transformer parameter class in Section~\ref{sec:estimator}, including the trainable input and positional embedding, 
and the associated clipped logit class $\F(\Theta)$ from Appendix~\ref{app:cover}. 
The log-sum-exp operator is denoted by $\LSE$, 
\begin{align*}
     \LSE(x) := \log \sum_{j=1}^d e^{x_j}, \quad \text{for } x\in \bR^d.
\end{align*}
Its neural-network approximation is given in Lemma \ref{lem:nn-lse-approx}.

\begin{lemma}[Single-tree parallel construction]
\label{lem:single-tree-parallel}
Let $q$ be a $J$-state hidden Markov tree model on $[L]$, with root $o$, root distribution $\nu$, transition tables $A_i$, and emission tables $B_i$ satisfying $B_i(v,z)\ge\rho_\delta$. There exists a parameter class
$\Theta=\Theta(D,N,M,D_{\rm ff},B,S;\sqrt3,G_\delta)$ with budgets depending only on $(L,V,J,\delta,\varepsilon)$, and a parameter $\theta \in \Theta$ satisfying
\begin{align}
 \sup_{c:\,W(c)\ne[L]}\max_{i\notin W(c),\,v\in\bV}
 |F_{\theta,i,v}(c)-\log q_i(v\mid c)|\le\varepsilon.
 \label{eq:single-tree-parallel-goal}
\end{align}
The architectural budgets can be chosen as
\begin{align}
 N&\le Cb_*^3,\qquad M\le2,\qquad \log B\le Cb_*^2,\notag\\
 D&\le C\big[L+V+(J^3+JV)b_*^3\big],\qquad
 D_{\rm ff}\le C\big[LV+(J^3+JV)b_*^3\big],\notag\\
 S&\le C\big[L(J^2+JV)+J^3+JV\big]b_*^{10}.
 \label{eq:single-tree-parallel-budgets}
\end{align}
Before the output embedding, the construction also contains a coordinate $\widehat h_i(c)$ at every row satisfying
\begin{align}
 \sup_{c\in\overline{\bV}^L}\max_{i\in[L]}
 |\widehat h_i(c)-\log q(c)|\le\varepsilon,
 \label{eq:single-tree-evidence-goal}
\end{align}
which approximates the log context likelihood $\log q(c)$. 
The budgets are independent of the topology and probability tables. 
\end{lemma}

By Lemma~\ref{lem:regularization}, the lemma applies to $q^\delta=P_\delta q$ for any $q\in\mathcal M_{J,1}$. 
In this case, \eqref{eq:single-tree-parallel-goal} approximates $F^{\delta,q}$ on masked positions. 
The additional coordinate in \eqref{eq:single-tree-evidence-goal} will be used later to combine different components.

\paragraph{Proof of Lemma~\ref{lem:single-tree-parallel}.}
We first eliminate vertices while preserving the marginal law on the remaining tree. 
Reversing these operations recovers all hidden-state posteriors. 
We then implement the contraction and reconstruction steps by transformer blocks and control their errors.

\paragraph{Positive hidden probabilities.}
The emission lower bound ensures positive conditional probabilities, but the root distribution and transition tables may still contain zeros. 
We therefore perturb these hidden probabilities before working in logarithmic coordinates. 
Let $u_J=J^{-1}\ind$ and define
\begin{align}
 \gamma:=\frac{\varepsilon\rho_\delta^L}{64L},\qquad
 \nu^\gamma:=(1-\gamma)\nu+\gamma u_J,\qquad
 A_i^\gamma:=(1-\gamma)A_i+\gamma u_J\ind^\top.
 \label{eq:single-tree-smoothing}
\end{align}
Denote the resulting observed law by $q^\gamma$. 
A coupling of the $L$ hidden draws gives
$\mathsf{TV}(q,q^{\gamma}) \le L\gamma$.
Both laws assign probability at least $\rho_\delta^{|W|}$ to a context on $W$, and at least $\rho_\delta^{|W|+1}$ to that context together with one masked token. Thus,
\begin{align}
 |\log q_i^\gamma(v\mid c)-\log q_i(v\mid c)|
 &\le 2L\gamma\rho_\delta^{-L}=\varepsilon/32,\notag\\
 |\log q^\gamma(c)-\log q(c)|
 &\le L\gamma\rho_\delta^{-L}=\varepsilon/64.
 \label{eq:single-tree-smoothing-error}
\end{align}
Here we use $|\log a-\log b|\le |a-b|/\min\{a,b\}$. 
Below, the root distribution and transition matrices refer to the regularized model, and we suppress their superscript $\gamma$. Notice that
\[
 H=G_\delta+\log(J/\gamma),\qquad
 (\gamma/J)^L\rho_\delta^L=e^{-LH}.
\]

\paragraph{Contraction and the marginal law.}
In the following, we consider iterative tree contraction and denote $\mathcal{V}_i$ as the vertices of the remaining tree at stage $i$.
Let $\mathcal V_0=[L]$. Initially, set
\[
 g_i(z,c):=\begin{cases}B_i(c_i,z),&i\in W(c),\\1,&i\notin W(c),\end{cases}
 \qquad
 \psi_o^{(0)}(z):=\nu(z)g_o(z,c),\quad
 \psi_i^{(0)}(z):=g_i(z,c)\quad(i\ne o),
\]
and $T_i^{(0)}:=A_i$ for $i\ne o$. Here $T_i^{(s)}$ denotes an effective edge factor; $B_i$ continues to denote an emission table. Write $\pa_s$ for the parent map of the remaining tree after round $s$. 
We maintain
\begin{align}
 \bP_{q^\gamma}[Z_{\mathcal V_s}=z_{\mathcal V_s},c]
 =\prod_{i\in\mathcal V_s}\psi_i^{(s)}(z_i)
  \prod_{i\in\mathcal V_s\setminus\{o\}}
       T_i^{(s)}(z_i,z_{\pa_s(i)}).
 \label{eq:single-tree-factor-invariant}
\end{align}
This holds initially by the hidden-tree factorization. 
There are two elimination operations. In the formulas below, all factors on the right are taken before the round.
\begin{itemize}
 \item If $v$ is a nonroot leaf with parent $u$, define
 \[
  \ell_v(a):=\sum_{b=1}^J T_v(b,a)\psi_v(b),\qquad
  \psi_u(a)\leftarrow\psi_u(a)\ell_v(a),
 \]
 and remove $v$.
 \item If $v$ is a nonroot vertex with parent $u$ and unique child $w$, define
 \begin{align}
  C_v(z,a):=\sum_{b=1}^J T_w(z,b)\psi_v(b)T_v(b,a).
  \label{eq:single-tree-contract-edge}
 \end{align}
 Remove $v$, connect $w$ to $u$, and replace $T_w$ by $C_v$.
\end{itemize}
Both operations sum out $Z_v$ in \eqref{eq:single-tree-factor-invariant}. As in tree contraction \citep{miller1985contraction}, we apply them to nonadjacent vertices. 
If several selected leaves share a parent, their factors $\ell_v$ are multiplied into its node factor. 
The eliminated variables have no edges between them, so simultaneous elimination gives exactly the marginal on the remaining tree.

At round $s$, consider the nonroot vertices in $\mathcal V_{s-1}$ with at most one child. Select the larger of their even- and odd-depth sets, breaking ties by choosing even depth, and call it $I_s$. 
Take $\mathcal V_s:=\mathcal V_{s-1}\setminus I_s$. For a tree with $n\ge2$ vertices, if $b$ vertices have degree at least three, then
\[
 2n-2\ge3b+(n-b),\qquad b\le(n-2)/2.
\]
After excluding the root, at least $n/2$ vertices have at most one child. 
One depth parity therefore contains at least $n/4$ nonroot vertices with at most one child, and vertices of the same parity are nonadjacent. Consequently,
\begin{align}
 |I_s|\ge |\mathcal V_{s-1}|/4,\qquad
 |\mathcal V_s|\le\tfrac34|\mathcal V_{s-1}|,\qquad
 s_0\le1+\lceil\log_{4/3}L\rceil,
 \label{eq:single-tree-rounds}
\end{align}
where $\mathcal V_{s_0}=\{o\}$. 
The schedule depends only on the topology, and shows that the number of rounds is controlled by $O(\log L)$.

\paragraph{Reconstruction of the posteriors.}
When removing $v\in I_s$, save $\psi_v^{(s-1)}$ and $T_v^{(s-1)}$, together with $\ell_v$ if $v$ is a leaf, or $T_w^{(s-1)}$ and $C_v$ if it has one child $w$. Each row is removed once, so these saved factors require $O(J^2)$ coordinates per row.

At the end of contraction, \eqref{eq:single-tree-factor-invariant} gives
\[
 \psi_o^{(s_0)}(a)=\bP_{q^\gamma}[Z_o=a,c],\qquad
 q^\gamma(c)=\sum_a\psi_o^{(s_0)}(a),\qquad
 p_o(a):=\frac{\psi_o^{(s_0)}(a)}{q^\gamma(c)}.
\]
Suppose that, before reversing round $s$, we know
\begin{align*}
      p_u(a)&=\bP_{q^\gamma}[Z_u=a\mid c]\quad(\text{for }u\in\mathcal V_s),\\
 P_u^{(s)}(z,a)&=\bP_{q^\gamma}[Z_u=z,Z_{\pa_s(u)}=a\mid c]
 \quad(\text{for } u\in\mathcal V_s\setminus\{o\}).
\end{align*}
We recover the node and edge posteriors on $\mathcal V_{s-1}$ as follows.
\begin{itemize}
 \item If $v$ was a leaf with parent $u$, then
 \begin{align}
  P_v^{(s-1)}(b,a)
  =p_u(a)\frac{\psi_v^{(s-1)}(b)T_v^{(s-1)}(b,a)}{\ell_v(a)},
  \qquad p_v(b)=\sum_aP_v^{(s-1)}(b,a).
  \label{eq:single-tree-restore-leaf}
 \end{align}
 \item If $v$ had parent $u$ and child $w$, form
 \begin{align}
  \Gamma_v(z,b,a)
  :=P_w^{(s)}(z,a)
   \frac{T_w^{(s-1)}(z,b)\psi_v^{(s-1)}(b)T_v^{(s-1)}(b,a)}{C_v(z,a)}.
  \label{eq:single-tree-restore-triple}
 \end{align}
 Then
 \begin{align*}
      P_v^{(s-1)}(b,a)=\sum_z\Gamma_v(z,b,a),&\quad
  P_w^{(s-1)}(z,b)=\sum_a\Gamma_v(z,b,a),\\
  p_v(b)= &\sum_aP_v^{(s-1)}(b,a).
 \end{align*}

\end{itemize}
Indeed, conditioned on the surviving neighbors, the factors involving $Z_v$ in \eqref{eq:single-tree-factor-invariant} are precisely the numerators in these kernels; 
summing over $Z_v$ gives their denominators. 
Thus $\Gamma_v$ is the joint posterior of $(Z_w,Z_v,Z_u)$ given $c$. 
Nonadjacency ensures that all required neighbors belong to $\mathcal V_s$. 
The updated posterior values use only the saved factors and the posteriors available before the reverse round. 
We form all candidates before updating the corresponding edge posteriors, and retain the existing node and unchanged edge posteriors. This proves the induction and recovers every $p_i$ after $s_0$ reverse rounds.

\paragraph{Logarithmic coordinates and initialization.}
All current factors are at most one. 
Indeed, node factors are initially at most one, and the initial edge matrices are column stochastic. 
Leaf elimination preserves this bound, while
\[
 \sum_z C_v(z,a)
 =\sum_b\psi_v(b)T_v(b,a)\sum_zT_w(z,b)\le1
\]
shows that effective edge matrices remain column substochastic. 
Every factor is a sum of products of distinct original factors. 
Retaining one compatible assignment bounds it below by $e^{-LH}$. The same lower bound holds for node, edge, and three-node posteriors: retain one compatible full hidden assignment and divide by $q^\gamma(c)\le1$. 
It also holds for each reconstruction kernel, whose numerator uses disjoint original factors and whose denominator is at most one. We therefore clip their logarithms to $[-LH,0]$; clipping cannot increase error relative to the exact quantities.

Use distinct standard basis vectors $\mathsf p_i$ for positions, $e_v$ for tokens and the mask, and $b$ for a constant. Choose the trainable embeddings as
\[
 E_{v,\cdot}=e_v^\top+b^\top,\qquad P_{i,\cdot}=\mathsf p_i^\top,
 \qquad \Embed_{E,P}(c)_{i,\cdot}=\mathsf p_i^\top+e_{c_i}^\top+b^\top.
\]
Thus $R_{\embed}=\sqrt3$, $\|E\|_F =\sqrt{2(V+1)}$ and $\|P\|_F =\sqrt{L}$, and these matrices use $2(V+1)+L$ nonzero parameters. Auxiliary coordinates are initially zero. Besides the $JV$ emission entries, allocate $O(J^2)$ coordinates for the live and saved factors, node and edge log posteriors, and receive and candidate quantities. The temporary $J^3$ triple entries and the arithmetic workspace are reused across rounds.

For completeness, the position-dependent log-probability tables can be loaded exactly by one attention head in the auxiliary coordinates. 
Let $\mathsf{t}_i$ be the vector containing the entries of $\log A_i$, $\log B_i$, and $\ind\{i=o\}\log\nu$, in separate auxiliary coordinates; 
set entries reserved for the root's edge factor to zero.
Set
\[
 Q=\sum_j\mathsf{p}_j\mathsf{p}_j^\top+bb^\top,\qquad
 K=2\log3\sum_j\mathsf{p}_j\mathsf{p}_j^\top-\log3\,bb^\top,
\]
\[
 V_{\mathrm{tab}}=2L\sum_j\mathsf{p}_j\bigg(\mathsf{t}_j-\frac1{L+2}\sum_\ell \mathsf{t}_\ell\bigg)^\top.
\]
The sigmoid attention weights are $3/4$ on the diagonal and $1/4$ elsewhere. Hence the attention output, including the factor $1/L$, equals $\mathsf{t}_i^\top$ at row $i$. A rowwise FFN then writes
\[
 \log g_i(z,c)
 =-\sum_{v\in\bV}\ReLU\big(-\log B_i(v,z)
                   +G_\delta\ind\{c_i=v\}-G_\delta\big).
\]
Adding $\log\nu$ at the root initializes $\log\psi_i^{(0)}$. These operations use $O(L(J^2+JV))$ nonzeros and FFN width $O(JV)$, including the input embeddings.

\paragraph{Routing and rowwise operations.}
For a prescribed relation $\mathcal E\subseteq[L]^2$, let
$\mathcal J(\mathcal E):=\{j:(i,j)\in\mathcal E\text{ for some }i\}$. The pair $(i,j)$ means that row $i$ receives from row $j$. For source coordinates $f_\ell$ and initially zero receive coordinates $r_\ell$, set
\begin{align}
 Q_{\mathcal E}&=\sum_{(i,j)\in\mathcal E}\mathsf p_i\mathsf{p}_j^\top+bb^\top,\quad K_{\mathcal E} =2\lambda\sum_{j\in\mathcal J(\mathcal E)}\mathsf p_j \mathsf{p}_j^\top-\lambda bb^\top,\quad V_{\mathcal E} =L\sum_\ell f_\ell r_\ell^\top.
 \label{eq:single-tree-routing}
\end{align}
The scores equal $\lambda$ on $\mathcal E$ and $-\lambda$ elsewhere. 
If source entries lie in $[-LH,0]$, each received entry differs from its prescribed sum by at most $L^2He^{-\lambda}$. 
This includes contributions from rows outside $\mathcal J(\mathcal E)$.

A positional affine map writes $d=\ind\{i\in I\}$ for any prescribed active set $I$. For $|x|\le LH$ and $d\in\{0,1\}$, the following identities implement selection, clipping, and reset by residual FFNs:
\begin{align*}
 d\cdot x&=\ReLU(x)-\ReLU(x-LHd)-\ReLU(-x)+\ReLU(-x-LHd) ,\\
 \Clip_{[-LH,0]}(x)&=-LH+\ReLU(x+LH)-\ReLU(x),\\
 -x&=-\ReLU(x)+\ReLU(-x).
\end{align*}
To update a stored value, clip the candidate first and apply the first identity to the candidate minus the old value. Inactive rows are then preserved exactly. The constant coordinate supplies affine offsets. For $m$ scalar coordinates, these operations have width and sparsity $O(m)$; the active-set indicator uses $O(|I|+1)$ nonzeros.

Let
\begin{align}
 \xi:=\frac{\varepsilon}{2^{14}L(s_0+1)},\qquad
 t:=\log(eJ/\xi),\qquad \lambda:=\log(L^2H/\xi).
 \label{eq:single-tree-parallel-accuracy}
\end{align}
By Lemmas~\ref{lem:nn-lse-approx} and~\ref{lem:nn-compilation}, a $d$-input LSE can be approximated to accuracy $\xi$ using residual rowwise FFNs with depth $Ct^2$, width $Cdt^3$, sparsity $Cdt^8$, and log Frobenius norm at most $Ct^2$.
Parallel copies use disjoint workspace coordinates. 
We reset the receive, candidate, indicator, and workspace coordinates after every stage, and leave the coordinates storing quantities for later updates unchanged during the arithmetic.

In a contraction round, rowwise FFNs compute
\[
 \log\ell_v(a)=\LSE_b\big(\log T_v(b,a)+\log\psi_v(b)\big),
\]
\[
 \log C_v(z,a)=\LSE_b\big(\log T_w(z,b)+\log\psi_v(b)+\log T_v(b,a)\big).
\]
For a selected one-child vertex, attention first copies its child's edge table. 
After computing the candidates, attention aggregates the selected leaf log messages at their current parents and sends each replacement edge table to its current child. 
The updates affect only the corresponding rows. The eliminated rows retain their saved factors throughout all later contraction rounds.

At the root, one LSE gives the log evidence, and subtraction initializes the log posterior. 
In a reverse round, attention copies the parent's node posterior to a removed leaf, or the current child's edge posterior to a removed one-child vertex. 
Affine operations form the log kernels in \eqref{eq:single-tree-restore-leaf}--\eqref{eq:single-tree-restore-triple}. Further LSEs marginalize the joint posteriors. 
The restored child-edge table is sent to its child only after all candidates have been computed. Each round uses a constant number of routing and arithmetic groups. The largest group consists of $O(J^2)$ LSEs with $J$ inputs, and therefore has width $O(J^3t^3)$ and sparsity $O(J^3t^8)$. These are shared rowwise FFNs, so their parameter count has no additional factor $L$.

\paragraph{Error control.}
Initialize auxiliary weights to the initial factors by
\[
w(\psi_o^{(0)})=2,
\qquad
w(\psi_i^{(0)})=w(T_i^{(0)})=1
\quad \text{for } i\ne o.
\]
Under elimination, define
\begin{align*}
w(\ell_v)
&=w(T_v)+w(\psi_v),\\
w(\psi_u^{\mathrm{new}})
&=w(\psi_u)+\sum_{v\in A_u}w(\ell_v),\\
w(T_w^{\mathrm{new}})
&=w(C_v)=w(T_w)+w(\psi_v)+w(T_v),
\end{align*}
where $A_u$ is the set of selected leaves with parent $u$.
Here $w$ counts the original factors $\nu,g_i$ and $A_i$ incorporated into each quantity, counting each factor by its vertex or edge of origin.
There are $2L$ such factors. Since each update combines disjoint collections, every weight is at most $2L$.

All error bounds below are uniform over $c\in\overline{\bV}^L$. 
Write $\widehat{\log\psi_i^{(s)}}$ and $\widehat{\log T_i^{(s)}}$ for their stored logarithmic
approximations. 
Initialization is exact. Each attention operation contributes at most $\xi$ of routing error and each $\LSE$ approximation contributes at most $\xi$. 
Using the $1$-Lipschitz property of $\LSE$ and induction over the contraction rounds gives
\begin{align*}
\Big\|
\widehat{\log\psi_i^{(s)}}-\log\psi_i^{(s)}
\Big\|_\infty
&\le 4\bigl(w(\psi_i^{(s)})-1\bigr)\xi
\le 8L\xi,\\
\Big\|
\widehat{\log T_i^{(s)}}-\log T_i^{(s)}
\Big\|_\infty
&\le 4\bigl(w(T_i^{(s)})-1\bigr)\xi
\le 8L\xi.
\end{align*}
Including the copying operations, all logarithmic factors stored for
reconstruction have error at most $10L\xi$.

At the root, one LSE computes $\log q^\gamma(c)$, and subtraction
gives $\log p_o$. Hence,
\begin{align*}
\big|
\widehat{\log q^\gamma(c)}-\log q^\gamma(c)
\big|
&\le (8L+1)\xi,\\
\big\|
\widehat{\log p_o}-\log p_o
\big\|_\infty
&\le (16L+1)\xi
\le 17L\xi.
\end{align*}
Let $e_s$ denote the maximum error among all stored node and edge log posteriors on $\mathcal{V}_s$, uniformly over $c\in\overline{\bV}^L$, before reversing round $s$.
Each reconstruction uses one existing posterior and at most four stored factors. Attention and
marginalization add at most $4\xi$, so
\[
e_{s_0}\le 17L\xi,
\qquad
e_{s-1}\le e_s+40L\xi+4\xi
\le e_s+50L\xi.
\]
Therefore,
\[
e_0\le 17L\xi+50Ls_0\xi
\le 70L(s_0+1)\xi.
\]

For $i\notin W(c)$, the identity
\[
\log q_i^\gamma(v\mid c)
=
\operatorname{LSE}_{z\in[J]}
\bigl(\log p_i(z)+\log B_i(v,z)\bigr)
\]
requires one further LSE. After clipping to $[-G_\delta,0]$, the
hidden-state regularization bound and the choice
$\xi=\varepsilon/[2^{14}L(s_0+1)]$ yield
\begin{align*}
\sup_{\substack{c\in\overline{\mathcal V}^{\,L}\\W(c)\ne[L]}}
\max_{\substack{i\notin W(c)\\v\in\mathcal V}}
\big|F_{\theta,i,v}(c)-\log q_i(v\mid c)\big|
\le e_0+\xi+\frac{\varepsilon}{32} \le
\Big(\frac{71}{2^{14}}+\frac{1}{32}\Big)\varepsilon
<\varepsilon.
\end{align*}
Store these logits in coordinates $o_v$ and choose
$W_{\mathrm{out}}=(o_v)_{v\in\mathcal V}$.
This matrix has $V$ nonzero entries and Frobenius norm $\sqrt V$,
proving the conditional-logit approximation bound.

Finally, copying the clipped root log context likelihood to every
row adds at most $\xi$. Thus,
\[
\sup_{c\in\overline{\mathcal V}^{\,L}}
\max_{i\in[L]}
\big|\widehat h_i(c)-\log q(c)\big|
\le (8L+2)\xi+\frac{\varepsilon}{64}
<\varepsilon,
\]
which proves the log context likelihood approximation bound.

\paragraph{Architectural budgets.}
Initialization uses $O(L(J^2+JV))$ nonzeros. Each forward or reverse round uses a constant number of routing relations with $O(L)$ pairs, together with the rowwise arithmetic described above. The final emissions use $O(JVt^8)$ nonzeros. Thus,
\begin{align*}
 N&\le C(s_0+1)t^2,\qquad M\le2,\\
 D&\le C\big[L+V+J^2+JV+(J^3+JV)t^3\big],\\
 D_{\rm ff}&\le C\big[LV+J^2+JV+(J^3+JV)t^3\big],\\
 S&\le C\big\{L(J^2+JV)+(s_0+1)(L+J^3t^8)+JVt^8\big\}.
\end{align*}
The input embeddings, output embedding, gates, and evidence broadcast are included in these bounds. Each row is eliminated once, so saving the factors introduces no factor $s_0$ in the width. Since $s_0,t\le Cb_*$ and $J^2\le J^3$, these bounds imply \eqref{eq:single-tree-parallel-budgets}. Table loading, routing, and gates have coefficients polynomial in $(L,J,V,H,\lambda)$, whereas the LSE coefficients are bounded by $\exp(Ct^2)$. Converting to matrix Frobenius norms adds only polynomial dimension factors, giving
\[
 \log B\le C\big[t^2+\log(eL^2JVH)\big]\le Cb_*^2.
\]
For each topology, the schedule fixes the routing and selection locations. All table and arithmetic locations can be allocated in advance, independently of their values. This proves the lemma.

\subsection{Approximation bound for mixtures of trees}
\label{app:tree-parallel-mixture}
\label{app:mixture-parallel-construction}
In this section, we combine the single-tree constructions to approximate $F^{\delta,q}$ for $q\in\mathcal M_{J,R}$. The component context likelihoods determine their posterior mixture weights and must be retained along with their conditional logits.

\begin{proposition}[Uniform approximation of regularized mixture logits]
\label{prop:appendix-parallel-mixture}
Fix $R\ge1$ and set
\[
 H_R:=(L+1)G_\delta+\log(64LJR/\varepsilon),\qquad
 b_R:=\log(eLJVRH_R/\varepsilon),
\]
where $G_{\delta} := \log (V/\delta)$.
There exists one parameter class $\Theta=\Theta(D,N,M,D_{\rm ff},B,S;R_{\embed},G_\delta)$ with $R_{\embed} = \sqrt{3}$ 
such that
\begin{align}
 \sup_{q\in\mathcal M_{J,R}}
 d_{\infty,\mathrm{mask}}(F^{\delta,q},\F(\Theta))\le\varepsilon,
 \label{eq:appendix-tree-uniform-realization}
\end{align}
with
\begin{align}
 N&\le Cb_R^3,\qquad M\le2R,\qquad \log B\le Cb_R^2,\notag\\
 D&\le C\big[L+V+R(J^3+JV)b_R^3\big],\qquad
 D_{\rm ff}\le C\big[LV+R(J^3+JV)b_R^3\big],\notag\\
 S&\le CR\big[L(J^2+JV)+J^3+JV\big]b_R^{10}.
 \label{eq:appendix-tree-parallel-budgets}
\end{align}
For mixtures sharing a common rooted topology, two heads suffice. The same bounds hold when the component topologies are restricted to any prescribed family of rooted trees on $[L]$.
\end{proposition}

\paragraph{Proof of Proposition~\ref{prop:appendix-parallel-mixture}.}
Fix $q=\sum_{r=1}^R\pi_rq_r\in\mathcal M_{J,R}$, allowing zero weights. By Lemma~\ref{lem:regularization},
$q^\delta=\sum_r\pi_rq_r^\delta$, and each component has emissions
$B_{i,r}^\delta(v,z)=(1-\delta)B_{i,r}(v,z)+\delta/V$.

\paragraph{Component constructions and mixture weights.}
Apply Lemma~\ref{lem:single-tree-parallel} to each $q_r^\delta$ with accuracy $a:=\varepsilon/64$. Retain its conditional logits and log evidence as internal coordinates, denoted by $\widehat f_{i,r,v}$ and $\widehat h_{i,r}$. They satisfy
\[
 |\widehat f_{i,r,v}(c)-\log q_{r,i}^\delta(v\mid c)|\le a,
 \qquad |\widehat h_{i,r}(c)-\log q_r^\delta(c)|\le a.
\]
The first bound is for masked positions, and the second holds for all contexts. The hidden-table perturbation is already included in these bounds. The component constructions occupy disjoint auxiliary coordinates, share the input embedding, and run in parallel. Shorter contraction schedules are padded with identity rounds.

To make the mixture log weights finite, set
\[
 \kappa:=\frac{\varepsilon\rho_\delta^L}{64},\qquad
 \pi_r^\kappa:=(1-\kappa)\pi_r+\frac\kappa R,
 \qquad \bar q:=\sum_{r=1}^R\pi_r^\kappa q_r^\delta.
\]
The posterior component weights are proportional to $\pi_rq_r^\delta(c)$ for $q^{\delta}$ and to $\pi_r^{\kappa} q_r^{\delta}(c)$ for $\bar q$.
A coupling of the component label gives
$\sup_E|\bar q(E)-q^\delta(E)|\le\kappa$. As both distributions have the same emission lower bound, the argument in \eqref{eq:single-tree-smoothing-error} yields
\begin{align}
 |\log\bar q_i(v\mid c)-\log q_i^\delta(v\mid c)|
 \le2\kappa\rho_\delta^{-L}=\varepsilon/32.
 \label{eq:parallel-mixture-weight-error}
\end{align}
Moreover, $|\log\pi_r^\kappa|\le\log(R/\kappa)\le H_R$.

\paragraph{Combining the components.}
For every $i \notin W(c)$ and $v\in \bV$, the conditional log probabilities of $\bar q$ satisfy
\begin{align*}
 \log\bar q_i(v\mid c)
  =\LSE_r\big(\log\pi_r^\kappa+\log q_r^\delta(c)
                         +\log q_{r,i}^\delta(v\mid c)\big) -\LSE_r\big(\log\pi_r^\kappa+\log q_r^\delta(c)\big).
\end{align*}
Use $V+1$ parallel LSE networks of accuracy $a$ to compute
\[
 \widehat h_i=\Phi_a\big((\log\pi_r^\kappa+\widehat h_{i,r})_{r=1}^R\big),
 \qquad
 \widehat v_{i,v}=\Phi_a\big((\log\pi_r^\kappa+\widehat h_{i,r}
                                      +\widehat f_{i,r,v})_{r=1}^R\big).
\]
The mixture weights enter as exact affine offsets. Since LSE is one-Lipschitz in the maximum norm, the two errors are at most $2a$ and $3a$, respectively. Subtracting and clipping to $[-G_\delta,0]$ gives error at most $5a$ relative to $\log\bar q_i(v\mid c)$. Together with \eqref{eq:parallel-mixture-weight-error}, this gives
\[
 |F_{\theta,i,v}(c)-F_{i,v}^{\delta,q}(c)|
 \le5\varepsilon/64+\varepsilon/32<\varepsilon.
\]
This explains why averaging the component conditionals with the prior weights $\pi_r$ would not suffice: the weights given $c$ are proportional to $\pi_rq_r^\delta(c)$.

\paragraph{A common architectural class.}
Let $s_0$ be the largest number of contraction rounds among the components, and take $t=C\log(eLRJ/\varepsilon)$ with a sufficiently large universal constant. The branch depths are maximized, whereas their auxiliary widths and sparsities add. Initialization requires $O(LR(J^2+JV))$ nonzeros. The component constructions and final mixture operations therefore satisfy
\begin{align*}
 N&\le C(s_0+1)t^2,\\
 S&\le CR\big\{L(J^2+JV)+(s_0+1)(L+J^3t^8)+JVt^8\big\}.
\end{align*}
The $R$ evidence broadcasts cost $O(RL)$ nonzeros, and the final mixture LSEs cost $O(RVt^8)$; both are absorbed above. Replacing the single-tree accuracy by $\varepsilon/64$ changes its logarithmic factors only by a universal constant, so $s_0,t\le Cb_R$. This proves \eqref{eq:appendix-tree-parallel-budgets}, including the Frobenius bound. At most $2R$ heads suffice. If the topologies agree, use the same contraction schedule and attention matrices, and place the component value maps in disjoint coordinates; two heads then suffice.

For fixed component topologies, the coefficient locations can be chosen independently of their probability tables. The displayed budgets are uniform over all topologies. Padding to these budgets and allowing the sparse supports to vary gives a single parameter class $\Theta$ for every $q\in\mathcal M_{J,R}$. This proves \eqref{eq:appendix-tree-uniform-realization}.

\subsection{Proof of Proposition~\ref{prop:tree-mixture-approximation}}
\label{app:pf-of-prop:tree-mixture-approx}
Let $\Theta$ be the transformer parameter class as in Proposition \ref{prop:appendix-parallel-mixture}.
\eqref{eq:appendix-tree-uniform-realization} gives
\begin{align*}
    \sup_{q\in \mathcal{M}_{J,R}} d_{\infty, \mathrm{mask}} \big(F^{\delta,q} ,\mathcal{F}(\Theta)\big) \le \varepsilon.
\end{align*}
Applying Lemma~\ref{lem:decomp} therefore gives
\[
 \varepsilon_{\mathrm{approx}}(\Theta)
 \le\frac{s_{\max}}L\Proj_{J,R}(p^\star)
     +s_{\max}\log\frac1{1-\delta}+2\varepsilon,
\]
for every distribution $p^\star$ on $\bV^L$. Together with the architectural bounds in \eqref{eq:appendix-tree-parallel-budgets},
this proves Proposition~\ref{prop:tree-mixture-approximation}.

\section{Shallow-tree result}
\label{app:shallow tree}
In this appendix, we derive complementary estimation and sampling guarantees for mixtures of hidden Markov trees with bounded height.
We replace the contraction-based construction in Appendix \ref{app:single-tree-parallel} by upward and downward message passing, while using similar techniques as in Proposition \ref{prop:gen error} and Theorem \ref{thm:end-to-end}.

Appendix \ref{app:message-identities} establishes the message identities.
Appendix \ref{app:single-tree-message} implements them for a single tree structure, and 
Appendix \ref{app:height-dependent-mixture} extends this construction to mixtures.
Appendices \ref{app:height-dep-estimation-error} and \ref{app:height-depe-end-to-end} derive the resulting estimation error and sampling guarantees.

Let $L,V\ge 2$ and $J,R\ge 1$ be integers. 
The height of a rooted tree is the maximum number of edges from its root to a vertex.
For an integer $h \in [L-1]$, let $\mathcal{M}_{J,R}^h$ denote the distributions admitting a mixture representation with $R$ $J$-state hidden Markov tree components, each of height at most $h$. 
To be more specific, let $\mathfrak{T}_h \subseteq \mathfrak{T}$ denote the subset of trees on $[L]$ of height at most $h$. 
Then 
\begin{align*}
    \mathcal{M}_{J,R}^h := \Big\{ q(x) = \sum_{r=1}^R \pi_r q_r(x) \mid q_r \text{ satisfies } \eqref{eq:hidden-tree-law} \text{ with tree structure } \mathcal{T}_r \in \mathfrak{T}_{h}, \quad \sum_{r=1}^R \pi_r = 1, \text{ and }\pi_r \in [0,1] \Big\}.
\end{align*}
Define 
\begin{align*}
    \Proj^h_{J,R} (p^\star) := \Proj_{\mathcal{M}_{J,R}^h} (p^\star) = \inf_{q\in \mathcal{M}_{J,R}^h} \KL(p^\star \| q). 
\end{align*}
For the approximation construction, take $0<\delta<1$ and $0<\varepsilon \le 1/2$, and use the notation $(\rho_{\delta},G_{\delta},b_*,H, H_R,b_R)$ from Appendices \ref{app:single-tree-parallel} and \ref{app:mixture-parallel-construction}.

\subsection{Message identities}
\label{app:message-identities}
In this section, we state the messages to help derive the target conditionals.
They provide the upward and downward updates used in the height-dependent construction.

In the following, for a rooted tree with root $o$, let $\ch(i)$ denote the set of children of vertex $i$:
\begin{align*}
     \ch(i):= \{u\in[L]\setminus \{o\}:\pa(u)=i\}, 
\end{align*}
for any $i \in [L]$.

\begin{lemma}
  \label{lem:conditional-mixture}
  Fix $c\in\overline{\bV}^L$ and write $W=W(c)$. Let $q$ be a single hidden Markov tree model with root $o$, root distribution $\nu$, transitions $A_i$, and emissions $B_i$. 
  For each vertex $i$, let $\mathcal V_i$ be the subtree rooted at $i$, including $i$, and write $\mathcal{V}_i^c:= [L]\setminus \mathcal{V}_i$. For $S\subseteq[L]$, write $c_S$ for the event $\{X_{W\cap S}=c_{W\cap S}\}$, and define
\[
 g_i(z,c):=\begin{cases}B_i(c_i,z),&i\in W,\\1,&i\notin W.\end{cases}
\]
Define the upward messages by
\begin{align}
 \alpha_i(z,c)&:=g_i(z,c)\prod_{u\in\ch(i)}m_u(z,c),\qquad i\in[L],\notag\\
 m_i(a,c)&:=\sum_{z=1}^J A_i(z,a)\alpha_i(z,c),\qquad i\ne o,
 \label{eq:message-identities-upward}
\end{align}
and the downward messages by $\beta_o(z,c):=\nu(z)$ and
\begin{align}
 \beta_i(z,c)
 :=\sum_{a=1}^J A_i(z,a)\beta_{\pa(i)}(a,c)g_{\pa(i)}(a,c)
      \prod_{u\in\ch(\pa(i))\setminus\{i\}}m_u(a,c),\qquad i\ne o.
 \label{eq:message-identities-downward}
\end{align}
Here empty products equal one. These messages satisfy
\begin{align}
 \alpha_i(z,c)\beta_i(z,c)
 &=\bP_q[Z_i=z,c],\qquad
 q(c)=\sum_{z=1}^J\alpha_i(z,c)\beta_i(z,c).
 \label{eq:message-identities-joint}
\end{align}
Consequently, whenever $q(c)>0$, for every $i\notin W$ and $v\in\bV$,
\begin{align}
 q_i(v\mid c)
 =\frac{\sum_{z=1}^J\alpha_i(z,c)\beta_i(z,c)B_i(v,z)}
        {\sum_{z=1}^J\alpha_i(z,c)\beta_i(z,c)}.
 \label{eq:message-identities-conditional}
\end{align}
For a mixture $q=\sum_{r=1}^R\pi_rq_r\in\mathcal M_{J,R}$, compute the messages on each component tree. Then, whenever $q(c)>0$, for every $i\notin W$ and $v\in\bV$,
\begin{align}
 q_i(v\mid c)
 =\frac{\sum_{r=1}^R\pi_r\sum_{z=1}^J
          \alpha_{i,r}(z,c)\beta_{i,r}(z,c)B_{i,r}(v,z)}
        {\sum_{r=1}^R\pi_r\sum_{z=1}^J
          \alpha_{i,r}(z,c)\beta_{i,r}(z,c)}.
 \label{eq:message-identities-mixture}
\end{align}

\end{lemma}

\paragraph{Proof of Lemma \ref{lem:conditional-mixture}.}
Fixing $z_i=z$ in the subtree factorization gives
\[
 \alpha_i(z,c)
 =\sum_{(z_u)_{u\in\mathcal V_i\setminus\{i\}}}
      \prod_{u\in\mathcal V_i\setminus\{i\}}A_u(z_u,z_{\pa(u)})
      \prod_{u\in\mathcal V_i}g_u(z_u,c).
\]
Each summed state ranges over $[J]$. Grouping the sum by child subtrees gives \eqref{eq:message-identities-upward}, starting with $\alpha_i=g_i$ at leaves. Thus $\alpha_i(z,c)$ is the likelihood of the subtree evidence with its root state fixed to $z$. This finite-sum definition also applies to states with $\bP_q[Z_i=z]=0$.

We next verify that $\beta_i(z,c)=\bP_q[Z_i=z,c_{\mathcal V_i^c}]$. It holds at the root since $\mathcal V_o^c=\varnothing$. For $i\ne o$, let $j=\pa(i)$. The evidence outside $\mathcal V_i$ consists of the evidence outside $\mathcal V_j$, the observation at $j$ if present, and the evidence in the sibling subtrees. The tree factorization therefore gives
\begin{align*}
 \bP_q[Z_i=z,c_{\mathcal V_i^c}]
 &=\sum_{a=1}^J A_i(z,a)\bP_q[Z_j=a,c_{\mathcal V_i^c}]\\
 &=\sum_{a=1}^J A_i(z,a)\beta_j(a,c)g_j(a,c)
            \prod_{u\in\ch(j)\setminus\{i\}}m_u(a,c)
 =\beta_i(z,c).
\end{align*}
Factoring the evidence into $\mathcal V_i$ and its complement proves \eqref{eq:message-identities-joint}. At a masked position, the emission at $i$ is absent from the context, so
\[
 \bP_q[X_i=v,c]=\sum_{z=1}^J B_i(v,z)\bP_q[Z_i=z,c].
\]
Dividing by $q(c)>0$ gives \eqref{eq:message-identities-conditional}. Finally, for the mixture,
\[
 \bP_q[X_i=v,c]=\sum_{r=1}^R\pi_r\bP_{q_r}[X_i=v,c],
 \qquad q(c)=\sum_{r=1}^R\pi_rq_r(c).
\]
Apply the single-tree joint identities to each component and take the ratio to obtain \eqref{eq:message-identities-mixture}. This argument does not require every component to assign positive probability to the context.

\subsection{Height-dependent predictor construction for single-tree}
\label{app:single-tree-message}
In this section, we implement the recursions above by processing the tree levels. We use the regularization, embedding, routing, and rowwise arithmetic from Appendix~\ref{app:single-tree-parallel}.

\begin{lemma}[Single-tree message construction]
\label{lem:single-tree-message}
Under the assumptions of Lemma~\ref{lem:single-tree-parallel}, suppose the rooted tree has height $h$.
There exists a parameter class
$\Theta=\Theta(D,N,M,D_{\rm ff},B,S;\sqrt3,G_\delta)$ and a parameter $\theta\in \Theta$ satisfying
\begin{align}
 \sup_{c:\,W(c)\ne[L]}\max_{i\notin W(c),\,v\in\bV}
 |F_{\theta,i,v}(c)-\log q_i(v\mid c)|\le\varepsilon,
 \label{eq:single-tree-goal}
\end{align}
with
\begin{align}
 N&\le C(h+1)b_*^2,\qquad M\le1,\qquad\log B\le Cb_*^2,\notag\\
 D&\le C\big[L+V+(J^2+JV)b_*^3\big],\qquad
 D_{\rm ff}\le C\big[LV+(J^2+JV)b_*^3\big],\notag\\
 S&\le CL(J^2+JV)b_*^8.
 \label{eq:single-tree-message-budgets}
\end{align}
Before the output embedding, the construction also retains a coordinate $\widehat h_i(c)$ at every row satisfying
\begin{align}
 \sup_{c\in\overline{\bV}^L}\max_{i\in[L]}
 |\widehat h_i(c)-\log q(c)|\le\varepsilon.
 \label{eq:single-tree-message-evidence}
\end{align}
The budgets depend on the topology only through its height. For a fixed topology, the permitted coefficient locations can be fixed independently of the probability tables.
\end{lemma}

\paragraph{Proof of Lemma~\ref{lem:single-tree-message}.}
Apply \eqref{eq:single-tree-smoothing} and work with $q^\gamma$. The conditional-logit and log-evidence perturbations are bounded by \eqref{eq:single-tree-smoothing-error}. As in Appendix~\ref{app:single-tree-parallel}, all relevant log probabilities lie in $[-LH,0]$. Below, the tables and messages refer to $q^\gamma$, and the superscript $\gamma$ on the hidden tables is suppressed.

Use the embedding and exact table initialization from Appendix~\ref{app:single-tree-parallel} to store $\log A_i$, $\log B_i$, $\log g_i$, and $\log\nu$ at the root. Allocate separate coordinates for the upward and downward log messages, received values, candidates, and LSE workspace. Initialize $\widehat\beta_o^{\log}=\log\nu$, with other message and temporary coordinates zero. Here the superscript $\log$ indicates a stored approximation to a log message. Set
\[
 \eta:=\frac{\varepsilon}{128L(h+1)},\qquad
 t:=\log(eJ/\eta),\qquad \lambda:=\log(L^2H/\eta).
\]
The attention maps in \eqref{eq:single-tree-routing} have error at most $\eta$ per received coordinate. We use the exact clipping, selection, and reset operations from Appendix~\ref{app:single-tree-parallel}, and LSE networks of accuracy $\eta$ from Lemma~\ref{lem:nn-lse-approx}.

\paragraph{Upward pass.}
Let $h_i$ be the height of the subtree rooted at $i$, and process $I_s:=\{i:h_i=s\}$ for $s=0,\ldots,h$. Every child of an active vertex has already been processed. Using the relation
\[
 \mathcal E_s^\uparrow:=\{(i,u):i\in I_s,\ u\in\ch(i)\},
\]
attention aggregates the stored child log messages into initially zero receive coordinates $\widehat r_i$. Clip these coordinates to $[-LH,0]$. Rowwise FFNs then form the candidates
\begin{align}
 \widehat\alpha_i^{\log}(z)
 &\leftarrow\Clip_{[-LH,0]}\big(\log g_i(z,c)+\widehat r_i(z)\big),
 \label{eq:up-alpha-candidate}\\
 \widehat m_i^{\log}(a)
 &\leftarrow\Clip_{[-LH,0]}\Big(\Phi_\eta\big(
      (\log A_i(z,a)+\widehat\alpha_i^{\log}(z))_{z=1}^J\big)\Big),
      \qquad i\ne o.
 \label{eq:up-m-candidate}
\end{align}
The right-hand side of the second update uses the new candidate for $\widehat\alpha_i^{\log}$. We select the active rows only after forming the candidates and preserve the other stored messages. The root needs only the first update. At the end of each stage, reset all receive, candidate, indicator, and workspace coordinates.

Let $n_i:=|\mathcal V_i|$. By \eqref{eq:message-identities-upward}, induction from the leaves gives
\begin{align}
 \|\widehat\alpha_i^{\log}-\log\alpha_i\|_\infty
 &\le\eta+\sum_{u\in\ch(i)}2n_u\eta=(2n_i-1)\eta,\notag\\
 \|\widehat m_i^{\log}-\log m_i\|_\infty&\le2n_i\eta \qquad i \ne o.
 \label{eq:single-tree-up-error}
\end{align}
Indeed, routing contributes at most $\eta$, and LSE contributes another $\eta$. Adding exact local evidence and clipping do not increase these bounds.

\paragraph{Downward pass.}
Process nonroot vertices by increasing depth. At the beginning of each stage, form
\begin{align}
 \widehat u_i(z):=\Clip_{[-LH,0]}
       \big(\widehat\alpha_i^{\log}(z)+\widehat\beta_i^{\log}(z)\big).
 \label{eq:combined-coordinate}
\end{align}
For $D_s:=\{i:\mathrm{depth}(i)=s\}$ for $s=1,\ldots,h$, use
\[
 \mathcal E_s^\downarrow:=\{(i,\pa(i)):i\in D_s\}
\]
to copy the parent's combined log message. Denote the clipped received vector by $\widehat r_i$ and set
\[
 \widehat d_i(a):=\Clip_{[-LH,0]}
                     \big(\widehat r_i(a)-\widehat m_i^{\log}(a)\big).
\]
For $j=\pa(i)$, its exact target is
\begin{align*}
 \log\alpha_j(a,c)+\log\beta_j(a,c)-\log m_i(a,c)
 &=\log\beta_j(a,c)+\log g_j(a,c)
          +\sum_{u\in\ch(j)\setminus\{i\}}\log m_u(a,c)\\
 &=\log\bP_{q^\gamma}[Z_j=a,c_{\mathcal V_i^c}].
\end{align*}
Thus subtraction removes the evidence from $\mathcal V_i$. The downward recursion \eqref{eq:message-identities-downward} is implemented by
\begin{align}
 \widehat\beta_i^{\log}(z)
 \leftarrow\Clip_{[-LH,0]}\Big(\Phi_\eta\big(
       (\log A_i(z,a)+\widehat d_i(a))_{a=1}^J\big)\Big).
 \label{eq:down-candidate}
\end{align}
As before, form candidates, select the active rows, and reset temporary coordinates. All parent downward messages are complete before they are used. If
$e_i:=\|\widehat\beta_i^{\log}-\log\beta_i\|_\infty$, then
\begin{align}
 e_o=0,\qquad e_i\le e_{\pa(i)}+(4L+2)\eta
       \le\mathrm{depth}(i)(4L+2)\eta.
 \label{eq:single-tree-down-error}
\end{align}
The two upward-message errors contribute at most $4L\eta$, while routing and LSE contribute at most $2\eta$.

\paragraph{Conditional logits and context likelihood.}
With both passes complete, form $\widehat u_i$ again as in \eqref{eq:combined-coordinate}. By \eqref{eq:message-identities-joint}, its target is $\log\bP_{q^\gamma}[Z_i=z,c]$. Apply $V+1$ parallel LSE networks to obtain
\[
 \widehat h_i:=\Clip_{[-LH,0]}\Big(\Phi_\eta\big((\widehat u_i(z))_{z=1}^J\big)\Big),
 \qquad
 \widehat v_{i,v}:=\Phi_\eta\big((\widehat u_i(z)+\log B_i(v,z))_{z=1}^J\big).
\]
Their targets are $\log q^\gamma(c)$ and $\log\bP_{q^\gamma}[X_i=v,c]$, respectively; the second identity is needed only when $i$ is masked. Each error is at most
\[
 \big[2L+h(4L+2)+1\big]\eta\le8L(h+1)\eta.
\]
Clip $\widehat v_{i,v}-\widehat h_i$ to $[-G_\delta,0]$, store the result in output coordinates $o_v$, and use $W_{\rm out}=(o_v)_{v\in\bV}$. This gives conditional-logit error at most $16L(h+1)\eta=\varepsilon/8$ relative to $q^\gamma$. Combining this with \eqref{eq:single-tree-smoothing-error} proves \eqref{eq:single-tree-goal}. The retained coordinate $\widehat h_i$ has error at most $\varepsilon/16+\varepsilon/64$ relative to $\log q(c)$, proving \eqref{eq:single-tree-message-evidence} for all contexts.

\paragraph{Architectural budgets.}
Each message stage uses $J$ parallel $J$-input LSE networks, with depth $Ct^2$, arithmetic width $CJ^2t^3$, and arithmetic sparsity $CJ^2t^8$. For positional routing, every edge and source child is used once in the upward pass. In the downward pass, every edge is used once and each source parent is used at only one depth. The active sets partition the processed vertices, so
\[
 \sum_s\big(|\mathcal E_s|+|\mathcal J(\mathcal E_s)|+|I_s|+1\big)\le CL
\]
for the upward pass, with $I_s$ replaced by $D_s$ for the downward pass. Here $\mathcal J$ is defined before \eqref{eq:single-tree-routing}. Thus positional routing and indicators cost $O(L)$ over each pass. Value routing and resets use $O(J^2)$ coefficients per stage and are absorbed in the arithmetic count.

Adding initialization, both passes, and the output group yields
\begin{align*}
 N&\le C(h+1)t^2,\\
 S&\le C\big\{L(J^2+JV)+(h+1)J^2t^8+JVt^8\big\}
   \le CL(J^2+JV)t^8.
\end{align*}
Here we use only $h\le L-1$. Persistent coordinates have dimension $O(L+V+J^2+JV)$, and workspace is reused across stages. The input embeddings and output matrix are counted as in Appendix~\ref{app:single-tree-parallel}; retaining $\widehat h_i$ adds no routing step. The same coefficient bounds give
\[
 \log B\le C\big[t^2+\log(eL^2JVH)\big]\le Cb_*^2.
\]
Since $t\le Cb_*$, these estimates give \eqref{eq:single-tree-message-budgets}. The topology fixes the routing and active sets, and all table and arithmetic locations can be allocated independently of their values.

\subsection{Approximation bound for mixtures of trees}
\label{app:height-dependent-mixture}
In this section, we give and prove the height-dependent mixture result of Proposition~\ref{prop:appendix-message-mixture}. We replace the component construction in Appendix~\ref{app:tree-parallel-mixture} by Lemma~\ref{lem:single-tree-message}; the mixture-weight perturbation and final LSE operation are unchanged.

\begin{proposition}[Height-dependent approximation of regularized mixture logits]
\label{prop:appendix-message-mixture}
Let $1\le h\le L-1$. With $H_R$ and $b_R$ as in Proposition~\ref{prop:appendix-parallel-mixture}, there exists one class
$\Theta=\Theta(D,N,M,D_{\rm ff},B,S;\sqrt3,G_\delta)$ such that
\begin{align}
 \sup_{q\in\mathcal M_{J,R}^h}
 d_{\infty,\mathrm{mask}}(F^{\delta,q},\F(\Theta))\le\varepsilon,
 \label{eq:height-mixture-uniform-realization}
\end{align}
with
\begin{align}
 N&\le C(h+1)b_R^2,\qquad M\le R,\qquad\log B\le Cb_R^2,\notag\\
 D&\le C\big[L+V+R(J^2+JV)b_R^3\big],\qquad
 D_{\rm ff}\le C\big[LV+R(J^2+JV)b_R^3\big],\notag\\
 S&\le CRL(J^2+JV)b_R^8.
 \label{eq:height-mixture-budgets}
\end{align}
If all component trees share a common rooted topology, one head suffices. The same bounds hold for any further restriction on the component topologies.
\end{proposition}

\paragraph{Proof of Proposition~\ref{prop:appendix-message-mixture}.}
Fix a representation $q=\sum_{r=1}^R\pi_rq_r\in\mathcal M_{J,R}^h$ with component heights at most $h$. Regularize the emissions as in Lemma~\ref{lem:regularization} and apply Lemma~\ref{lem:single-tree-message} to each $q_r^\delta$ with accuracy $a:=\varepsilon/64$. The resulting internal coordinates satisfy
\[
 |\widehat f_{i,r,v}(c)-\log q_{r,i}^\delta(v\mid c)|\le a,
 \qquad |\widehat h_{i,r}(c)-\log q_r^\delta(c)|\le a,
\]
where the first bound is for masked positions and the second holds for every context. These estimates already include the hidden-table perturbation. The branches occupy disjoint auxiliary coordinates and share the input embedding. Run the upward stages for all components, followed by their downward stages, with inactive stages padded by identities.

Use the weights $\pi_r^\kappa$ and the final $V+1$ LSE networks from Appendix~\ref{app:tree-parallel-mixture}. That calculation uses only the two displayed approximation bounds. The two LSE errors are at most $2a$ and $3a$, and the weight perturbation contributes at most $\varepsilon/32$ by \eqref{eq:parallel-mixture-weight-error}. Thus,
\[
 \sup_{c:\,W(c)\ne[L]}\max_{i\notin W(c),\,v\in\bV}
 |F_{\theta,i,v}(c)-F_{i,v}^{\delta,q}(c)|
 \le5a+\varepsilon/32<\varepsilon.
\]

It remains to count the architecture. Take $t=C\log(eLRJ/\varepsilon)$ with a sufficiently large universal constant. Each message stage has depth $Ct^2$, total arithmetic width $CRJ^2t^3$, and arithmetic sparsity $CRJ^2t^8$. The positional routing count in Lemma~\ref{lem:single-tree-message} gives $O(RL)$ nonzeros over each complete pass. Initialization costs $O(LR(J^2+JV))$, the component output groups cost $O(RJVt^8)$, and the final mixture LSEs cost $O(RVt^8)$. Therefore,
\begin{align*}
 N&\le C(h+1)t^2,\\
 S&\le CR\big\{L(J^2+JV)+(h+1)J^2t^8+JVt^8\big\}
   \le CRL(J^2+JV)t^8.
\end{align*}
Widths and sparsities add across components, while their depths are maximized. All workspace is reused across stages. Since $t\le Cb_R$, the width and Frobenius estimates from Lemma~\ref{lem:single-tree-message} give \eqref{eq:height-mixture-budgets}. There is at most one routing head per component at each step. If the rooted topologies coincide, their routing matrices are shared, and the value maps can be combined into one head.

These budgets are uniform over all component trees of height at most $h$. For fixed topologies, the permitted coefficient locations are independent of the tables. Padding to the displayed budgets and allowing the supports to vary gives one class $\Theta$ satisfying \eqref{eq:height-mixture-uniform-realization}. 
This proves the proposition.
Finally, Lemma~\ref{lem:decomp} gives the corresponding approximation bound
\[
 \varepsilon_{\mathsf{approx}}(\Theta)
 \le\frac{s_{\max}}L\Proj_{\mathcal M_{J,R}^h}(p^\star)
     +s_{\max}\log\frac1{1-\delta}+2\varepsilon.
\]
The height bound enters the transformer depth through $h+1$. No logarithmic-height assumption is used.

\subsection{Estimation error analysis}
\label{app:height-dep-estimation-error}

\begin{theorem}[Estimation error for shallow tree]
\label{thm:shallow-tree-mixture-estimation}
Let $n,L,V\ge 3$, $J,R \ge 1$, and $h\in [L-1]$. 
Use the fixed-cardinality masking schedule $s$ in Section \ref{sec:prelim} for training.
There exists a transformer class $\Theta_h$ of the form \eqref{eq:transformer predictor}, with an architecture depending only on $(n,L,V,J,R,h)$ (specified in Proposition \ref{prop:appendix-message-mixture} with $\delta = \varepsilon = 1/n$), such that for any distribution $p^\star$ on $\bV^L$, the ERM $\widehat{\theta}_h$ over $\Theta_h$ computed from $n$ i.i.d.~samples drawn from $p^\star$ satisfies
\begin{align}
    \bE[\mathcal{L}(\widehat{\theta}_h)] \leq C_{3} \bigg(\frac{s_{\mathsf{max}}}{L} \Proj^h_{J,R}(p^\star) + \frac{K_{\rm sh}}{n}\, \polylog(n,L,V,J,R) \bigg).  \label{eq:shallow tree estimation error}
\end{align}
Here, $C_{3}>0$ is a universal constant, the expectation is over training samples, and 
\begin{align}\label{eq:Ksh}
    K_{\rm sh} := RL(J^2+JV)h^2.
\end{align}
\end{theorem}

\paragraph{Proof of Theorem \ref{thm:shallow-tree-mixture-estimation}.}
Recall $\beta:= \log (enLVJR)$. 
Lemma~\ref{lem:decomp} gives
\[
 \varepsilon_{\mathsf{approx}}(\Theta_h)
 \le\frac{s_{\max}}L\Proj_{\mathcal M_{J,R}^h}(p^\star)
      +\frac{2s_{\max}+2}{n}.
\]
The height-dependent budgets satisfy
\[
 N\le C(h+1)\beta^2,\qquad
 S\le CRL(J^2+JV)\beta^8,\qquad
 \log B\le C\beta^2,\qquad \log P_{\rm tr}\le C\beta.
\]
Applying Proposition~\ref{prop:gen error} with 
$$ \log A_{\mathrm{TF}} \lesssim h^2\beta^6, \quad C(\Theta_h) \lesssim K_{\rm sh} \beta^{14}$$
yields 
\begin{align}
 \bE\big[\loss(\widehat\theta_h)\big]
 \le C\bigg\{\frac{s_{\max}}L\Proj_{J,R}^h(p^\star)
       +\frac{K_{\mathrm{sh}}\beta^{16}}{n}\bigg\} . \label{eq:score estimation of bounded height}
\end{align}
This proves Theorem \ref{thm:shallow-tree-mixture-estimation}.

\subsection{End-to-end sampling guarantees}
\label{app:height-depe-end-to-end}

\begin{theorem}[Sampling guarantees under shallow tree structure]
\label{thm:shallow end-to-end}
Assume the conditions and adopt the notation of Theorem \ref{thm:shallow-tree-mixture-estimation}.
Let $p_{Y\mid S}$ denote the output distribution of the $T$-step MDM sampler using the ERM-based mask predictor $p_{\widehat{\theta}_h}$ and the realized unmasking sets $\mathcal{S}=(S^1,\ldots,S^T)$ with $|S^t| = s_t$.
Then for any $1\leq T \leq L$, one has 
\begin{align}
    & \bE_{\{X^{(i)}\}_{i=1}^n,S} \big[\KL \big( p^\star \,\|\, p_{Y\mid S}\big)\big] \notag \\
    &\quad \leq C_{4}\bigg( \varepsilon_{\mathsf{parall}}(p^\star;s) + s_{\mathsf{max}} \Proj^h_{J,R}(p^\star) + \frac{L K_{\rm sh}}{n}\, \polylog (n,L,V,J,R)\bigg), \label{eq:shallow tree sampling error}
\end{align}
where $C_{4}>0$ is a universal constant, and the expectation is over the training samples and random unmasking sets $S$. 
\end{theorem}

\paragraph{Proof of Theorem \ref{thm:shallow end-to-end}.}
For $1\le h\le L-1$, use the parameter class $\Theta_h$ from Proposition~\ref{prop:appendix-message-mixture} with $\delta = \varepsilon = 1/n$. 
Denote $\widehat{\theta}_h $ as the ERM estimator over $\Theta_h$, and let $p_{Y\mid S}$ be the output distribution of the sampler using $p_{\widehat{\theta}_h}$ as mask predictor and $S=(S^1,\ldots,S^T)$ as the unmasking partition.
Then applying the decomposition \eqref{eq:sampling-decomposition} with $\theta = \widehat{\theta}_h$, 
\begin{align*}
    \bE_{\{X^{(j)}\}, S}\big[
        \KL(p^\star\|p_{Y\mid S})
    \big]
    &=\varepsilon_{\parall}(p^\star; s)
      +L\bE_{ \{X^{(j)}\}}\big[\mathcal L(\widehat\theta_h)\big]\\
    &\lesssim \varepsilon_{\parall}(p^\star; s) + s_{\max}\Proj_{J,R}^h(p^\star) + \frac{LK_{\rm sh}}{n} \polylog(n,L,V,J,R). 
\end{align*}
Here we apply \eqref{eq:score estimation of bounded height} and $K_{\rm sh}:= RL (J^2+JV)h^2$.

\section{Auxiliary neural-network approximation lemmas}
\label{app:nn}

We write $\NN(W,d,B,S)$ for feed-forward ReLU networks of maximum width at most $W$, depth at most $d$, coefficient magnitude at most $B$, and at most $S$ nonzero scalar coefficients: 
\begin{align}
 \NN (W,d,B,S) := \Big\{ \phi(x) &=(A_d \, \sigma + b_d ) \circ \cdots \circ (A_1 I + b_1 )(x) \notag \\
 & \big |\,  A_i \in \bR^{W_{i+1}\times W_{i}}, 
 b_i \in \bR^{W_{i+1}}, 
 \max_i \big\{ \|A_i\|_{\infty,\infty} \vee \|b_i\|_{\infty} \big\} \leq B,  \notag \\
 &
 \sum_{i=1}^d \big(\|A_i\|_{0} + \|b_i\|_0 \big) \leq S
 \Big\}
\end{align}
Here $I$ denotes the identity map, $\sigma$ is applied entry-wise using ${\rm ReLU} (x) = x \vee 0$, and $W_i \leq W$, $\forall i\in [d+1]$. 
As a remark, in the auxiliary neural-network class, \(B\) bounds individual weights and biases, while in the transformer class defined in (\ref{eq:transformer predictor}), $B$ bounds the Frobenius norm of each parameter matrix. 
We will convert them when applying results to transformer structure in Lemma \ref{lem:nn-compilation}. 

\subsection{Exact operations and residual implementation}
\begin{lemma}[Exact realization of clipping]
\label{lem:nn-clipping}
For $a\le b$, define
\[
 \operatorname{Clip}_{[a,b]}(x):=\min\{b,\max\{a,x\}\}.
\]
This function is represented exactly by a ReLU network in
$\NN(2,2,B_{\rm clip},7)$, where
$B_{\rm clip}:=\max\{1,|a|,|b|\}$.
Moreover, for all $x,x'\in\bR$ and $y\in[a,b]$,
\[
 |\operatorname{Clip}_{[a,b]}(x)-\operatorname{Clip}_{[a,b]}(x')|
 \le |x-x'|,
 \quad
 |\operatorname{Clip}_{[a,b]}(x)-y|\le |x-y|.
\]
\end{lemma}
\paragraph{Proof of Lemma \ref{lem:nn-clipping}.}
The identity
\[
 \operatorname{Clip}_{[a,b]}(x)
 =a+\relu(x-a)-\relu(x-b)
\]
gives an exact representation with two hidden units, two affine layers,
and at most seven nonzero weights and biases, each of magnitude at most
$B_{\rm clip}$. The clipping function is continuous and piecewise affine
with slopes in $\{0,1\}$, hence is $1$-Lipschitz.
Since $\operatorname{Clip}_{[a,b]}(y)=y$ for $y\in[a,b]$,
the second inequality follows from the first.

\begin{lemma}[Implementation of ReLU DNNs by transformer FFN blocks]
\label{lem:nn-compilation}
Suppose a feed-forward ReLU network has maximum layer dimension $W$, depth $d$, coefficient bound $B\ge1$, and sparsity $S$. 
Using a preserved constant coordinate and two reusable buffers, the function can be implemented rowwise by $O(d+1)$ residual FFN blocks as defined in \eqref{eq:single transformer block} with zero attention updates, auxiliary dimension $O(W)$, FFN width $O(W)$, matrix Frobenius norms at most $C(W+1)(B+1)$, and total sparsity
\[
 C\big(S+(d+1)W\big).
\]
For parallel realization, the sparsities and buffer dimensions add; the squared Frobenius norms add across the branches.
\end{lemma}

\paragraph{Proof of Lemma \ref{lem:nn-compilation}.}
Write the ReLU network as
\[
z_0=x,\quad
z_\ell=\relu(A_\ell z_{\ell-1}+b_\ell)
\quad (1\le\ell<d), \quad z_d = A_d z_{d-1}+b_d. 
\]
and pad all layer vectors to dimension $W$.
Set all attention updates to zero, and reserve two groups
of $W$ auxiliary coordinates together with a constant coordinate.

Suppose the current row is $h=(z^\top,v^\top,1)$, where
$z$ contains the current NN input and $v$ contains the old
values in the destination coordinates. To implement
$z\mapsto\relu(Az+b)$, choose
\[
W_{\rm TF}=
\begin{pmatrix}
A^\top&0&0\\
0&I_W&-I_W\\
b^\top&0&0
\end{pmatrix},
\qquad
U_{\rm TF}=
\begin{pmatrix}
0&I_W&0\\
0&-I_W&0\\
0&I_W&0
\end{pmatrix}.
\]
Then the residual FFN satisfies
\begin{align*}
h+\relu(hW_{\rm TF})U_{\rm TF}
&=\big(z^\top,\,
[v+\relu(Az+b)-\relu(v)+\relu(-v)]^\top,\,1\big)\\
&=\big(z^\top,\relu(Az+b)^\top,1\big).
\end{align*}
Thus one block writes the next hidden layer exactly.
Alternating the source and destination groups implements
successive layers:
\[
(z_0,0)\longmapsto(z_0,z_1)
\longmapsto(z_2,z_1)\longmapsto(z_2,z_3)
\longmapsto\cdots.
\]

For the final affine layer, take $r=Az+b$ and use four
groups of hidden activations to produce
\[
\Delta v
=\relu(r)-\relu(-r)-\relu(v)+\relu(-v)
=r-v.
\]
The residual update therefore gives $v+\Delta v=Az+b$.

To count parameters, let
$s_\ell=\|A_\ell\|_0+\|b_\ell\|_0$.
For each hidden layer, the displayed matrices have
\[
\|W_{\rm{TF},\ell}\|_0=s_\ell+2W,\quad
\|U_{\rm{TF},\ell}\|_0=3W,
\]
while the final affine layer uses at most $2s_d+6W$
nonzero coefficients. Including the initial input copy
and final coordinate selection yields
\[
S_{\rm TF}\le C\big(S+(d+1)W\big).
\]
The construction uses $O(d+1)$ blocks, auxiliary
dimension $O(W)$, and FFN width at most $4W$.
Moreover, for each hidden-layer block,
\[
\|W_{\rm{TF},\ell}\|_F^2
=\|A_\ell\|_F^2+\|b_\ell\|_2^2+2W
\le (W^2+W)B^2+2W,
\quad
\|U_{\rm{TF},\ell}\|_F^2=3W.
\]
The final block satisfies the same bounds up to constants,
so every FFN matrix has Frobenius norm at most
$C(W+1)(B+1)$.
For parallel networks, use disjoint auxiliary and FFN
coordinates; sparsities, auxiliary dimensions, and
squared Frobenius norms then add across branches.

\subsection{Scalar arithmetic approximations}
\begin{lemma}[Multiplication on a bounded rectangle]
\label{lem:nn-product}
For $C_0\geq 1$ and $0<\epsilon\le1/2$, there exists $\phi_{\rm prod}\in\NN(W,d,B,S)$ such that
\[
 \sup_{|x|,|y|\le C_0}| \phi_{\rm prod}(x,y)-xy|\le4C_0^2\epsilon,
\]
where $W=O(1)$, $d=O(\log(e/\epsilon))$, $B=C_0^{2}$, and $S=O(\log(e/\epsilon))$.
\end{lemma}
This lemma is a direct corollary of Lemma F.6 in \citet{oko2023diffusion} with two inputs, followed by rescaling.

\begin{lemma}[Uniform approximation of the exponential function (Lemma F.12 in the arXiv version of \citet{oko2023diffusion})]
\label{lem:nn-exponential}
For $0<\epsilon\le1/2$, there exists $\phi_{\exp}\in \NN (W,d,B,S)$ with
\[
 W=O(\log \epsilon ^{-1}),\quad d=O(\log^2 \epsilon ^{-1}),\quad
 B=\exp\big(O(\log^2 \epsilon^{-1} )\big),\quad
 S=O(\log^2 \epsilon^{-1} ),
\]
such that for any $x,x'\ge0$,
\[
 | e^{-x'}-\phi_{\exp}(x) |\le\epsilon+|x-x'|.
\]
Moreover, $|\phi_{\exp}(x)|\le\epsilon$ whenever $x\ge\log 3\epsilon^{-1}$.
\end{lemma}

\begin{lemma}[Uniform approximation of logarithm on the interval]
\label{lem:nn-logarithm}
For $0<\tau\le1/2$ and $0<\epsilon\le1/2$, there exists $\phi_{\log}\in\NN(W,d,B,S)$ with
\begin{align*}
 W&=O\big(\log^2 \epsilon^{-1}\cdot \log \tau^{-1}\big),&
    d&=O\big(\log^2 \epsilon^{-1} \vee\log \tau^{-1} \big),\\
 B&=O\big(\tau^{-1}\vee\log^2 \tau^{-1}\big),&
 S&=O\big((\log \epsilon^{-1}+\log \tau^{-1} )^8\big),
\end{align*}
such that $\sup_{x\in[\tau,1]}|\phi_{\log}(x)-\log x|\le\epsilon$.
\end{lemma}
This result follows from Lemma C.9 in \citet{wu2026neural} by using $S\lesssim d(W+1)^2$ to control the network sparsity. 

\subsection{Softmax stability and LSE approximation}
\begin{lemma}[Global softmax stability]
\label{lem:nn-softmax-stability}
For any $a,b\in\bR^k$,
\[
 \|\softmax(a)-\softmax(b)\|_1\le2\|a-b\|_\infty.
\]
\end{lemma}
\paragraph{Proof of Lemma \ref{lem:nn-softmax-stability}.}
Let $h=a-b$, $z(t)=b+th$, and $p(t)=\softmax(z(t))$ for $0\le t\le1$. Differentiating with respect to $t$ gives
\[
 p'_i(t)=p_i(t)\Big(h_i-\sum_s p_s(t)h_s\Big), \quad \forall i \in [k].
\]
Observe that $p_i(t) \in (0,1]$ together with $\sum_i p_i(t)=1$, and thus
\[
 \sum_{i=1}^k|p'_i(t)| \leq  2\|h\|_{\infty}.
\]
Integrating over $t\in[0,1]$ then proves the result.

\begin{lemma}[Uniform log-sum-exp approximation]\label{lem:nn-lse-approx}
For $n\ge1$ and $0<\xi\le1/2$, there exists a ReLU DNN $\Phi_\xi \in \NN(W,d,B,S)$ such that
\[
 \sup_{x\in\bR^n}|\Phi_\xi(x)-\LSE(x)|\le\xi,
 \quad \text{where }\LSE(x):=\log\sum_{j=1}^n e^{x_j}.
\]
With $t=\log(en/\xi)$ and some universal constant $C>0$, its network parameters can be chosen as follows,
\[
 d\le Ct^2,\quad W\le Cnt^3,\quad
 S\le Cnt^8,\quad \log B \le Ct^2.
\]
It can be implemented by residual rowwise FFNs with the same orders of depth, width, and sparsity, using reusable auxiliary coordinates and a constant input coordinate. The corresponding matrix Frobenius norms also have logarithms at most $Ct^2$. Moreover, $\LSE$ is $1$-Lipschitz with respect to the infinity norm:
\begin{align}
    |\LSE(x)-\LSE(y)|\le\|x-y\|_\infty. \label{eq:nn-lse-con}
\end{align}
\end{lemma}
\paragraph{Proof of Lemma \ref{lem:nn-lse-approx}.}
For $n=1$, we just take $\Phi_\xi(x)=x_1$. In the following, assume that $n\ge2$ and denote
\[
 \nu:=\frac{\xi}{4n},\quad \eta:=\frac{\xi}{2},
 \quad \text{and }t:=\log(\frac{en}{\xi}).
\]
We construct the approximation by composing the exponential and logarithm networks in Lemmas~\ref{lem:nn-exponential} and~\ref{lem:nn-logarithm} respectively, together with exact maximum, affine, and clipping operations.

\paragraph{Step 1: exact maximum shift.}
For $x\in\bR^n$, define
\[
 a(x):=\max_{j\in[n]}x_j,\quad
 u_j(x):=a(x)-x_j,\quad
 Z(x):=\sum_{j=1}^n e^{-u_j(x)}.
\]
Then $u_j(x)\ge0$, at least one $u_j(x)$ equals zero, and
\begin{equation}\label{eq:nn-lse-shift}
 1\le Z(x)\le n,\quad
 \LSE(x)
 =\log\Big(e^{a(x)}\sum_{j=1}^n e^{x_j-a(x)}\Big)
 =a(x)+\log Z(x).
\end{equation}
Apply the following pairwise maximum operation along a balanced binary tree, with comparisons at each level performed in parallel:
\[
 \max\{v,w\}=\relu(v-w)+w. 
\]
Retaining the original coordinates until the final exact realization of $a(x)$ gives the map
\[
 \Gamma(x):=\big(a(x),u_1(x),\ldots,u_n(x)\big),
\]
realized by ReLU DNNs with depth $O(\log(en))$, width $O(n)$, sparsity $O(n\log(en))$, and coefficient bound $O(1)$. 
Retaining original coordinates can be realized via ReLU layers using $v=\relu(v)-\relu(-v)$.

\paragraph{Step 2: approximate the exponential sum.}
Apply Lemma~\ref{lem:nn-exponential} with accuracy $\nu$ to obtain
\[
 \phi_{\exp,\nu}\in\NN\, (W_{\exp},d_{\exp},B_{\exp},S_{\exp}),
 \quad\text{such that }
 \sup_{u\ge0}|\phi_{\exp,\nu}(u)-e^{-u}|\le\nu,
\]
where
\begin{align*}
 W_{\exp}\le Ct,\quad d_{\exp}\le Ct^2,\quad 
 S_{\exp}\le Ct^2,\quad \log B_{\exp}&\le Ct^2.
\end{align*}
Run $n$ copies in parallel and set the approximation of $Z(x)$: 
\[
 \widehat Z(x)
 :=\Clip_{[1,n]}\Big(\sum_{j=1}^n \phi_{\exp,\nu}(u_j(x))\Big).
\]
Clipping is exact by Lemma~\ref{lem:nn-clipping}. Since $Z(x)\in[1,n]$, the clipping operator does not introduce additional error. 
Hence, 
\begin{align}
 |\widehat Z(x)-Z(x)|
 &\le\Big|\sum_{j=1}^n \phi_{\exp,\nu}(u_j(x))
                  -\sum_{j=1}^n e^{-u_j(x)}\Big|\notag\\
 &\le\sum_{j=1}^n|\phi_{\exp,\nu}(u_j(x))-e^{-u_j(x)}|
 \le n\nu=\frac{\xi}{4},
 \quad \text{with }\widehat Z(x)\in[1,n].\label{eq:nn-lse-sum-error}
\end{align}
The parallel exponential networks have width $O(nt)$, depth $O(t^2)$, and sparsity $O(nt^2)$. Summation and clipping add $O(n)$ coefficients, of magnitude at most $Cn$.

\paragraph{Step 3: rescale the logarithm network and compose.}
Apply Lemma~\ref{lem:nn-logarithm} with interval lower bound $\tau=1/n$ and accuracy $\eta$, and hence there exists 
\[
 \phi_{\log, \eta}\in\NN\, (W_{\log},d_{\log},B_{\log},S_{\log}),
 \quad \text{such that}
 \sup_{v\in[1/n,1]}|\phi_{\log, \eta}(v)-\log v|\le\eta,
\]
where
\begin{align*}
 W_{\log}
 &\le C\log^2(2/\xi)\log n\le Ct^3,&
 d_{\log}
 &\le C\big(\log^2(2/\xi)\vee\log n\big)\le Ct^2,\\
 B_{\log}
 &\le C\big(n\vee\log^2 (n)\big)\le Cn,&
 S_{\log}
 &\le C\big(\log(2/\xi)+\log n\big)^8\le Ct^8.
\end{align*}
To transfer the logarithm approximation from $[1/n,1]$ to $[1,n]$, define the affine maps $A_n(z):=z/n$ and $T_n(v):=v+\log n$. Denote the composition
\[
 \Lambda_\eta:=T_n\circ  \phi_{\log, \eta} \circ A_n.
\]
It then satisfies
\[
 \sup_{z\in[1,n]}|\Lambda_\eta(z)-\log z|
 =\sup_{z\in[1,n]}| \phi_{\log, \eta}(z/n)- \log(z/n)|
 \le\eta.
\]
Consequently, we denote the network construction based on \eqref{eq:nn-lse-shift} as
\begin{equation}\label{eq:nn-lse-network}
 \Phi_\xi(x)
 :=a(x)+\Lambda_\eta\big(\widehat Z(x)\big)
 =a(x)+\log n+
 \phi_{\log,\eta}\Big(\frac{1}{n}
 \Clip_{[1,n]}\Big(\sum_{j=1}^n
 \phi_{\exp,\nu}\big(a(x)-x_j\big)\Big)\Big).
\end{equation}

Using \eqref{eq:nn-lse-shift}--\eqref{eq:nn-lse-sum-error} and the $1$-Lipschitz property of $\log x$ on $[1,n]$, we obtain that, for every $x\in\bR^n$,
\begin{align*}
 |\Phi_\xi(x)-\LSE(x)|
 &\le \big|\Lambda_\eta\big(\widehat{Z}(x)\big)-\log \widehat{Z}(x) \big|
       +|\log\widehat Z(x)-\log Z(x)|\\
 &\le\eta+|\widehat Z(x)-Z(x)|
 \le\frac{\xi}{2}+\frac{\xi}{4}\le\xi.
\end{align*}
This gives the construction of uniform approximation of log-sum-exp operator for $x\in \bR^n$. Then we summarize its network parameters and its residual realization. 

\paragraph{Step 4: network budgets and residual implementation.}
Compose the preceding networks without multiplying adjacent affine matrices: an intermediate signed vector $v$ can be represented by $(\relu(v),\relu(-v))$ and reconstructed in the next affine layer. This incurs only constant-factor overhead in the corresponding layer widths and nonzero coefficients; see also Lemmas~F.1--F.3 in \citet{oko2023diffusion}. The scalar $a(x)$ can be carried through the exponential and logarithm layers and then recovered using two signed channels. Therefore,
\begin{align*}
 d
 &\le C\big(\log(en)+d_{\exp}+d_{\log}+1\big)\le Ct^2,\\
 W
 &\le C\big(n+nW_{\exp}+W_{\log}\big)
 \le C(nt+t^3)\le Cnt^3,\\
 S
 &\le C\big(n\log(en)+nS_{\exp}+S_{\log}
                +d_{\exp}+d_{\log}+n\big)
 \le C(nt^2+t^8)\le Cnt^8,\\
 B
 &\le C\max\{1,n,B_{\exp},B_{\log}\} \leq \exp\big(O(t^2)\big).
\end{align*}
Here sparsities add across the parallel exponential branches; the signed carry channels contribute only $O(d_{\exp}+d_{\log})$ coefficients.

Apply Lemma~\ref{lem:nn-compilation} to this circuit, retaining a constant input coordinate to supply biases. With all attention updates zero, the resulting rowwise residual FFNs satisfy
\begin{align*}
 N_{\rm FFN}&\le C(d+1)\le Ct^2,\\
 D_{\rm aux},\ D_{\rm ff}&\le C(W+n+1)\le Cnt^3,\\
 S_{\rm FFN}
 &\le C\big(S+(d+1)(W+n+1)\big)
 \le C(nt^8+nt^5)\le Cnt^8.
\end{align*}
The maximum Frobenius norm $B_{\rm F}$ of the FFN parameter matrices obeys
\[
 B_{\rm F}\le C(W+n+1)(B+1) \leq \exp \big(O(t^2)\big).
\]
These estimates include the reusable workspaces and their reset operations.

\paragraph{Proof of \eqref{eq:nn-lse-con}.} Let $r:=\|x-y\|_\infty$. Then
\[
 y_j-r\le x_j\le y_j+r\quad\text{for every }j
 \quad\Longrightarrow\quad
 e^{-r}\sum_{j=1}^n e^{y_j}
 \le\sum_{j=1}^n e^{x_j}
 \le e^r\sum_{j=1}^n e^{y_j}.
\]
Taking logarithms yields
$\LSE(y)-r\le\LSE(x)\le\LSE(y)+r$, and thus proves the 1-Lipschitz continuity.
\section{Parallelization error}
\label{app:TC and DTC parallel error}
This appendix follows \citet{chen2025optimal} to bound the parallelization error in \eqref{eq:TC and DTC control for parallelization error}. 
We include the derivations here to make the presentation self-contained. 

Recall the definition of TC (\textit{total correlation}) and DTC (\textit{dual total correlation}):
\[
    \TC(p) := \sum_{i=1}^L H(X_i) - H(X), \quad
    \DTC(p) := H(X) - \sum_{i=1}^L H(X_i\mid X_{[L]\setminus \{i\}}),
\]
where $X\sim p$ and $H$ denotes the entropy.

We first write the parallelization error using conditional total correlation, and then construct the TC and DTC schedules following Section 5 of \citet{chen2025optimal}.

Fix a distribution $p^\star$ on $\mathbb V^L$ and a forward masking schedule $s=(s_1,\ldots,s_T)$, with $s_t\in\mathbb N_+$ and $\sum_{t=1}^T s_t=L$. 
Recall the index sets $S^t$ and $W^t$ defined in the reverse process in Section \ref{sec:prelim}. 
The partition sets $ S:=(S^1,\ldots,S^T)$ are selected uniformly with $|S^t|=s_t$, and
\[
    W^t=\bigcup_{u=t+1}^T S^u.
\]
Thus, $W^t$ contains the unmasked positions before sampling $S^t$. 
Conditioned on $S$, the sampler using the exact conditional marginals has output distribution
\[
    p_{Y^\star \mid S}(x):=\prod_{t=1}^T\prod_{i\in S^t}p_i^\star(x_i\mid x_{W^t}).
\]

Recall the definition of the parallelization error in \eqref{eq:parallelization error}:
\begin{align}
    \varepsilon_{\parall}(p^\star;s)
    &:=\mathbb E_{ S}\big[\KL(p^\star\, \|\, p_{Y^\star \mid S})\big].
\end{align}
For sets $B,W\subseteq[L]$ and $B\cap W=\emptyset$, define the conditional TC as,
\[
    \TC_p(X_B\mid X_W)
    :=\sum_{i\in B}H(X_i\mid X_W)-H(X_B\mid X_W), \quad \text{with } X \sim p.
\]
Then notice that for a fixed unmasking partition $S$,
\begin{align*}
    \KL(p^\star\, \|\, p_{Y^\star \mid S}) &= \bE_{X\sim p^\star} \Big[ \log \frac{\prod_{t=1}^T p^\star(X_{S^t}\mid X_{W^t})}{\prod_{t=1}^T \prod_{i\in S^t}p_i^\star(X_i\mid X_{W^t})}\Big] \\
    & = \sum_{t=1}^T \bigg( \bE_{X\sim p^\star} \big[ \log p^\star(X_{S^t}\mid X_{W^t})\big] - \bE_{X\sim p^\star}\big[ \log \prod_{i\in S^t}p_i^\star(X_i\mid X_{W^t})\big] \bigg) \\
    & = \sum_{t=1}^T \TC_{p^\star}(X_{S^t}\mid X_{W^t}). 
\end{align*}
Hence, 
\begin{align}
    \varepsilon_{\parall}(p^\star;s) = \mathbb E_{S}\bigg[\sum_{t=1}^T \TC_{p^\star}(X_{S^t}\mid X_{W^t})\bigg], \label{eq:parallel-error-definition}
\end{align}

The following lemma provides bounds on the parallelization error for specific masking schedules.

\begin{lemma}[TC and DTC schedules]
\label{lem:tc-dtc-schedules}
Fix $L\geq2$ and an integer $T$ satisfying $2+2\log(eL)\leq T<L$. 
There exist forward masking schedules $s^{\TC}$ and $s^{\DTC}$, each consisting of $T$ positive integers summing to $L$ and depending only on $(L,T)$, such that, for every target distribution $p^\star$ on $\mathbb V^L$,
\begin{align}
    \varepsilon_{\parall}(p^{\star};s^{\TC})
    \leq\frac{6\log(eL)}{T}\TC(p^\star),
    \quad \varepsilon_{\parall}(p^\star;s^{\DTC})
    \leq\frac{6\log(eL)}{T}\DTC(p^\star).
    \label{eq:tc-dtc-schedule-bounds}
\end{align}
For $T=L$, the singleton schedule has zero parallelization error.
\end{lemma}

\paragraph{Proof of Lemma \ref{lem:tc-dtc-schedules}.}
We first write the parallelization error in terms of average mutual information. For $j\in[L]$, define
\[
    Z_j:=\mathbb E_{\substack{|W|=j-1\\i\notin W}}\big[I(X_i;X_W)\big],
    \qquad d_r:=Z_{r+1}-Z_r,\quad r\in[L-1],
\]
where $I(\cdot;\cdot)$ denotes the mutual information, $(W,i)$ is uniform among pairs of the indicated sizes and $X \sim p^\star$. 
For a fixed $i$, coupling nested uniform subsets of $[L]\setminus\{i\}$ gives $I(X_i;X_W)\leq I(X_i;X_{W'})$ whenever $W\subseteq W'$. Averaging over $i$ therefore gives $0=Z_1\leq\cdots\leq Z_L$, so $d_r\geq0$.

Let $H_j$ be the average of $H(X_W)$ over uniform subsets $W\subseteq[L]$ of size $j$, with $H_0=0$. Then by the connection between entropy and mutual information,
\begin{align}
    Z_j = \mathbb E_{\substack{|W|=j-1\\i\notin W}}\big[H(X_i)+ H(X_W) - H(X_{W\cup\{i\}}) \big]  = H_1 + H_{j-1} - H_j. \notag
\end{align}
Moreover, 
\begin{align*}
    \sum_{j=1}^L Z_j=LH_1-H_L=\TC(p^\star), \quad LZ_L=\sum_{i=1}^L I(X_i;X_{[L]\setminus\{i\}})=\TC(p^\star)+\DTC(p^\star). 
\end{align*}
Therefore, 
\begin{align}
    \TC(p^\star)=\sum_{r=1}^{L-1}(L-r)d_r,
    \qquad
    \DTC(p^\star)=\sum_{r=1}^{L-1}r\,d_r.
    \label{eq:tc-dtc-increments}
\end{align}

For the construction below, we list batch sizes in generation order as $s^{\rm rev}=(\tilde{s}_1,\cdots,\tilde{s}_T) := (s_T,\ldots,s_1)$.
Set $N_j:=\sum_{\ell=1}^j \tilde{s}_\ell$ as the total number of elements in the first $j$ unmasking batches, and $N_0=0$. 
Hence, $|W^j| = N_{T-j}$ and $|W^j \cup S^j| = N_{T-j+1}$. 
Consider
\begin{align}
    E_j &:= \bE_S \big[ \TC_{p^\star} (X_{S^j} \mid X_{W^j})\big]\notag \\
    &= \bE_S \big[ \sum_{i\in S^j} H(X_i \mid X_{W^j}) - H(X_{S^j} \mid X_{W^j}) \big] \notag \\
    & = s_j (H_1 - Z_{N_{T-j}+1}) - (H_{N_{T-j+1}} - H_{N_{T-j}}) \notag\\
    & = \sum_{v=N_{T-j}+1}^{N_{T-j+1}}(Z_v - Z_{N_{T-j}+1}) = \sum_{r=N_{T-j}+1}^{N_{T-j+1}-1}(N_{T-j+1}-r)d_r. \label{eq:batch-parallel-cost}
\end{align}
Summing over batches gives
\begin{align}
    \varepsilon_{\parall}(p^\star;s)=\sum_{j=1}^T E_j.   \label{eq:parallel-cost-sum}
\end{align}

\paragraph{TC schedule.}
Fix an integer $2\leq\zeta\leq L$. 
Starting with $r_0 = L$ remaining positions in the reverse process, for $j=1,\ldots$, take a batch of size $\tilde{s}^{\TC}_{j} = \lfloor r_{j-1}/\zeta\rfloor$ whenever $r_{j-1}\geq\zeta$, so that
\begin{align}
    r_{j}=r_{j-1}-\tilde{s}_j^{\TC} = r_{j-1}-\lfloor r_{j-1}/\zeta\rfloor. \label{eq:def of r}
\end{align}
Once fewer than $\zeta$ positions remain, generate them in singleton batches, i.e., $\tilde{s}_{j}^{\TC}=1$. 
Since $r_{j-1}$ and $\zeta$ are integers, it holds that $\lfloor r_{j-1}/\zeta\rfloor \ge r_{j-1}/\zeta - 1 + 1/\zeta$ and thus
\[
    r_{j}-\zeta+1
    \leq\bigg(1-\frac1\zeta\bigg)(r_{j-1}-\zeta+1).
\]
Therefore, at most
\[
    \lambda:=\bigg\lfloor
    \frac{\log(L-\zeta+1)}{-\log(1-1/\zeta)}
    \bigg\rfloor+1\leq1+\zeta\log L
\]
recursive steps are needed to reach $r_j<\zeta$. 
Including the remaining singleton batches, the total number of batches satisfies $m\leq\lambda+\zeta-1\leq \zeta\log(eL)$.
Define $r_j = r_{j-1} - \tilde{s}_j^{\TC}$ throughout the construction including the final singleton batches, so that $r_m=0$.
The corresponding forward masking schedule $s^{TC} = (\tilde{s}_m^{\TC},\ldots,\tilde{s}_1^{\TC})$. 

For $s^{\TC}$, the corresponding $N_j^{\TC} := \sum_{i=1}^j \tilde{s}_i^{\TC} = L-r_j$.
Hence, by \eqref{eq:batch-parallel-cost}, 
\begin{align*}
    E_{m-j+1}  = \sum_{r=N_{j-1}^{\TC}+1}^{N_{j}^{\TC}-1} (N_{j}^{\TC}-r) d_r.
\end{align*}
For every nonsingleton batch, we have 
\begin{align*}
    \tilde{s}_j^{\TC} = \lfloor r_{j-1}/\zeta\rfloor \leq r_{j-1}/\zeta = \frac{r_j + \tilde{s}_j^{\TC}}{\zeta}, \quad \text{ which implies } \, \tilde{s}^{\TC}_j \le \frac{r_j}{\zeta-1}.
\end{align*}
Therefore, for any $N_{j-1}^{\TC}\leq r \leq N_{j}^{\TC}-1$, it holds that
\[
    N_{j}^{\TC}-r \leq N_{j}^{\TC}-N_{j-1}^{\TC} = \tilde{s}_j^{\TC} \leq \frac{L-N_j^{\TC}}{\zeta-1} \leq \frac{L-r}{\zeta-1}.
\]
Hence, by \eqref{eq:parallel-cost-sum}
\begin{align}
    \varepsilon_{\parall}(p^\star;s^{\TC}) = \sum_{j=1}^m E_{m-j+1} \le \sum_{j=1}^m \sum_{r=N_{j-1}^{\TC}+1}^{N_{j}^{\TC}-1} \frac{L-r}{\zeta-1} d_r \le \frac{\TC(p^\star)}{\zeta-1}. \label{eq:tc-zeta-bound}
\end{align}
Here we use the fact that singleton batches contribute zero error, together with the expression of TC in \eqref{eq:tc-dtc-increments}.

\paragraph{DTC schedule.}
For this schedule, we just take the reverse of $s^{\TC}$. 
That is, $s^{\DTC} = \tilde{s}^{\TC}$. 
Similarly, for $s^{\DTC}$, the corresponding $N_j^{\DTC} := \sum_{i=1}^j \tilde{s}_i^{\DTC} = \sum_{i=1}^j s_i^{\TC} = r_{m-j}$, with $r$ defined in \eqref{eq:def of r}.

For every nonsingleton batch, its corresponding TC batch is also nonsingleton, and thus
\begin{align*}
    \tilde{s}_j^{\DTC} = \tilde{s}_{m-j+1}^{\TC} \le \frac{r_{m-j+1}}{\zeta-1} = \frac{N_{j-1}^{\DTC}}{\zeta-1}.
\end{align*}
Hence for every $N_{j-1}^{\DTC}\leq r \leq N_{j}^{\DTC}-1$: 
\begin{align*}
    N_j^{\DTC} - r \le \tilde{s}_j^{\DTC} \le \frac{N_{j-1}^{\DTC}}{\zeta-1} \le \frac{r}{\zeta-1}. 
\end{align*}
Notice that singleton batches contribute zero error, so we have
\begin{align}
    \varepsilon_{\parall}(p^\star;s^{\DTC}) = \sum_{j=1}^m E_{m-j+1} \le \sum_{j=1}^m \sum_{r=N_{j-1}^{\DTC}+1}^{N_{j}^{\DTC}-1} \frac{r}{\zeta-1} d_r \le \frac{\DTC(p^\star)}{\zeta-1}. \label{eq:dtc-zeta-bound}
\end{align}
Here we apply the expression of DTC in \eqref{eq:tc-dtc-increments}. 

\paragraph{Determine $\zeta$.}
Finally, we choose $\zeta$ and refine the two lists to obtain exactly $T$ batches. Let
\[
    A_L:=\log(eL),\qquad
    \zeta:=\bigg\lfloor\frac{T-2}{A_L}\bigg\rfloor.
\]
The assumption on $T$ ensures $2\leq\zeta\leq L$, and each list has at most $\zeta A_L\leq T-2$ batches.
By \eqref{eq:batch-parallel-cost}, splitting a batch cannot increase the parallelization error. 
Repeatedly splitting nonsingleton batches thus produces exactly $T$ batches, while preserving the bounds in \eqref{eq:tc-zeta-bound} and \eqref{eq:dtc-zeta-bound}.
Denote these adjusted forward schedules by our $s^{\TC}$ and $s^{\DTC}$, which depends only on $(L,T)$.

For $x=(T-2)/A_L\geq2$, the inequality $\lfloor x\rfloor-1\geq x/3$ implies
\[
    \frac1{\zeta-1}\leq\frac{3A_L}{T-2}
    \leq\frac{6A_L}{T},
\]
where the last inequality uses $T\geq4$.
Substituting this into \eqref{eq:tc-zeta-bound} and \eqref{eq:dtc-zeta-bound} proves \eqref{eq:tc-dtc-schedule-bounds}. 
For $T=L$, take singleton batches, for which every term in \eqref{eq:parallel-cost-sum} is zero.

\end{document}